\documentclass{article}

\usepackage{iclr,times}

\usepackage[utf8]{inputenc} 
\usepackage[T1]{fontenc}    
\usepackage{hyperref}       
\usepackage{url}            
\usepackage{booktabs}       
\usepackage{amsfonts}       
\usepackage{nicefrac}       
\usepackage{microtype}      
\usepackage{xcolor}         

\usepackage{amsmath,bbm,bm,graphicx,amsthm}
\usepackage{subfigure}

\usepackage[ruled]{algorithm2e}

\usepackage[utf8]{inputenc}
\usepackage{authblk}

\newtheorem{theorem}{Theorem}

\DeclareMathOperator{\Tr}{Tr}

\title{On the Power-Law Hessian Spectrums in Deep Learning}

\author[1]{Zeke Xie}
\author[2]{Qian-Yuan Tang}
\author[1]{Yunfeng Cai}
\author[1]{Mingming Sun}
\author[1]{Ping Li}
\affil[1]{Cognitive Computing Lab, Baidu Research}
\affil[2]{Lab for Neural Computation and Adaptation, RIKEN Center for Brain Science}

\affil[ ]{\textit {xiezeke@baidu.com}}

\iclrfinalcopy 
\begin{document}

\maketitle
 
\begin{abstract}
It is well-known that the Hessian of deep loss landscape matters to optimization, generalization, and even robustness of deep learning. Recent works empirically discovered that the Hessian spectrum in deep learning has a two-component structure that consists of a small number of large eigenvalues and a large number of nearly-zero eigenvalues. However, the theoretical mechanism or the mathematical behind the Hessian spectrum is still largely under-explored. To the best of our knowledge, we are the first to demonstrate that the Hessian spectrums of well-trained deep neural networks exhibit simple power-law structures. Inspired by the statistical physical theories and the spectral analysis of natural proteins, we provide a maximum-entropy theoretical interpretation for explaining why the power-law structure exist and suggest a spectral parallel between protein evolution and training of deep neural networks. By conducing extensive experiments, we further use the power-law spectral framework as a useful tool to explore multiple novel behaviors of deep learning.
\end{abstract}

\section{Introduction}

It is well-known that the Hessian matters to optimization, generalization, and even robustness of deep learning \citep{li2020hessian,ghorbani2019investigation,zhao2019bridging,jacot2019asymptotic,yao2018hessian,dauphin2014identifying,byrd2011use}. 
Deep learning usually finds flat minima that generalize well \citep{hochreiter1995simplifying,hochreiter1997flat}. The Hessian is one of the most important measures of the minima flatness and directly relates to generalization in deep learning \citep{hoffer2017train,neyshabur2017exploring,dinh2017sharp,wu2017towards,tsuzuku2020normalized}. \citet{jiang2019fantastic} reported that minima-flatness-based generalization bound is still the most reliable metric in extensive experiments. \citet{wu2017towards} reported that the low-complexity solutions that generalize well have a small norm of Hessian matrix with respect to model parameters. \citet{yao2018hessian} reported that the spectrum of the Hessian closely connects to large-batch training and adversarial robustness.

A number of works empirically studied the Hessian in deep neural networks. Some papers \citep{sagun2016eigenvalues,sagun2017empirical,wu2017towards} empirically reported a two-component structure that, in the context of deep learning, most eigenvalues of the Hessian are nearly zero, while a small number of eigenvalues are large. \citet{sankar2021deeper} revealed that the layerwise Hessian spectrum is similar to the entire Hessian spectrum. However, the theoretical mechanism behind the spectrum is under-explored. 

\textbf{Motivation.} Why does the Hessian spectrum consist of a small number of large eigenvalues and a large number of nearly zero eigenvalues? Does an elegant mathematical structure hide behind the Hessian spectrum? Inspired by protein science, our work provides a novel tool to understand and analyze deep learning from a spectral perspective.

\textbf{Contributions.} This paper has two main contributions. 
\begin{enumerate}
\item First, to the best of our knowledge, we are the first to empirically discover and mathematically model the power-law Hessian spectrum in deep learning. We theoretically formulated a novel maximum entropy interpretation for explaining the power-law spectrum.
\item Second, we propose a framework of power-law spectral analysis for deep learning. We not only reveal how the power-law spectrum explains the theoretical origin of striking findings but also empirically demonstrate multiple novel insights of deep learning.
\end{enumerate}

\section{The Power-Law Hessian Spectrum}

\begin{figure*}
\center
\subfigure[MNIST]{\includegraphics[width =0.24\columnwidth ]{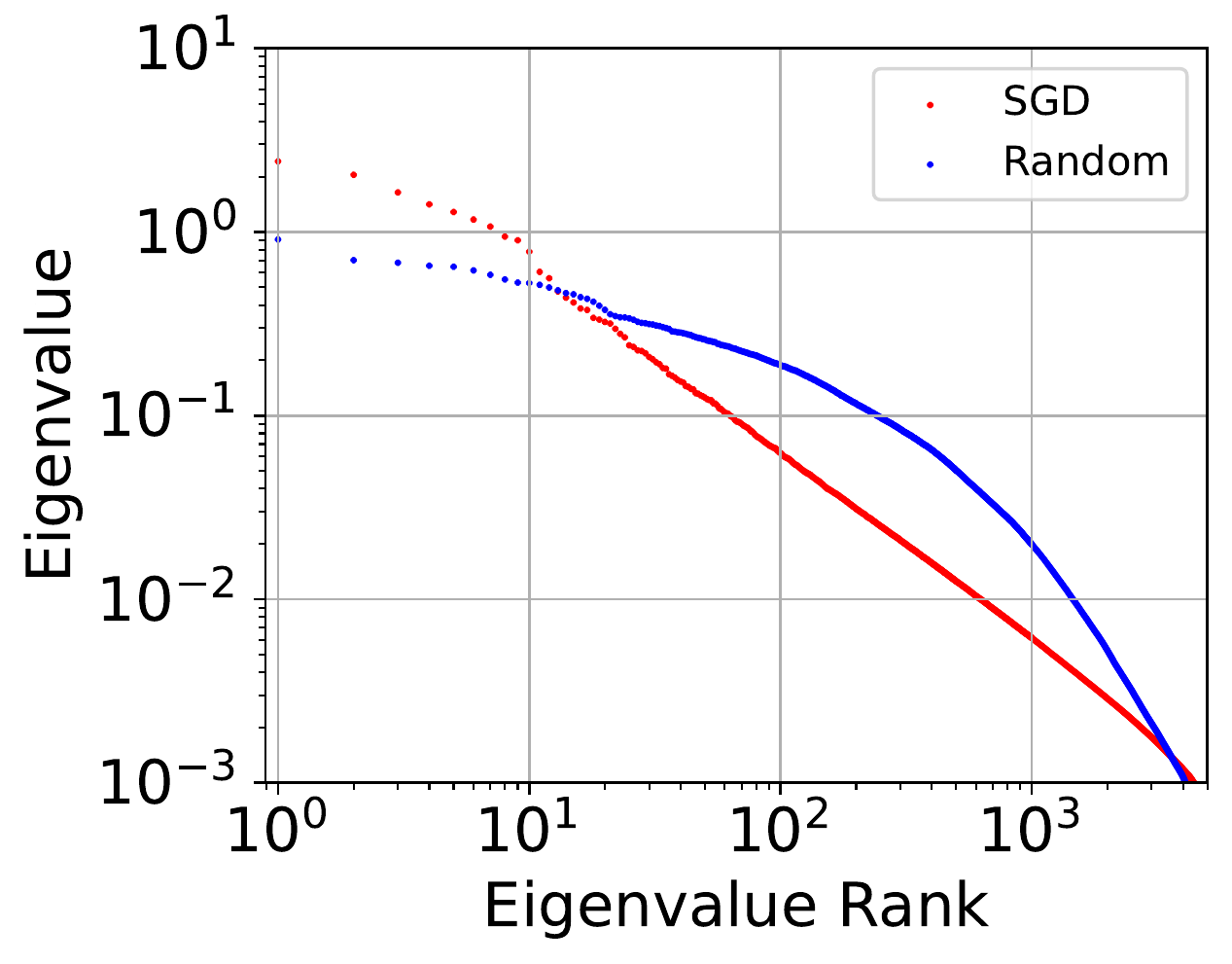}} 
\subfigure[Fashion-MNIST]{\includegraphics[width =0.24\columnwidth ]{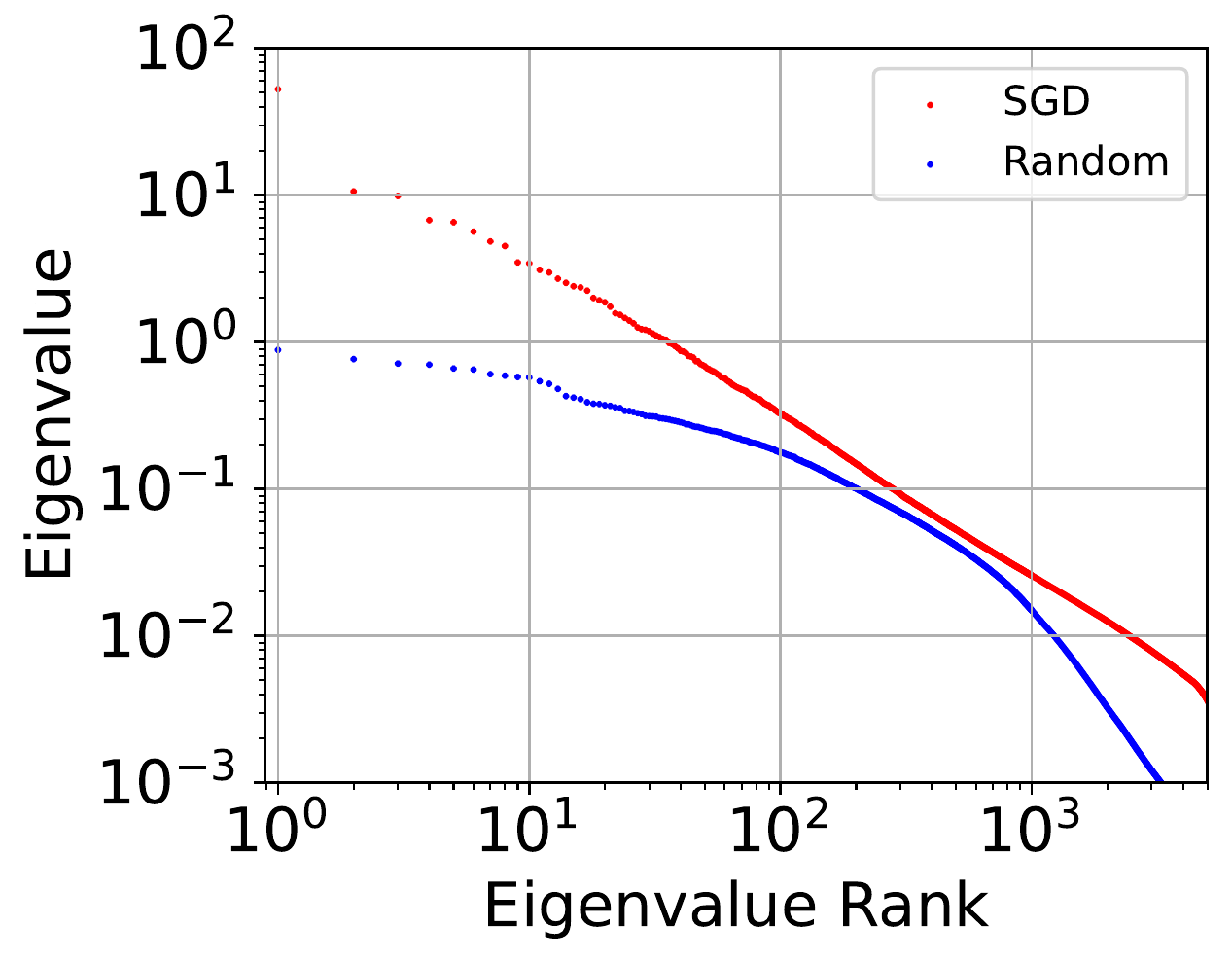}}
\subfigure[CIFAR-10]{\includegraphics[width =0.24\columnwidth ]{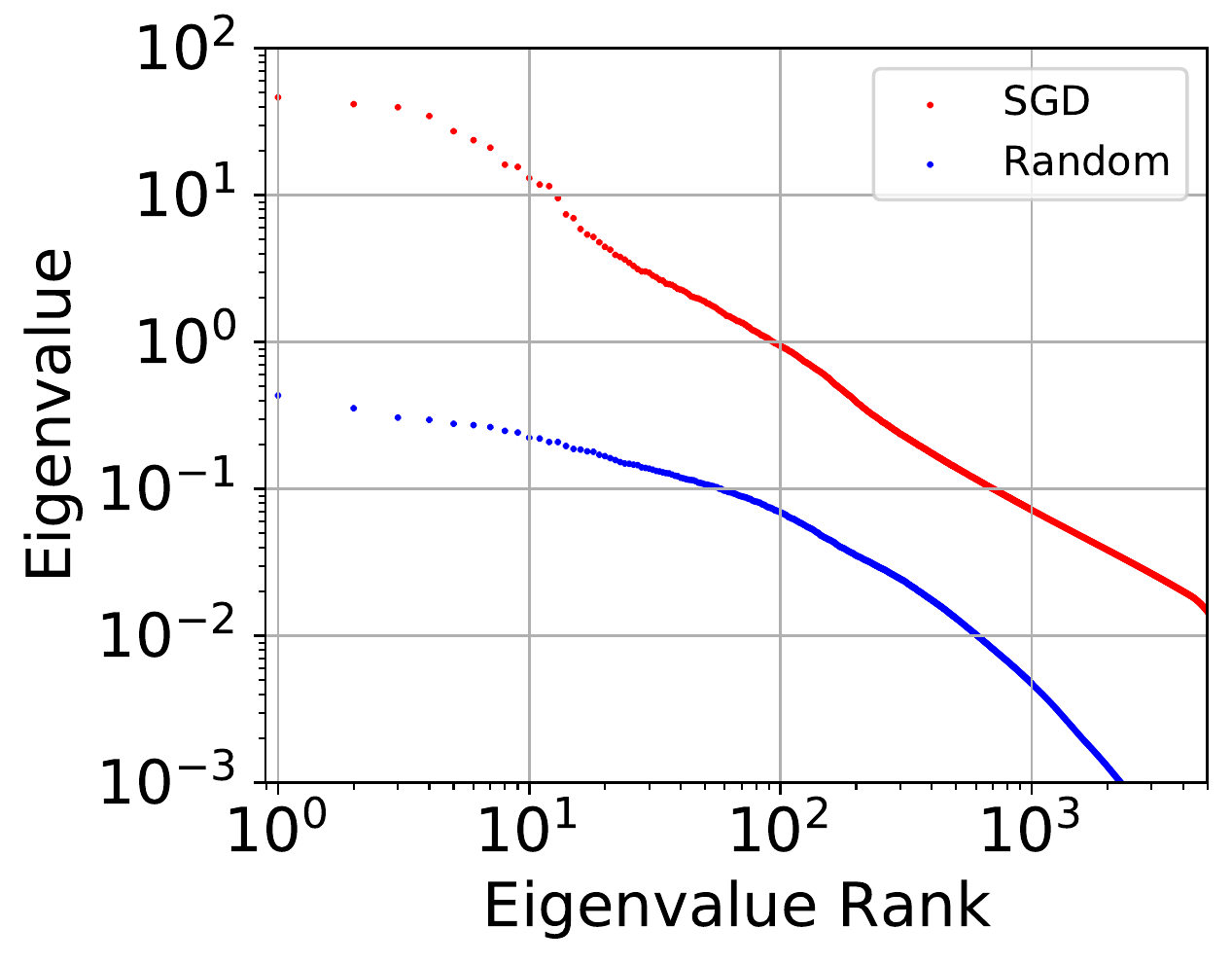}}  
\subfigure[CIFAR-100]{\includegraphics[width =0.24\columnwidth ]{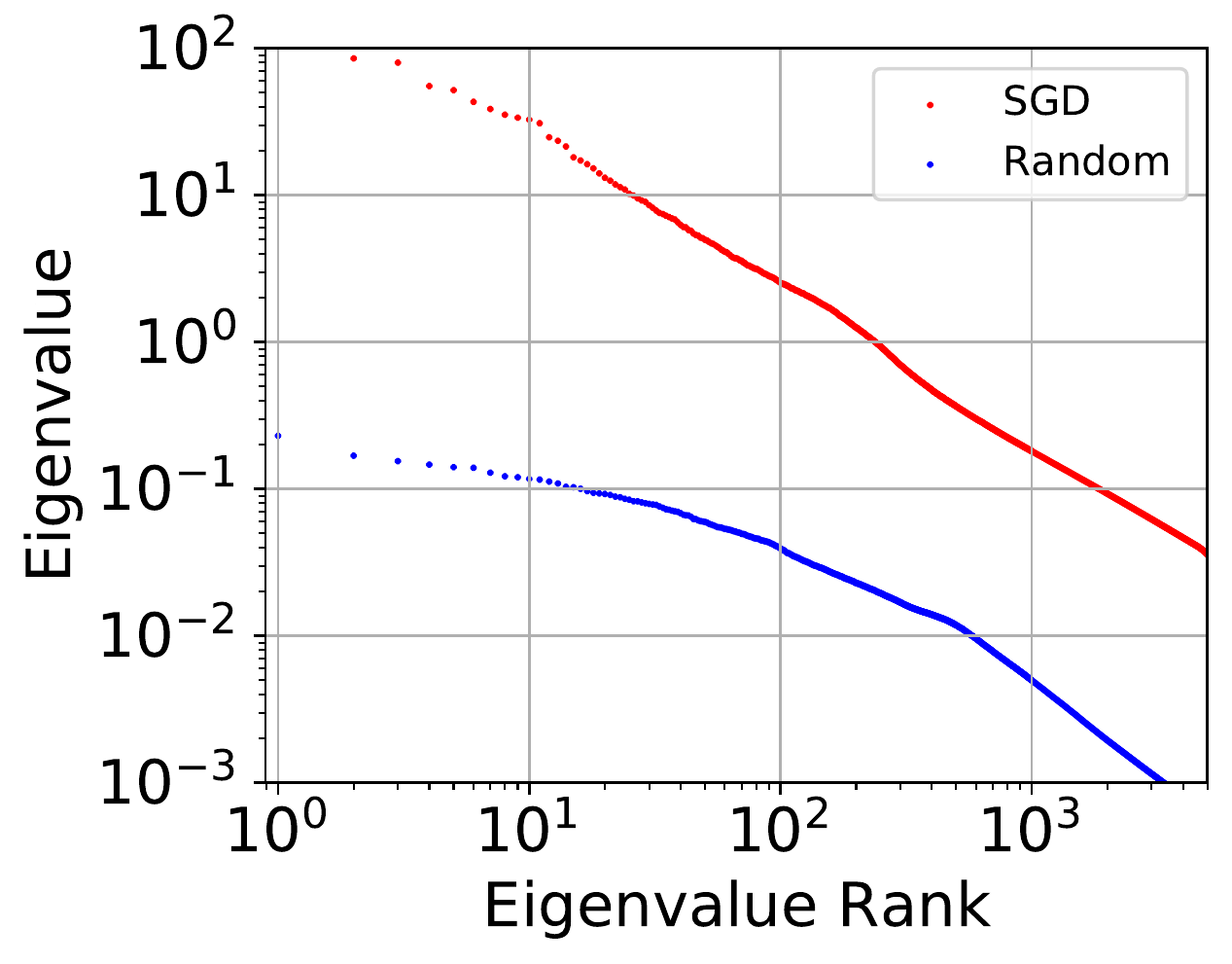}}  
\caption{ The power-law structure of the Hessian spectrum in deep learning. Model: LeNet. We may clearly observe that the power-law spectrums generally hold for well trained deep networks on various natural or artificial datasets, while do not hold for random neural networks. We also report that a small number of outlier eigenvalues ($\sim$10) slightly deviate from the fitted straight line.}
 \label{fig:powerlaw}
\end{figure*}

In this section, we demonstrate that the Hessian spectrums of well-trained deep neural networks have a simple power-law structure and how to theoretically derive the power-law structure. We also show that the discovered power-law structure provides novel insights and helps understand important properties of deep learning.

\textbf{Notations.} We denote the training dataset as $\{ (x, y) \} = \{(x_{j}, y_{j})\}_{j=1}^{N} $ drawn from the data distribution $\mathcal{S}$, the $n$ model parameters as $\theta$ and the loss function over one data sample $\{(x_{j}, y_{j})\}$ as $l(\theta, (x_{j}, y_{j}))$. For simplicity, we further denote the training loss as $L(\theta) =  \frac{1}{N} \sum_{j=1}^{N} l(\theta, (x_{j}, y_{j}) )$. We write the descending ordered eigenvalues of the Hessian $H$ as $\{\lambda_{1}, \lambda_{2}, \dots, \lambda_{n}\}$ and denote the spectral density function as $p(\lambda)$.  

\subsection{The Power-Law Structure and Empirical Evidence}

Recent papers studied the Hessian but failed to reveal its elegant mathematical structure. For understanding the distribution of the Hessian spectrum better, we first visualize the Hessian spectrum of a well-trained neural network and a randomly initialized neural network by using the Lanczos algorithm \citep{meurant2006lanczos,yao2020pyhessian} to estimate the eigenvalues and spectral densities. In Figure \ref{fig:powerlaw}, we display the top 6000 eigenvalues and their corresponding rank order. Both axes are $\log$-scale. And we surprisingly discover an approximately straight line fits the Hessian spectrum of the well-trained neural network surprisingly well, except that a small number of outliers ($\sim$10) slightly deviate from the fitted straight line. To the best of our knowledge, this fitted power-law Hessian spectrums were not empirically discovered or theoretically discussed by previous papers in deep learning. 

The well-fitted straight line means that the observed distribution of the Hessian eigenvalues of trained neural networks approximately obey a power-law distribution,
\begin{align}
\label{eq:PLDensity}
p(\lambda) = Z_{c}^{-1} \lambda^{-\beta},
\end{align}
where $Z_{c}$ is the normalization factor. The observed eigenvalues can be considered as $n$ samples from the power-law distribution $p(\lambda)$. We may also use a corresponding finite-sample power law for describing the observed law as 
\begin{align}
\label{eq:finitePL}
f_{k}= \frac{\lambda_{k}}{\Tr(H)} = Z_{d}^{-1} k^{-\frac{1}{\beta-1}},
\end{align}
where $f$ is the trace-normalized eigenvalue, $k$ is the rank order, the trace $\Tr(H) = \sum_{k=1}^{n} \lambda_{k} $, and $Z_{d} = \sum_{k=1}^{n}  k^{-\frac{1}{\beta-1}}$ is the normalization factor for the finite-sample power law. Note that the finite-sample power law is also called Zipf's law. This can also be equivalently written as
\begin{align}
\label{eq:finitePL2}
\lambda_{k} = \lambda_{1}k^{-s},
\end{align}
if we let $s = \frac{1}{\beta-1}$ denote the power exponent of Zipf's law. The empirical power-law spectrums suggest that the estimated $\hat{\beta}$ is close to $2$ and the estimated $\hat{s}$ is close to $1$.

\subsection{A Maximum-Entropy Theoretical Interpretation}


The maximum entropy principle \citep{guiasu1985principle}, also named the maximum entropy prior, states that the probability distribution which best represents the current state of knowledge about a system at equilibrium is the one with the highest entropy. This principle indicates that if we have no prior knowledge for suspecting one state over any other, then all states can be considered equally likely for a system at equilibrium. 

We start our theoretical analysis by maximum entropy in deep learning. The logarithmic space volume is often regarded a kind of entropy in statistical physics \citep{visser2013zipf}. Note that flat minima have larger space volume reflected by $\det (H^{-1})$. It means maximizing the minima space volume for better generalization may be regarded as a kind of entropy maximization principle. Following \citet{visser2013zipf}, we may explicitly write the volume entropy as 
\begin{align}
\label{eq:VolEntropy}
S_{\rm{vol}} =   \log \det (H^{-1}) = \int p(\lambda) \log \lambda d \lambda
\end{align}
and the spectral entropy as
\begin{align}
\label{eq:SpecEntropy}
S_{\mathrm{p}} = - \int p(\lambda) \log p(\lambda) d \lambda,
\end{align}
which is the entropy of the spectral density distribution. 

Considering the principle of maximum entropy with the two kinds of entropy, we need to maximize the total entropy with the spectral density normalization constraint
\begin{align}
\label{eq:TotalEntropy}
S_{\mathrm{total}} =  - \int p(\lambda) \log p(\lambda) d \lambda  +   \beta_{\rm{vol}} \int p(\lambda) \log \lambda d \lambda -  \beta_{\rm{norm}}(\int p(\lambda) d \lambda - 1),
\end{align}
where $S_{\mathrm{total}} = S_{\mathrm{p}} + \beta_{\rm{vol}} S_{\rm{vol}}$ and $\beta_{\rm{norm}}$ is a Lagrange multiplier. To find the optimal distribution $p^{\star}(\lambda)$ that maximizes the total entropy, we require the following
\begin{align}
\label{eq:MaxEntropy}
\frac{\partial S_{\mathrm{total}}}{\partial p(\lambda)} = -  \log p(\lambda) -   \beta_{\rm{vol}}  \log \lambda - \beta_{\rm{norm}}= 0 .
\end{align}
Thus, the optimal distribution $p^{\star}(\lambda)$ can be solved as
\begin{align}
\label{eq:OptimP}
p^{\star}(\lambda) =  e^{-\beta_{\rm{norm}}} \lambda^{\beta_{\rm{vol}}},
\end{align}
which has an amazingly similar form to Equation \eqref{eq:PLDensity} with $\beta_{\rm{norm}}= \log Z_{c}$ and $\beta = -\beta_{\rm{vol}}$.

We may interpret the power-law structure of the Hessian spectrum from two simple maximum entropy principles with the spectral density normalization constraint. It roughly means that simple rules can almost explain the power-law Hessian spectrum in deep learning. The spectrums have much simpler structures than previous work expected. 

Interestingly, similar well-fitted power laws have been widely discussed in neuroscience \citep{stringer2019high} and biology \citep{reuveni2008proteins,tang2020long}. This is exactly our motivation to further verify and study the power-law structure of the Hessian spectrum in the context of deep learning. We verify the elegant power-law structure indeed exists in well-trained deep neural networks just like bioactive proteins. In contrast, random neural networks have no such a power-law structure, just like deactivated (denatured or unfolded) proteins. Random neural networks which have no functional ability on the given task break the power-law spectrums similarly to deactivated proteins.

Statistical physical theories of neural networks \citep{bahri2020statistical,torlai2016learning,teh2003energy} supports that randomly initialized neural networks are far from the equilibrium, while well-trained neural networks are more close to equilibrium. The equilibrium condition may distinguish well-trained neural networks and randomly initialized neural networks and explains why the power-law structure breaks without training neural networks. However, we also note that there are still arguable disputes on the equilibrium of DNNs, which is beyond the main scope of this paper.

\subsection{Goodness-of-fit Test}

Following \citet{alstott2014powerlaw}, we use Maximum Likelihood Estimation (MLE) \citep{myung2003tutorial} for estimating the parameter $\beta$ of the fitted power-law distributions and the Kolmogorov-Smirnov Test (KS Test) \citep{massey1951kolmogorov,goldstein2004problems} for statistically testing the goodness of the fit. The KS test statistic is the KS distance $d_{\rm{ks}}$ between the hypothesized (fitted) distribution and the empirical data, which measures the goodness of fitting.

According to the practice of KS Test \citep{massey1951kolmogorov}, we first state {\it \textbf{the power-law hypothesis}} that the tested spectrum is power-law. If $d_{\rm{ks}}$ is higher than the critical value $d_{\rm{c}}$ at the $\alpha=0.05$ significance level, the KS test statistically will support (not reject) the power-law hypothesis. The test results associated with Figure \ref{fig:powerlaw} are presented in Table \ref{table:kslenet-mnist-cifar}. We leave the technical details and more test results (e.g. ResNet18) in Appendix \ref{sec:kstest}. 

When we say that a spectrum is (approximately) power-law in this paper, we mean that the KS test provides positive evidences to the power-law hypothesis instead of rejecting the power-law hypothesis. Our KS test results reject the power-law hypothesis for random neural networks and do not reject the power-law hypothesis for well-trained neural networks. Moreover, when the power-law hypothesis holds, the KS distance is usually significantly smaller than the critical value $d_{\rm{c}}$. For simplicity, the default $\alpha=0.05$ significance level is abbreviated in the following.

Following related work on the Hessian of neural networks \citep{thomas2020interplay}, our empirical analysis and statistical tests mainly focused on the top ($\sim 1000$) large eigenvalues larger than some minimal cutoff value $\lambda_{\rm{cutoff}}$ for three reasons. First, focusing on relatively large values is very reasonable and common in various fields' power-law studies, as real-world distributions typically follow power laws only after/large than some cutoff values \citep{clauset2009power} for ensuring the convergence of the probability distribution. Second, researchers are usually more interested in significantly large eigenvalues which contributes more to Hessian, minima sharpness, or generalization bound \citep{thomas2020interplay}. Third, empirically estimating a large number of nearly zero eigenvalues is very inaccurate and expensive.

\begin{table*}
\caption{The Kolmogorov-Smirnov statistics of the Hessian spectrums of LeNets on various datasets. The estimated power exponent $\hat{\beta}$ and slope magnitude $\hat{s}$ are also displayed.}
\label{table:kslenet-mnist-cifar}
\begin{center}
\begin{small}
\resizebox{\textwidth}{!}{%
\begin{tabular}{lll | lllll}
\toprule
Dataset & Model & Training   & $d_{\rm{ks}}$ & $d_{\rm{c}}$ & Power-Law & $\hat{\beta} \pm \sigma$ & $\hat{s}$ \\
\midrule
MNIST & LeNet & \textcolor{red}{Random}  & 0.0796 & 0.0430 &  \textcolor{red}{No}  \\ 
MNIST & LeNet & SGD    & 0.00900 & 0.0430 &  \textcolor{blue}{Yes}& $1.991 \pm 0.031$ & 1.009 \\  
\midrule
Fashion-MNIST & LeNet & \textcolor{red}{Random} & 0.0971 & 0.0430 &  \textcolor{red}{No}& \\  
Fashion-MNIST & LeNet & SGD  & 0.0132 & 0.0430 &  \textcolor{blue}{Yes}& $1.939 \pm 0.030 $ & 1.065\\
\midrule
CIFAR-10 & LeNet & \textcolor{red}{Random}   & 0.0663 & 0.0430 &  \textcolor{red}{No}& \\  
CIFAR-10 & LeNet & SGD   & 0.0279 & 0.0430 &  \textcolor{blue}{Yes}& $1.968 \pm 0.031 $ & 1.033 \\ 
\midrule
CIFAR-100 & LeNet & \textcolor{red}{Random}    & 0.0944 & 0.0430 &  \textcolor{red}{No}& \\  
CIFAR-100 & LeNet & SGD  & 0.0315 & 0.0430 &  \textcolor{blue}{Yes}& $1.908 \pm 0.029 $ & 1.101 \\
\bottomrule
\end{tabular}
}
\end{small}
\end{center}
\end{table*}


\subsection{Robust and Low-Dimensional Learning Space}
\label{sec:robustlow}

\begin{figure}
\center
\subfigure[Eigenvalue Rank]{\includegraphics[width =0.4\columnwidth ]{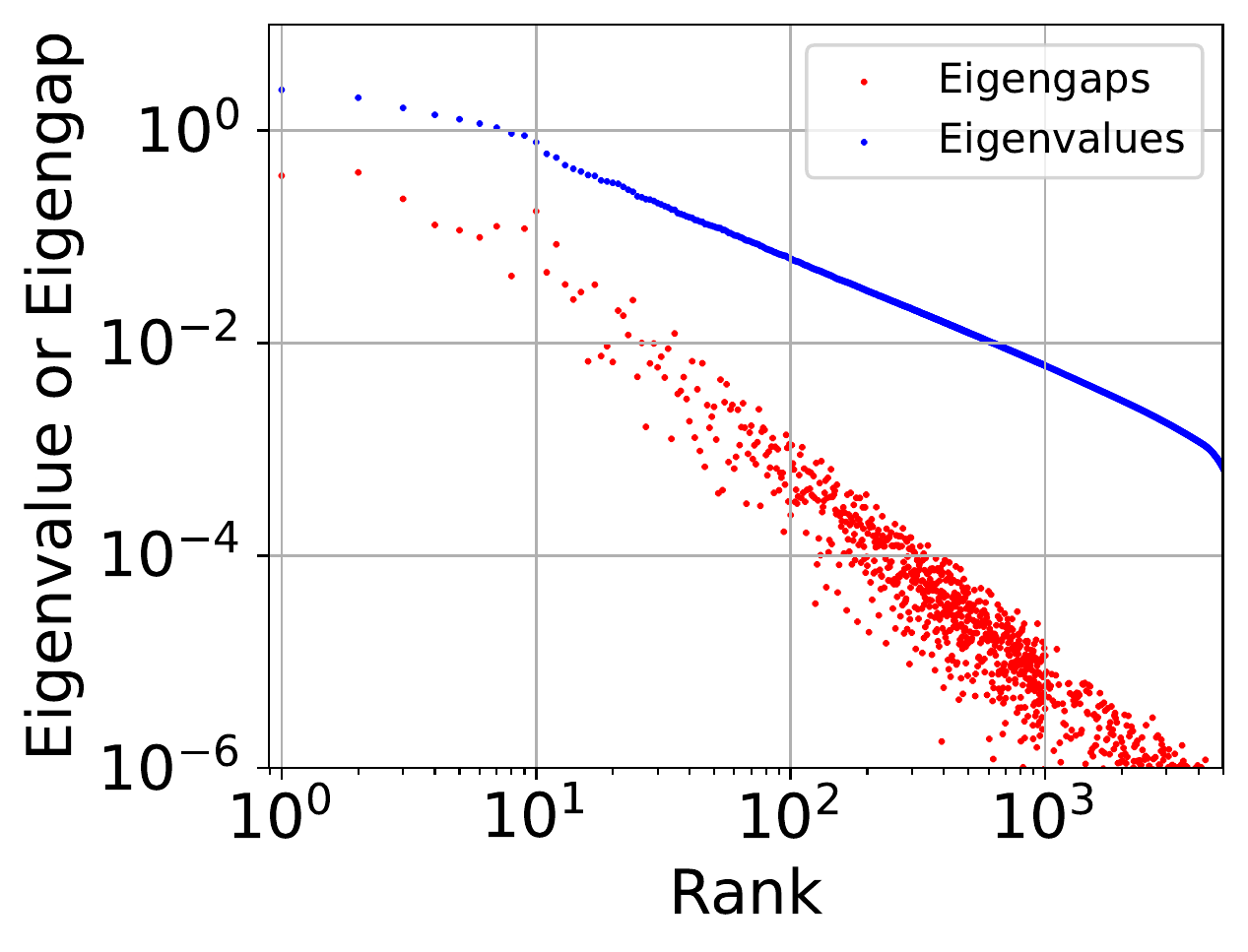}} 
\subfigure[Eigengap Rank]{\includegraphics[width =0.4\columnwidth ]{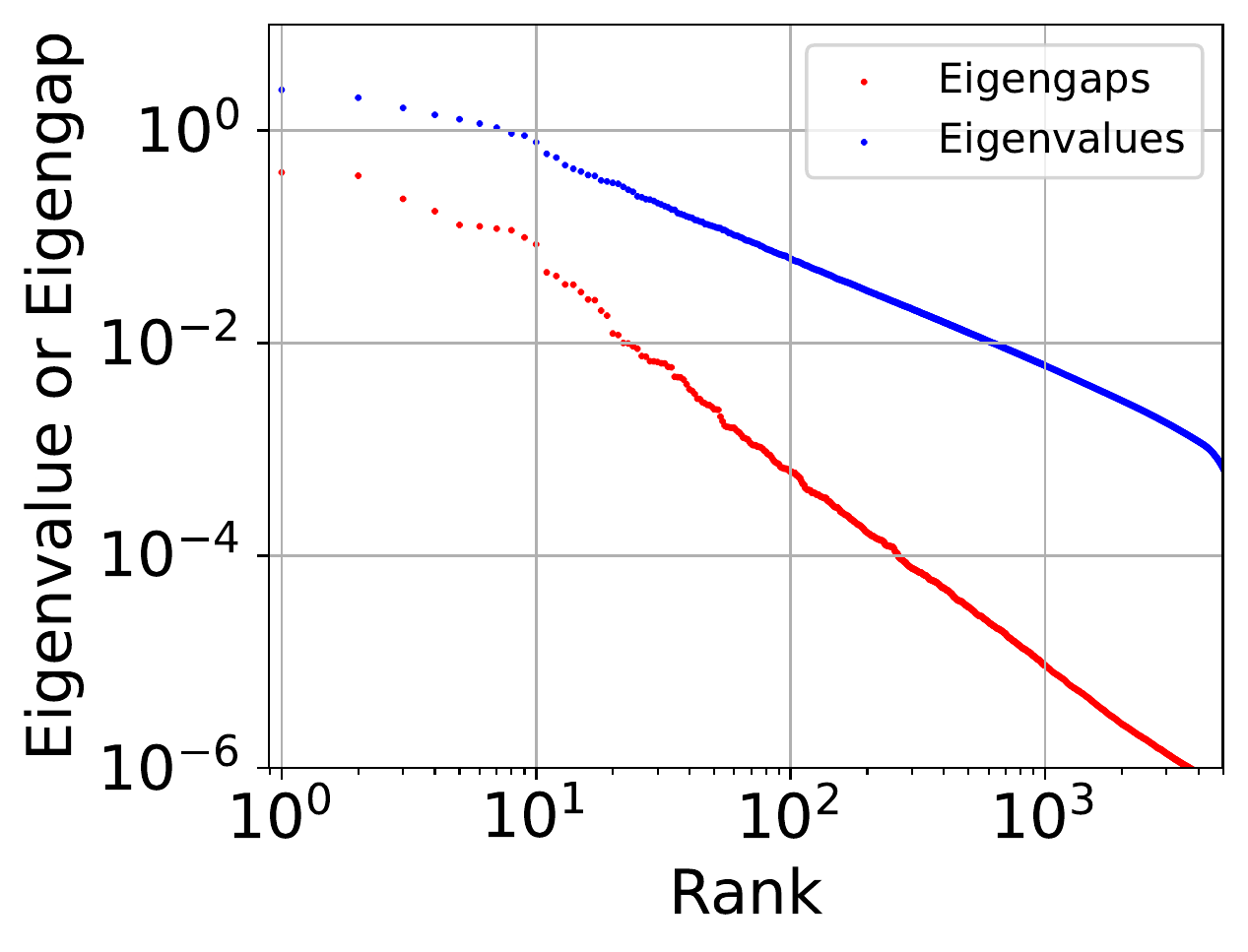}}
\caption{ The power-law Hessian eigengaps. Model: LeNet. Datsets: MNIST. Subfigure (a) displayed the eigengaps by original rank indices sorted by eigenvalues. Subfigure (b) displayed the eigengaps by rank indices re-sorted by eigengaps. We also present the results of Fashion-MNIST in Figure \ref{fig:fmnisteigengap} of Appendix \ref{sec:supresults}.}
 \label{fig:eigengap}
\end{figure}

\begin{table*}
\caption{The Kolmogorov-Smirnov statistics of the Hessian eigengaps on various datasets. The estimated power exponent $\hat{\beta}$ and slope magnitude $\hat{s}$ are also displayed.}
\label{table:kslenet-eigengap}
\begin{center}
\begin{small}
\resizebox{\textwidth}{!}{%
\begin{tabular}{lll | lllll}
\toprule
Dataset & Model & Training  & $d_{\rm{ks}}$ & $d_{\rm{c}}$ & Power-Law & $\hat{\beta} \pm \sigma$ & $\hat{s}$ \\
\midrule
MNIST & LeNet & SGD   & 0.0153 & 0.0430 &  \textcolor{blue}{Yes}& $1.550 \pm 0.017$ & 1.817 \\  
Fashion-MNIST & LeNet & SGD & 0.0240 & 0.0430 &  \textcolor{blue}{Yes}& $1.520 \pm 0.017$ & 1.922 \\  
\bottomrule
\end{tabular}
}
\end{small}
\end{center}
\end{table*}

\textbf{Deep learning happens in a low-dimensional space.} \citet{gur2018gradient} empirically observed that deep learning (via SGD) mainly happens in a low-dimensional space during the whole training process. \citet{ghorbani2019investigation} studied and reported that, throughout the optimization process, large isolated eigenvalues rapidly appear in the spectrum, along with a surprising concentration of the gradient in the corresponding eigenspace. \citet{xie2021diffusion} theoretically demonstrated that the learning space is a low-dimensional subspace spanned by the eigenvectors corresponding to large eigenvalues of the Hessian, because SGD diffusion mainly happens along these principal components. Note that the low-dimensional learning space implicitly reduces deep models' complexity. However, existing work cannot explain why the low-dimensional learning space is robust during training. In this paper, robust space means that the space's dimensions are stable during training.

We try to mathematically answer this question by studying the Hessian eigengaps. We define the $i$-th eigengap as $\delta_{k} = \lambda_{k} - \lambda_{k+1} $. According to Equation \eqref{eq:finitePL}, we have $\delta_{k}$ approximately meeting
\begin{align}
\label{eq:eigengap}
\delta_{k} =  \Tr(H) Z_{d}^{-1} (k^{-\frac{1}{\beta-1}} - (k+1)^{-\frac{1}{\beta-1}}) = \lambda_{k} \left[ 1 - (\frac{k}{k+1})^{s} \right].
\end{align}
Interestingly, it demonstrates that eigengaps also approximately exihibit a power-law distribution when $k$ is large. Particularly, we will have an approximate power law 
\begin{align}
\label{eq:PLeigengap}
\delta_{k} =  \Tr(H) Z_{d}^{-1} (k+1)^{- (s + 1)} 
\end{align}
under the approximation $s \approx 1$. The power exponent $s+1$ is larger than the one in Equation \eqref{eq:finitePL} by 1.

\textbf{Empirical analysis of the decaying Hessian eigengaps.} The empirical study about the Hessian eigengaps is missing in previous papers. Our experiments filled this gap. Our experiments show that top eigengaps dominate others in deep learning similarly to eigenvalues. We further empirically verified the approximate power-law distribution of the eigengaps in Figure \ref{fig:eigengap}. Moreover, the observation that the power exponent of eigengaps is larger than the power exponent of eigenvalues by $\sim1$ even fully matches our theoretical result by comparing Equations \eqref{eq:finitePL} and \eqref{eq:PLeigengap}. 

Note that the existence of top large eigenvalues does not necessarily indicate their gaps are also statistically large. Previous papers revealed that top eigenvalues dominate others but did not reveal if top eigengaps dominate others in deep learning. Fortunately, we theoretically and empirically demonstrate that both eigenvalues and eigengaps decay, are power-law as the rank order increase. Eigengaps even decay faster than eigenvalues due to the larger magnitude of the power exponent. We will show that this is the foundation of learning space robustness in deep learning.

\textbf{Eigengaps Bound Learning Space Robustness.} Based on the well-known Davis-Kahan $\sin (\Theta)$ Theorem \citep{davis1970rotation}, we use the angle of the original eigenvector $u_{k}$ and the perturbed eigenvector $\tilde{u}_{k}$, namely $\langle u_{k} , \tilde{u}_{k}\rangle$, to measure the robustness of space's dimensions. We directly apply Theorem \ref{pr:lspacerobustness}, a useful variant of Davis-Kahan Theorem \citep{yu2015useful}, to the Hessian in deep learning, which states that the eigenspace (spanned by eigenvector) robustness can be well bounded by the corresponding eigengap.

\begin{theorem}[Eigengaps Bound Eigenspace Robustness \citep{yu2015useful}]
 \label{pr:lspacerobustness}
Suppose the true Hessian is $H$, the perturbed Hessian is $\tilde{H} = H + \epsilon M$, the $i$-th eigenvector of $H$ is $u_{i}$ , and its corresponding perturbed eigenvector is $\tilde{u}_{i}$. Under the conditions of Davis-Kahan Theorem, we have 
 \[ \sin \langle u_{k} , \tilde{u}_{k}\rangle \leq \frac{2\epsilon \|M \|_{op}}{ \min( \lambda_{k-1} -  \lambda_{k},  \lambda_{k} -  \lambda_{k+1}) }, \]
 where $ \|M \|_{op}$ is the operator norm of $M$.
\end{theorem}

As we have a small number of large eigengaps corresponding to the large eigenvalues, the corresponding learning space robustness has a tight upper bound. For example, given the power-law eigengaps in Equation \eqref{eq:PLeigengap}, the upper bound of eigenvector robustness can be written as 
\begin{align}
\label{eq:robustspace}
\sup \sin \langle u_{k} , \tilde{u}_{k}\rangle =   \frac{2 \epsilon \|M \|_{op} (k+1)^{s+1} }{\lambda_{1} } ,  
\end{align}
which is relatively tight for top dimensions (small $k$) but becomes very loose for tailed dimensions (large $k$). A similar conclusion also holds given Equation \eqref{eq:eigengap}. As $s \approx 1$ suggests, the experimental results in Figure \ref{fig:eigengap} also well supports that the upper bound of $k=1000$ is $10^{4}$ times the upper bound of $k=10$. This indicates that non-top eigenspace can be highly unstable during training, because $\delta_{k} $ can decay to nearly zero for a large $k$. To the best of our knowledge, we are the first to demonstrate that the robustness of low-dimensional learning space directly depends on the eigengaps of the Hessian $H$.

\section{Related Works} 


A number of related works analyzed the spectral distribution of the Hessian in deep learning. \citet{pennington2017geometry} introduced an analytical framework from random matrix theory and reported that the shape of the spectrum depends strongly on the energy and the overparameterization parameter, $\phi$, which measures the ratio of parameters to data points. However, \citet{pennington2017geometry} mainly evaluated single-hidden-layer networks, which limits the scope of the conclusion. A followup work \citep{pennington2018spectrum} focused on a single-hidden-layer neural network with Gaussian data and weights in the limit of infinite width. Obviously, its theoretical and empirical analysis is far from practical deep models. \citet{jacot2019asymptotic} analyzed the limiting spectrum of the Hessian in neural networks with infinite width. \citet{fort2019goldilocks} analyzed the Hessian spectrums of initialized neural networks. \citet{fort2019emergent} studied the role of Hessian in learning in the low-dimensional subspace. \citet{papyan2019measurements} studied the three-level hierarchical structure and outliers in Hessian spectrums. \citet{liao2021hessian} studied the Hessian spectrums of more realistic nonlinear models. While a number of previous papers studied the Hessian spectrum, they failed to empirically discover the simple but important power-law structure and missed the theoretical interpretation.

\section{Empirical Analysis and Discussion}
\label{sec:empirical}

\begin{figure}
\begin{minipage}[t]{0.49\textwidth}
\centering
\subfigure{\includegraphics[width =0.49\columnwidth ]{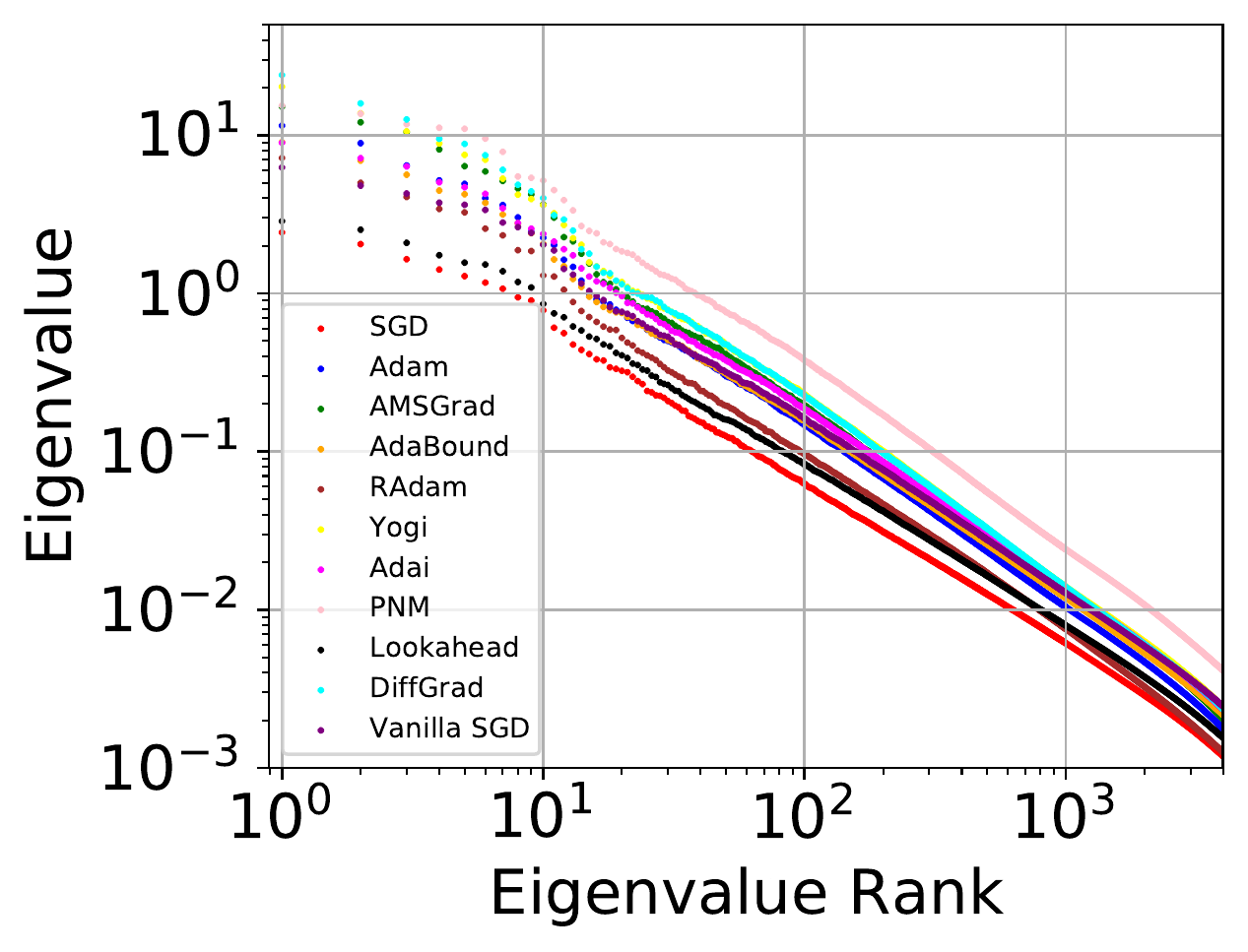}} 
\subfigure{\includegraphics[width =0.49\columnwidth ]{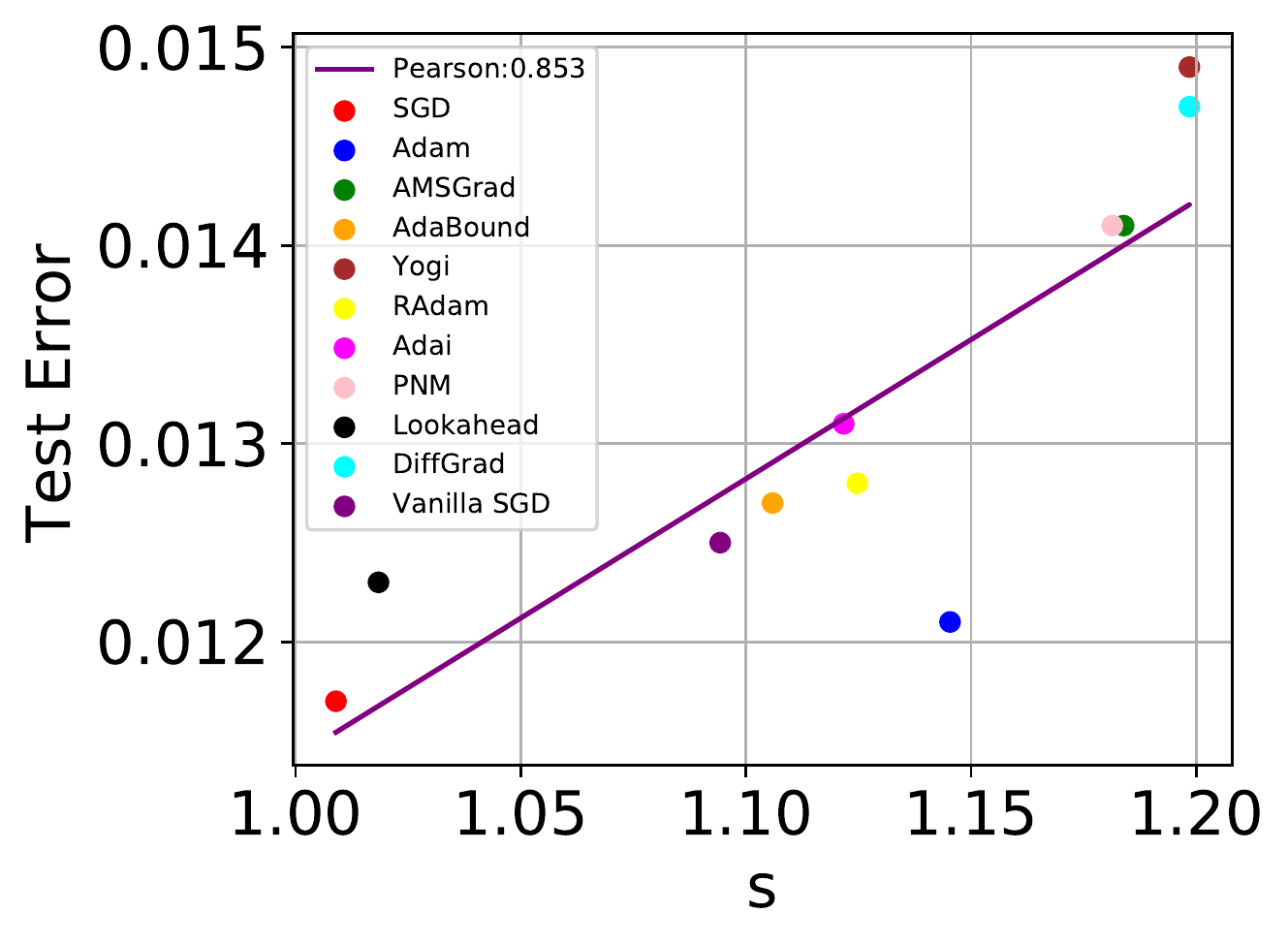}} 
\caption{The power-law spectrums hold across optimizers. Moreover, the slope magnitude $\hat{s}$ is an indicator of minima sharpness and a predictor of test performance. Model:LeNet. Dataset: MNIST.}
 \label{fig:optimizerspectrum-mnist}
\end{minipage} 
\hfill
\begin{minipage}[t]{0.24\textwidth}
\centering
\subfigure{\includegraphics[width =0.99\columnwidth ]{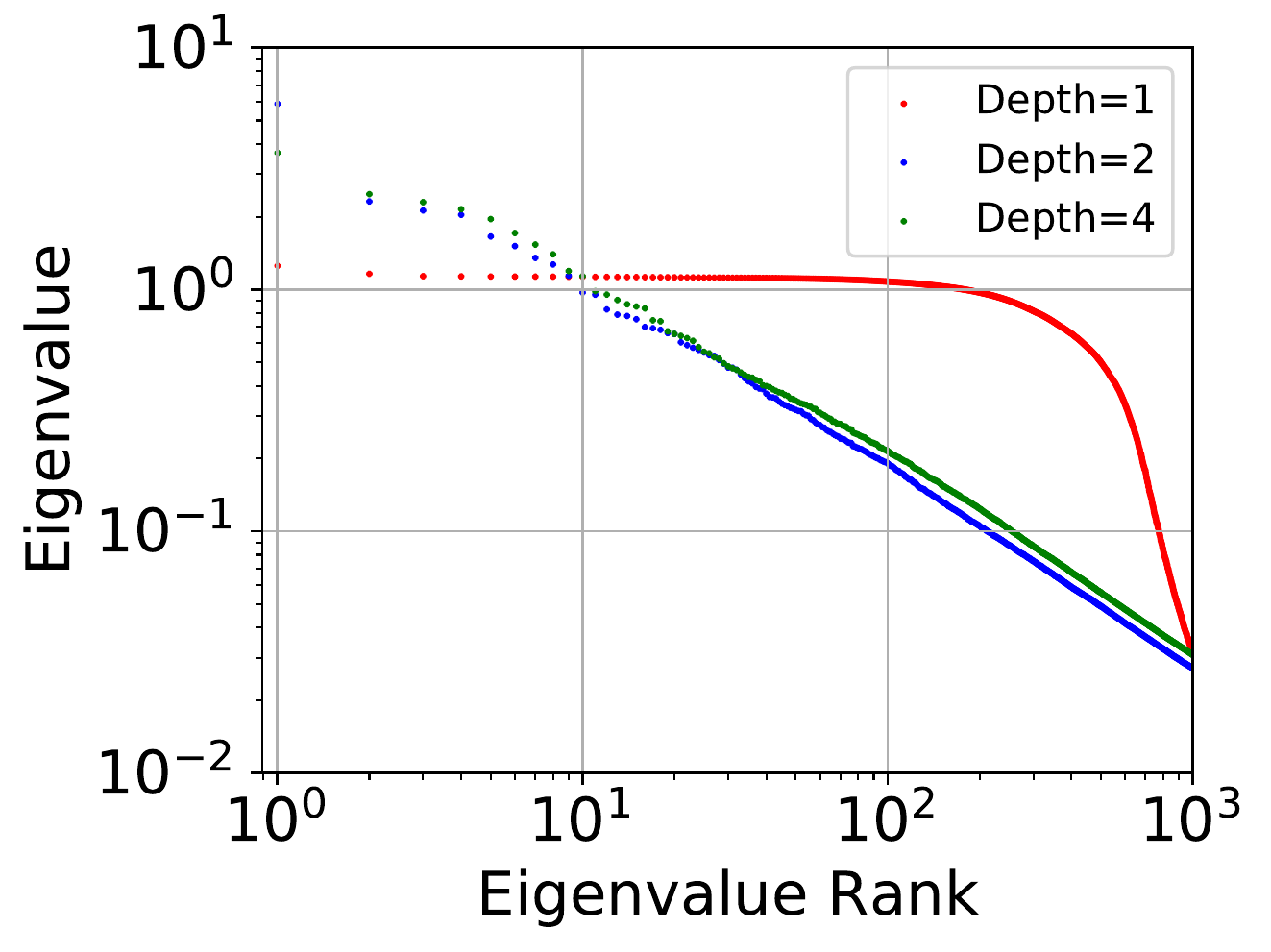}} 
\caption{The power-law spectrum holds well in overparameterized deep models but disappears in the underparameterized single-layer FCN.}
 \label{fig:depth}
 \end{minipage} 
 \hfill
\begin{minipage}[t]{0.24\textwidth}
\centering
\subfigure{\includegraphics[width =0.99\columnwidth ]{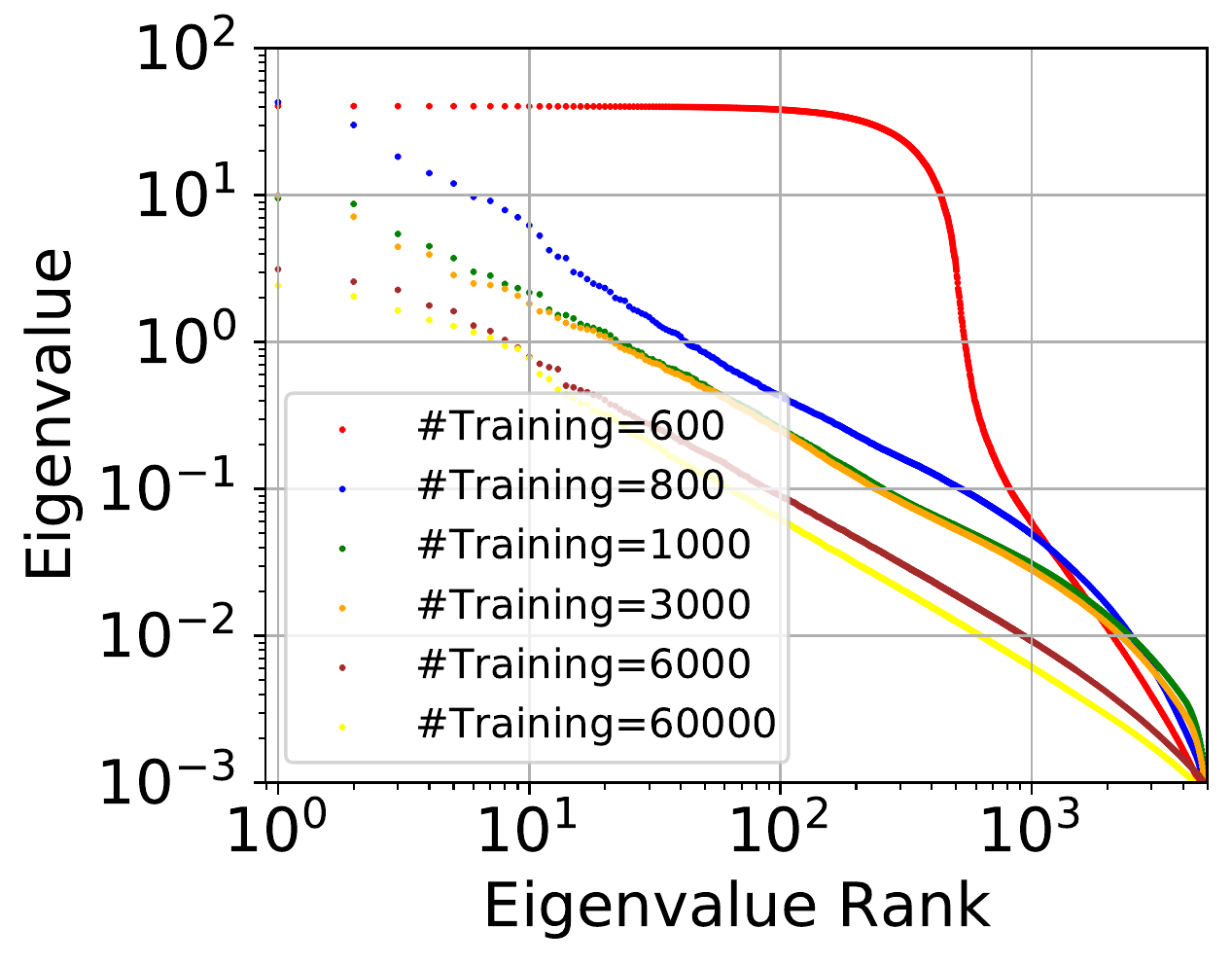}}
\caption{The spectrums of LeNet on MNIST with respect with various numbers of training samples.}
\label{fig:numtraining}
 \end{minipage} 
\end{figure}

In this section, we conduct extensive experimental results to explore novel behaviors of deep learning via power-law spectral analysis. 

\textbf{Model:} LeNet \citep{lecun1998gradient}, Fully Connected Networks (FCN), and ResNet18 \citep{he2016deep}. \textbf{Dataset}: MNIST \citep{lecun1998mnist}, Fashion-MNIST \citep{xiao2017fashion}, CIFAR-10/100 \citep{krizhevsky2009learning}, and non-image Avila \citep{de2018reliable}.

\textbf{1. Optimization and Generalization.} Figure \ref{fig:optimizerspectrum-mnist} discovered that the power-law spectrum consistently holds for various popular optimizers, such as SGD, Vanilla SGD, Adam \citep{kingma2014adam}, AMSGrad \citep{reddi2019convergence}, AdaBound \citep{luo2019adaptive}, Yogi \citep{zaheer2018adaptive}, RAdam \citep{liu2019variance}, Adai \citep{xie2022adaptive}, PNM \citep{xie2021positive}, Lookahead \citep{zhang2019lookahead}, and DiffGrad \citep{dubey2019diffgrad}, as long as the optimizers can train neural networks well.

We find that the slope magnitude $\hat{s}$ of the fitted straight line may serve as a nice indicator of minima sharpness and a predictor of test performance. It is common to measure minima sharpness by the largest Hessian eigenvalue or the Hessian trace. A smaller $\hat{s}$ highly correlates to a smaller largest eigenvalue and a smaller trace in Figure \ref{fig:slopesharpness-mnist} of Appendix \ref{sec:supresults}. The observation also holds on CIFAR-10 in Figures \ref{fig:optimizerspectrum-cifar10} and \ref{fig:slopesharpness-cifar10} of Appendix \ref{sec:supresults}. Interestingly, we also observe that $\hat{s} \approx 1$ is common in the spectrums of DNNs as well as the spectrums of natural proteins. 

\textbf{2. Overparameterization.} Figure \ref{fig:depth} shows that the power-law spectrum holds well in overparameterized models, but disappears in underparameterized models. Overparameterization is necessary for the power-law spectrum in deep learning, while proteins are also high-dimensional. It will be interesting to study the phase transition of overparameterization and underparameterization in future.

\begin{figure}
\begin{minipage}[t]{0.49\textwidth}
\centering
\subfigure{\includegraphics[width =0.49\columnwidth ]{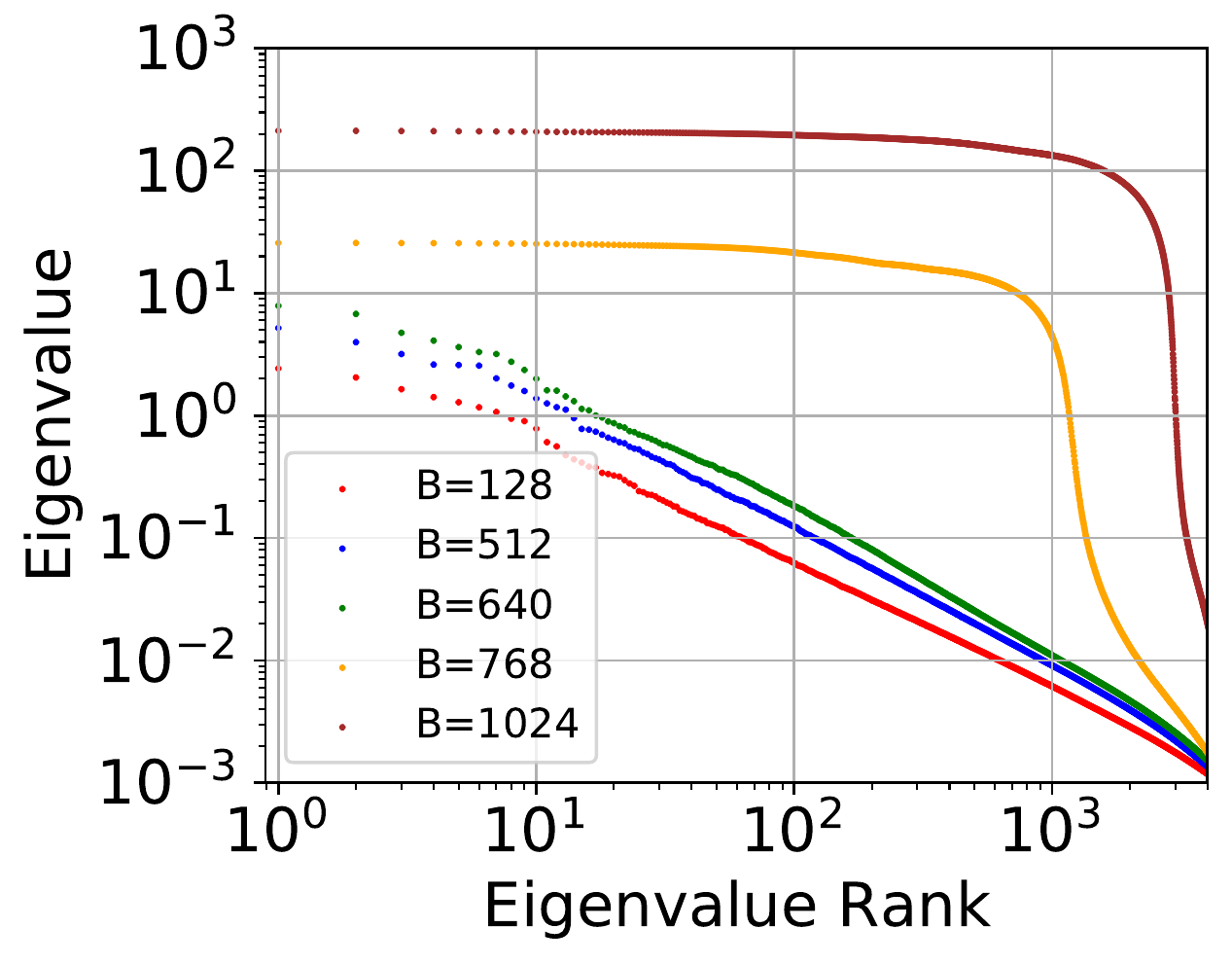}} 
\subfigure{\includegraphics[width =0.49\columnwidth ]{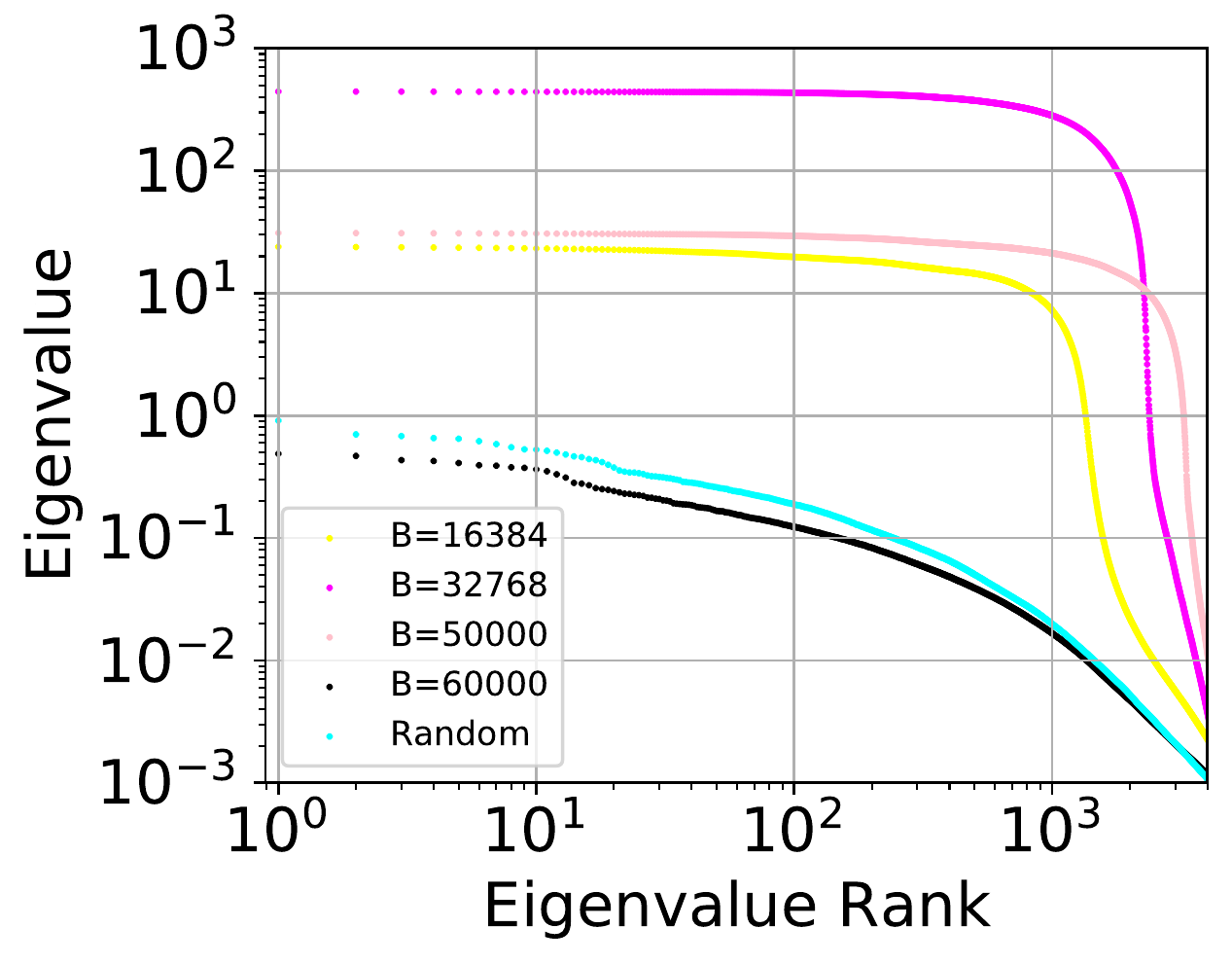}} 
\caption{Batch size matters to the spectrum. We discover three phases of the Hessian spectrums for large-batch training. Model:LeNet. Dataset: MNIST.}
 \label{fig:batchspectrum}
\end{minipage} 
\hfill
\begin{minipage}[t]{0.49\textwidth}
\centering
\subfigure{\includegraphics[width =0.49\columnwidth ]{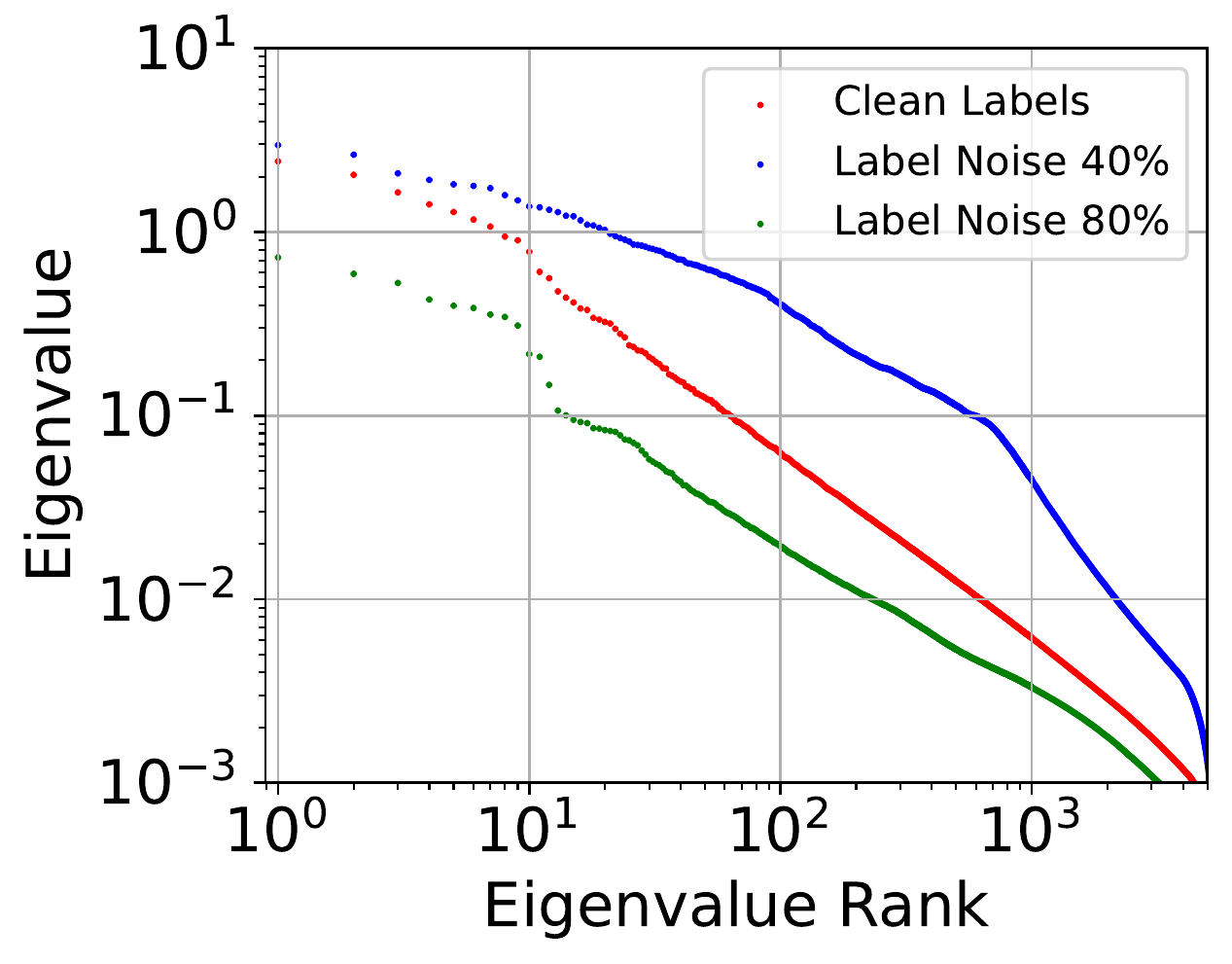}} 
\subfigure{\includegraphics[width =0.49\columnwidth ]{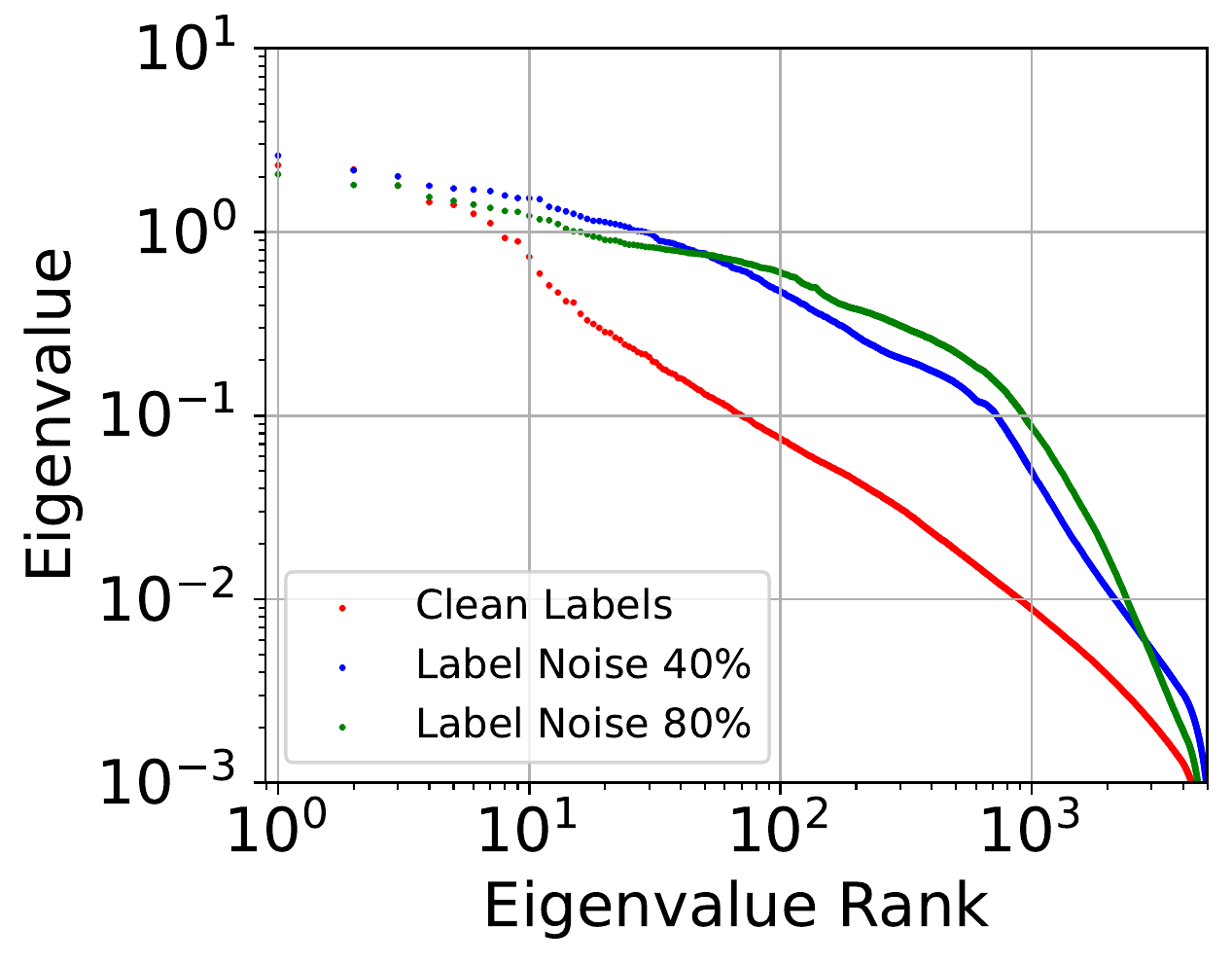}} 
\caption{The spectrum in the presence of noisy labels. Left: MNIST Trainset. Right: MNIST Testset.}
 \label{fig:noisyspectrum}
 \end{minipage} 
\end{figure}

 \begin{table*}[t]
\caption{The Kolmogorov-Smirnov statistics of the Hessian spectrums for various batch sizes.}
\label{table:kslenet-batch}
\begin{center}
\begin{small}
\resizebox{\textwidth}{!}{%
\begin{tabular}{llll | lllll}
\toprule
Dataset & Model & Training & Batch Size  & $d_{\rm{ks}}$ & $d_{\rm{c}}$ & Power-Law & $\hat{\beta} \pm \sigma$ & $\hat{s}$ \\
\midrule
MNIST & LeNet & SGD  & $B=128$  & 0.00900 & 0.0430 &  \textcolor{blue}{Yes}& $1.991 \pm 0.031$ & 1.009 \\  
MNIST & LeNet & SGD  & $B=512$  & 0.00787 & 0.0430 &  \textcolor{blue}{Yes}& $1.894 \pm 0.028$ & 1.119 \\  
MNIST & LeNet & SGD  & $B=640$  & 0.0125 & 0.0430 &  \textcolor{blue}{Yes}& $1.838 \pm 0.027$ & 1.194 \\ 
MNIST & LeNet & SGD  & $\textcolor{red}{B=768}$  & 0.278 & 0.0430 &  \textcolor{red}{No}  \\
MNIST & LeNet & SGD  & $\textcolor{red}{B=1024}$  & 0.129 & 0.0430 &  \textcolor{red}{No}  \\  
MNIST & LeNet & SGD & $\textcolor{red}{B=16384}$  & 0.249 & 0.0430 &  \textcolor{red}{No}  \\ 
MNIST & LeNet & SGD & $\textcolor{red}{B=32768}$  & 0.201 & 0.0430 &  \textcolor{red}{No}  \\ 
MNIST & LeNet & SGD & $\textcolor{red}{B=50000}$  & 0.139 & 0.0430 &  \textcolor{red}{No}  \\ 
MNIST & LeNet & SGD & $\textcolor{red}{B=60000}$  & 0.0936 & 0.0430 &  \textcolor{red}{No}  \\ 
\bottomrule
\end{tabular}
}
\end{small}
\end{center}
\end{table*}

\textbf{3. The Size of Training Dataset.} We evaluate the Hessian spectrums of DNNs trained over various training data sizes and evaluated on the same test dataset in Figure \ref{fig:numtraining}. The model trained with very limited training data finds minima with the Hessian spectrum that have multiple sharp directions, similarly to underparameterized models. 

\textbf{4. Batch Size.} We discover the three different phases for large-batch training via the curves in Figure \ref{fig:batchspectrum} and the KS test results in Table \ref{table:kslenet-batch}. 

First, in Phase I ($B\leq640$), moderately large-batch ($B=512$) training indeed finds sharper minima than small-batch ($B=128$) training, while the power-law spectrum still holds well. Power laws may guarantee that the top eigenvalues of large-batch trained networks are all larger than the corresponding eigenvalues of small-batch trained networks. The main challenge of large-batch training in Phase I is consistent with the common belief that large-batch training suffers from sharp minima and, thus, leads to bad generalization \citep{hoffer2017train,keskar2017large}. The minima sharpness measured by $\hat{s}$ obviously increases with the batch size.

Second, in Phase II ($768\leq B\leq50000$), the spectrum of large-batch ($B=1024$) trained networks does not exhibit power laws but is visually similar to the spectrum of underparameterized models in Figure \ref{fig:depth}. In Phase II, large-batch trained overparameterized models behave like underparameterized models from a spectral perspective, and, thus, can lead to bad generalization. The phase transition from Phase I to Phase II occurs in a narrow range of $640<B<768$, which is visually observable in Figure \ref{fig:batchspectrum}a and statistically observable in Table \ref{table:kslenet-batch}.

Third, in Phase III ($B \sim 60000$), extremely large-batch training ($B=60000$) cannot optimize the training loss well or find the Hessian spectrum similarly to random initialized neural networks. Phase III indicates that, sometimes, bad convergence rather than sharp minima can become the main performance bottleneck in large-batch training \citep{xie2020stable}, when the batch size is too large. 

To the best of our knowledge, we are the first to report Phase II, while Phase III was theoretically predicted by \citet{xie2020stable} but lacked direct empirical evidence before.

\textbf{5. Overfitting and Noisy Labels.} As DNNs overfit noisy labels easily, previous papers choose learning with noisy labels \citep{han2020survey} as an important setting for evaluating overfitting and generalization. Figure \ref{fig:noisyspectrum} shows that overfitting label noise makes the Hessian spectrums less power-law on both the corrupted training dataset and the clean test dataset. In contrast, in the absence of noisy labels, the power-law spectrums exist on both the training dataset and the test dataset. 

\textbf{6. Supplementary Results.} In Appendix \ref{sec:supresults}, we further discussed various interesting empirical results, including ResNet18, Random Labels, and Avila (a non-image dataset).

\section{A Spectral Parallel between Proteins and DNNs}

In this section, we make a pioneering discussion on a potential bridge between proteins and DNNs from a spectral perspective. This may be attract more attention to this topic from researchers with interdisciplinary backgrounds (e.g. protein science, biophysics, etc.).

As the basic building blocks of biological intelligence, proteins work as the main executors of various vital functions, including catalysis (enzyme), transportation (carrier proteins), defense (antibodies), and so on. The polypeptide chain made up of amino acid residues can fold into its native (energy-minimum) three-dimensional structure from a random coil, which is comparable with training of DNNs.
It is worth noting that, in a long timescale, the native structure (network parameter) $\vec{r}^0$ has been gradually shaped by evolution. Due to random mutations, the native structure (energy-minimum point) and corresponding elastic network have varied. Thus, protein evolution can be recognized as an optimization process of the network parameters \citep{tang2021dynamics}. With the second-order Taylor approximation near a minimum, for a given native structure $\vec{r}^0$, the potential energy can be calculated as:
\begin{equation}
E(\vec{r}^0) = E(\vec{r}^{\star}) + \frac{1}{2} (\vec{r}^0 - \vec{r}^{\star})^{\top} H(\vec{r}^{\star}) (\vec{r}^0 - \vec{r}^{\star}),
\label{eqn:protein}
\end{equation}
in which $\vec{r}^{\star}$ denote the reference structure, a selected ancestral structure in the evolution, and $H(\vec{r}^{\star})$ is its corresponding Hessian. In this way, the evolution of a protein become comparable to the training of artificial neural networks. 

We use the elastic network model (ENM) of the proteins \citep{atilgan2001anisotropy, bahar2010global} to calculate the vibrational spectrums of 9166 kinds of protein molecules from the Protein Data Bank (PDB) \citep{PDB}. To our knowledge, this is the most large-scale spectral study via ENM for proteins. We take a human protein assembly (shown in Figure \ref{fig:protein}a) as an example to obtain the vibration spectrums. The result is shown in Figure \ref{fig:protein}b.

\begin{figure}
\center
\includegraphics[width = 0.66\columnwidth]{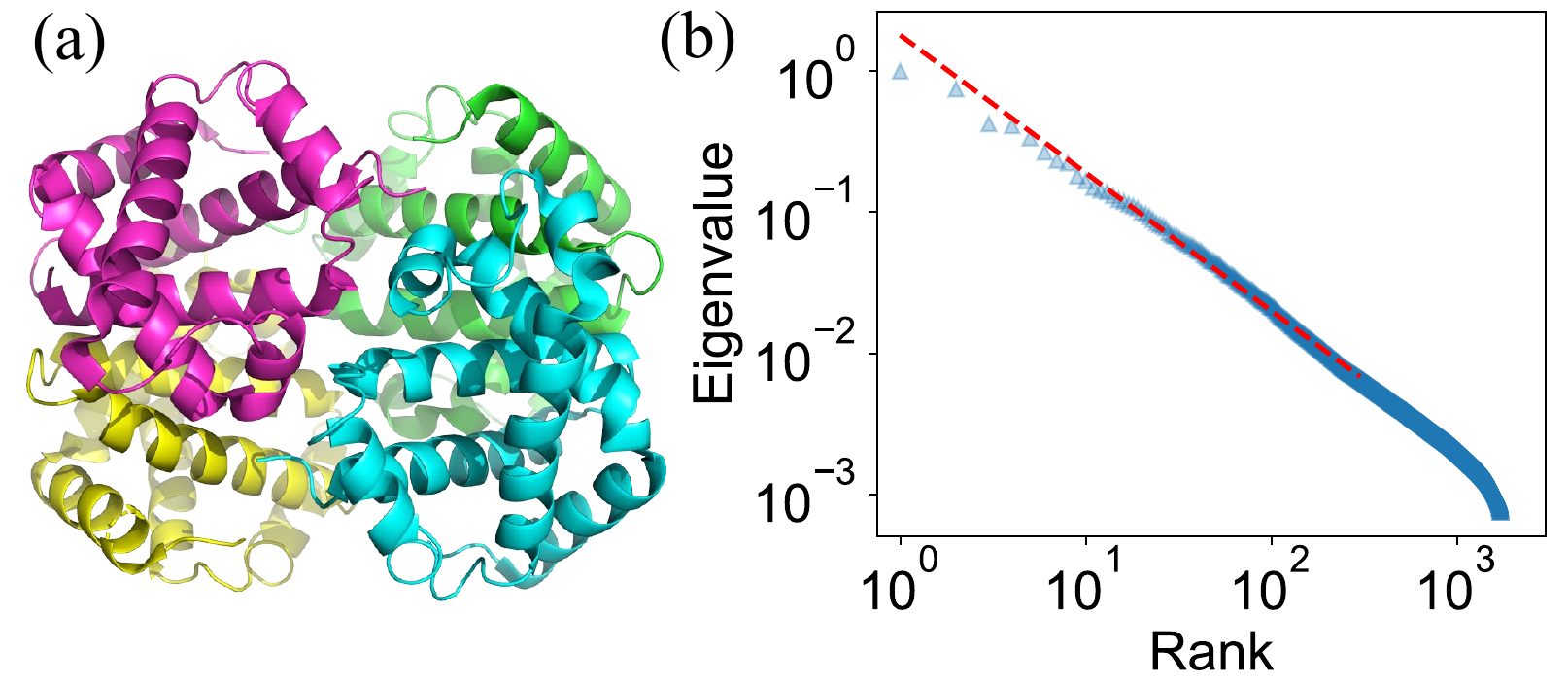}
\caption{(a) The cartoon illustration of human deoxyhemoglobin tetramer (PDB code: 4HHB). (b) The vibrational spectrum of the protein's elastic network. The estimated power exponent $\hat{s} = 0.992$ ($\hat{\beta} = 2.008 \pm 0.032$) is very close to the estimated power exponent in deep learning. }
\label{fig:protein}
\end{figure}

\begin{table}[t]
\caption{The estimated $\hat{s}$ for various sized proteins.}
\label{table:ksproteinsize}
\begin{center}
\begin{small}
\begin{sc}
\resizebox{0.6\textwidth}{!}{%
\begin{tabular}{l | lll}
\toprule
  Slope/SIze & $300 \leq 3N_{\rm{AA}} \leq 1000$  & $1000 \leq 3N_{\rm{AA}} \leq 3000$ & $3000 \leq 3N_{\rm{AA}} \leq 6000$   \\
  \midrule
$\hat{s}$  & $1.050 \pm 0.175$  &  $1.041 \pm 0.119$  &  $1.002 \pm 0.084$ \\
\bottomrule
\end{tabular}
}
\end{sc}
\end{small}
\end{center}
\end{table}

Surprisingly, we observed $\hat{s} \approx 1$ for the protein in Figure \ref{fig:protein} and in the spectrums of total 9166 kinds of protein molecules. Each of the protein molecule has $100\leq N_{\rm{AA}}\leq2000$ amino acid residues. All of the protein spectrums are statistically supported by the power-law KS tests with the estimated $mean(\hat{s})= 1.045$ ($mean(\hat{\beta})= 1.975$). The approximation $\hat{s} \approx 1$ also holds better for larger models (namely, proteins) in terms of the mean and the standard error, shown in Table \ref{table:ksproteinsize}. 

A parallel maximum-entropy theory has been proposed very recently for understanding the native structure and evolution of proteins in the statistical physical community \citep{tang2020functional}. Given the essential importance of protein science, our contribution to the similar power-law behaviors and theoretical interpretation may suggest a novel bridge between protein evolution and training of DNNs. In Appendix \ref{sec:protein} and \ref{sec:enmprotein}, we leave more formal discussion on the spectral similarity of protein and deep learning and present more technical details.

\section{Conclusion}

In this paper, we report the power-law spectrum in deep learning. Inspired by statistical physical theory \citep{visser2013zipf} and protein theory \citep{tang2020functional}, we successfully formulated a novel maximum-entropy interpretation and explain why the learning space may be lower dimensional and robust. The power-law spectrums provide us with a powerful tool to understand and analyze deep learning. We empirically demonstrate multiple novel behaviors of deep learning, particularly deep loss landscape, beyond previous studies. Particularly, those DNNs that do not exhibit power-law decaying Hessian eigenvalues after training usually have a large number of similarly large Hessian eigenvalues and cannot generalize well. Moreover, our theoretical interpretation and large-scale empirical study on proteins suggest a likely spectral bridge to deep learning. We believe our work will inspire more theories and empirical advancements on deep learning via power-law spectral analysis in the future.

\section*{Acknowledgement} 
We gratefully thank Profs. Zhanxing Zhu, Kunihiko Kaneko and Taro Toyoizumi for the helpful discussions. Q.-Y. T. thanks the support from the Brain/MINDS from AMED under Grant no. JP21dm0207001.

\bibliography{deeplearning}

\begin{thebibliography}{}

\bibitem[Alstott et~al., 2014]{alstott2014powerlaw}
Alstott, J., Bullmore, E., and Plenz, D. (2014).
\newblock powerlaw: a python package for analysis of heavy-tailed
  distributions.
\newblock {\em PloS one}, 9(1):e85777.

\bibitem[Atilgan et~al., 2001]{atilgan2001anisotropy}
Atilgan, A.~R., Durell, S., Jernigan, R.~L., Demirel, M.~C., Keskin, O., and
  Bahar, I. (2001).
\newblock Anisotropy of fluctuation dynamics of proteins with an elastic
  network model.
\newblock {\em Biophysical journal}, 80(1):505--515.

\bibitem[Bahar et~al., 2010]{bahar2010global}
Bahar, I., Lezon, T.~R., Yang, L.-W., and Eyal, E. (2010).
\newblock Global dynamics of proteins: bridging between structure and function.
\newblock {\em Annual review of biophysics}, 39:23--42.

\bibitem[Bahri et~al., 2020]{bahri2020statistical}
Bahri, Y., Kadmon, J., Pennington, J., Schoenholz, S.~S., Sohl-Dickstein, J.,
  and Ganguli, S. (2020).
\newblock Statistical mechanics of deep learning.
\newblock {\em Annual Review of Condensed Matter Physics}, 11(1).

\bibitem[Barron et~al., 1998]{barron1998minimum}
Barron, A., Rissanen, J., and Yu, B. (1998).
\newblock The minimum description length principle in coding and modeling.
\newblock {\em IEEE transactions on information theory}, 44(6):2743--2760.

\bibitem[Beal, 2003]{beal2003variational}
Beal, M.~J. (2003).
\newblock {\em Variational algorithms for approximate Bayesian inference}.
\newblock University of London, University College London (United Kingdom).

\bibitem[Berman et~al., 2000]{PDB}
Berman, H.~M., Westbrook, J., Feng, Z., Gilliland, G., Bhat, T.~N., Weissig,
  H., Shindyalov, I.~N., and Bourne, P.~E. (2000).
\newblock The protein data bank.
\newblock {\em Nucleic acids research}, 28(1):235--242.

\bibitem[Blier and Ollivier, 2018]{blier2018description}
Blier, L. and Ollivier, Y. (2018).
\newblock The description length of deep learning models.
\newblock In {\em Proceedings of the 32nd International Conference on Neural
  Information Processing Systems}, pages 2220--2230.

\bibitem[Byrd et~al., 2011]{byrd2011use}
Byrd, R.~H., Chin, G.~M., Neveitt, W., and Nocedal, J. (2011).
\newblock On the use of stochastic hessian information in optimization methods
  for machine learning.
\newblock {\em SIAM Journal on Optimization}, 21(3):977--995.

\bibitem[Clauset et~al., 2009]{clauset2009power}
Clauset, A., Shalizi, C.~R., and Newman, M.~E. (2009).
\newblock Power-law distributions in empirical data.
\newblock {\em SIAM review}, 51(4):661--703.

\bibitem[Dauphin et~al., 2014]{dauphin2014identifying}
Dauphin, Y.~N., Pascanu, R., Gulcehre, C., Cho, K., Ganguli, S., and Bengio, Y.
  (2014).
\newblock Identifying and attacking the saddle point problem in
  high-dimensional non-convex optimization.
\newblock {\em Advances in Neural Information Processing Systems},
  27:2933--2941.

\bibitem[Davis and Kahan, 1970]{davis1970rotation}
Davis, C. and Kahan, W.~M. (1970).
\newblock The rotation of eigenvectors by a perturbation. iii.
\newblock {\em SIAM Journal on Numerical Analysis}, 7(1):1--46.

\bibitem[De~Stefano et~al., 2018]{de2018reliable}
De~Stefano, C., Maniaci, M., Fontanella, F., and di~Freca, A.~S. (2018).
\newblock Reliable writer identification in medieval manuscripts through page
  layout features: The “avila” bible case.
\newblock {\em Engineering Applications of Artificial Intelligence},
  72:99--110.

\bibitem[Dinh et~al., 2017]{dinh2017sharp}
Dinh, L., Pascanu, R., Bengio, S., and Bengio, Y. (2017).
\newblock Sharp minima can generalize for deep nets.
\newblock In {\em International Conference on Machine Learning}, pages
  1019--1028.

\bibitem[Dubey et~al., 2019]{dubey2019diffgrad}
Dubey, S.~R., Chakraborty, S., Roy, S.~K., Mukherjee, S., Singh, S.~K., and
  Chaudhuri, B.~B. (2019).
\newblock diffgrad: An optimization method for convolutional neural networks.
\newblock {\em IEEE transactions on neural networks and learning systems},
  31(11):4500--4511.

\bibitem[Fort and Ganguli, 2019]{fort2019emergent}
Fort, S. and Ganguli, S. (2019).
\newblock Emergent properties of the local geometry of neural loss landscapes.
\newblock {\em arXiv preprint arXiv:1910.05929}.

\bibitem[Fort and Scherlis, 2019]{fort2019goldilocks}
Fort, S. and Scherlis, A. (2019).
\newblock The goldilocks zone: Towards better understanding of neural network
  loss landscapes.
\newblock In {\em Proceedings of the AAAI Conference on Artificial
  Intelligence}, volume~33, pages 3574--3581.

\bibitem[Ghorbani et~al., 2019]{ghorbani2019investigation}
Ghorbani, B., Krishnan, S., and Xiao, Y. (2019).
\newblock An investigation into neural net optimization via hessian eigenvalue
  density.
\newblock In {\em International Conference on Machine Learning}, pages
  2232--2241. PMLR.

\bibitem[Goldstein et~al., 2004]{goldstein2004problems}
Goldstein, M.~L., Morris, S.~A., and Yen, G.~G. (2004).
\newblock Problems with fitting to the power-law distribution.
\newblock {\em The European Physical Journal B-Condensed Matter and Complex
  Systems}, 41(2):255--258.

\bibitem[Graves, 2011]{graves2011practical}
Graves, A. (2011).
\newblock Practical variational inference for neural networks.
\newblock In {\em Advances in neural information processing systems}, pages
  2348--2356.

\bibitem[Gr{\"u}nwald, 2007]{grunwald2007minimum}
Gr{\"u}nwald, P.~D. (2007).
\newblock {\em The minimum description length principle}.
\newblock MIT press.

\bibitem[Guiasu and Shenitzer, 1985]{guiasu1985principle}
Guiasu, S. and Shenitzer, A. (1985).
\newblock The principle of maximum entropy.
\newblock {\em The mathematical intelligencer}, 7(1):42--48.

\bibitem[Guo et~al., 2004]{guo2004protein}
Guo, H.~H., Choe, J., and Loeb, L.~A. (2004).
\newblock Protein tolerance to random amino acid change.
\newblock {\em Proceedings of the National Academy of Sciences},
  101(25):9205--9210.

\bibitem[Gur-Ari et~al., 2018]{gur2018gradient}
Gur-Ari, G., Roberts, D.~A., and Dyer, E. (2018).
\newblock Gradient descent happens in a tiny subspace.
\newblock {\em arXiv preprint arXiv:1812.04754}.

\bibitem[Gurbuzbalaban et~al., 2021]{gurbuzbalaban2021heavy}
Gurbuzbalaban, M., Simsekli, U., and Zhu, L. (2021).
\newblock The heavy-tail phenomenon in sgd.
\newblock In {\em International Conference on Machine Learning}, pages
  3964--3975. PMLR.

\bibitem[Han et~al., 2020]{han2020survey}
Han, B., Yao, Q., Liu, T., Niu, G., Tsang, I.~W., Kwok, J.~T., and Sugiyama, M.
  (2020).
\newblock A survey of label-noise representation learning: Past, present and
  future.
\newblock {\em arXiv preprint arXiv:2011.04406}.

\bibitem[Han et~al., 2018]{han2018co}
Han, B., Yao, Q., Yu, X., Niu, G., Xu, M., Hu, W., Tsang, I., and Sugiyama, M.
  (2018).
\newblock Co-teaching: Robust training of deep neural networks with extremely
  noisy labels.
\newblock In {\em Advances in neural information processing systems}, pages
  8527--8537.

\bibitem[He et~al., 2019]{he2019control}
He, F., Liu, T., and Tao, D. (2019).
\newblock Control batch size and learning rate to generalize well: Theoretical
  and empirical evidence.
\newblock In {\em Advances in Neural Information Processing Systems}, pages
  1141--1150.

\bibitem[He et~al., 2016]{he2016deep}
He, K., Zhang, X., Ren, S., and Sun, J. (2016).
\newblock Deep residual learning for image recognition.
\newblock In {\em Proceedings of the IEEE conference on computer vision and
  pattern recognition}, pages 770--778.

\bibitem[Hinton and Camp, 1993]{hinton1993keeping}
Hinton, G.~E. and Camp, D.~V. (1993).
\newblock Keeping neural networks simple by minimising the description length
  of weights.
\newblock In {\em Proceedings of Conference on Learning Theory}, pages 5--13.

\bibitem[Hochreiter and Schmidhuber, 1995]{hochreiter1995simplifying}
Hochreiter, S. and Schmidhuber, J. (1995).
\newblock Simplifying neural nets by discovering flat minima.
\newblock In {\em Advances in neural information processing systems}, pages
  529--536.

\bibitem[Hochreiter and Schmidhuber, 1997]{hochreiter1997flat}
Hochreiter, S. and Schmidhuber, J. (1997).
\newblock Flat minima.
\newblock {\em Neural Computation}, 9(1):1--42.

\bibitem[Hodgkinson and Mahoney, 2021]{hodgkinson2021multiplicative}
Hodgkinson, L. and Mahoney, M. (2021).
\newblock Multiplicative noise and heavy tails in stochastic optimization.
\newblock In {\em International Conference on Machine Learning}, pages
  4262--4274. PMLR.

\bibitem[Hoffer et~al., 2017]{hoffer2017train}
Hoffer, E., Hubara, I., and Soudry, D. (2017).
\newblock Train longer, generalize better: closing the generalization gap in
  large batch training of neural networks.
\newblock In {\em Advances in Neural Information Processing Systems}, pages
  1729--1739.

\bibitem[Honkela and Valpola, 2004]{honkela2004variational}
Honkela, A. and Valpola, H. (2004).
\newblock Variational learning and bits-back coding: an information-theoretic
  view to bayesian learning.
\newblock {\em IEEE transactions on Neural Networks}, 15(4):800--810.

\bibitem[Jacot et~al., 2019]{jacot2019asymptotic}
Jacot, A., Gabriel, F., and Hongler, C. (2019).
\newblock The asymptotic spectrum of the hessian of dnn throughout training.
\newblock In {\em International Conference on Learning Representations}.

\bibitem[Jastrzkebski et~al., 2017]{jastrzkebski2017three}
Jastrzkebski, S., Kenton, Z., Arpit, D., Ballas, N., Fischer, A., Bengio, Y.,
  and Storkey, A. (2017).
\newblock Three factors influencing minima in sgd.
\newblock {\em arXiv preprint arXiv:1711.04623}.

\bibitem[Jiang et~al., 2019]{jiang2019fantastic}
Jiang, Y., Neyshabur, B., Mobahi, H., Krishnan, D., and Bengio, S. (2019).
\newblock Fantastic generalization measures and where to find them.
\newblock In {\em International Conference on Learning Representations}.

\bibitem[Keskar et~al., 2017]{keskar2017large}
Keskar, N.~S., Mudigere, D., Nocedal, J., Smelyanskiy, M., and Tang, P. T.~P.
  (2017).
\newblock On large-batch training for deep learning: Generalization gap and
  sharp minima.
\newblock In {\em International Conference on Learning Representations}.

\bibitem[Kingma and Ba, 2015]{kingma2014adam}
Kingma, D.~P. and Ba, J. (2015).
\newblock Adam: A method for stochastic optimization.
\newblock {\em 3rd International Conference on Learning Representations, ICLR
  2015}.

\bibitem[Krizhevsky and Hinton, 2009]{krizhevsky2009learning}
Krizhevsky, A. and Hinton, G. (2009).
\newblock Learning multiple layers of features from tiny images.

\bibitem[LeCun, 1998]{lecun1998mnist}
LeCun, Y. (1998).
\newblock The mnist database of handwritten digits.
\newblock {\em http://yann. lecun. com/exdb/mnist/}.

\bibitem[LeCun et~al., 1998]{lecun1998gradient}
LeCun, Y., Bottou, L., Bengio, Y., and Haffner, P. (1998).
\newblock Gradient-based learning applied to document recognition.
\newblock {\em Proceedings of the IEEE}, 86(11):2278--2324.

\bibitem[Li et~al., 2020]{li2020hessian}
Li, X., Gu, Q., Zhou, Y., Chen, T., and Banerjee, A. (2020).
\newblock Hessian based analysis of sgd for deep nets: Dynamics and
  generalization.
\newblock In {\em Proceedings of the 2020 SIAM International Conference on Data
  Mining}, pages 190--198. SIAM.

\bibitem[Li et~al., 2021]{li2021on}
Li, Z., Malladi, S., and Arora, S. (2021).
\newblock On the validity of modeling {SGD} with stochastic differential
  equations ({SDE}s).
\newblock In {\em Thirty-Fifth Conference on Neural Information Processing
  Systems}.

\bibitem[Liao and Mahoney, 2021]{liao2021hessian}
Liao, Z. and Mahoney, M.~W. (2021).
\newblock Hessian eigenspectra of more realistic nonlinear models.
\newblock {\em Advances in Neural Information Processing Systems}, 34.

\bibitem[Liu et~al., 2019]{liu2019variance}
Liu, L., Jiang, H., He, P., Chen, W., Liu, X., Gao, J., and Han, J. (2019).
\newblock On the variance of the adaptive learning rate and beyond.
\newblock In {\em International Conference on Learning Representations}.

\bibitem[Luo et~al., 2019]{luo2019adaptive}
Luo, L., Xiong, Y., Liu, Y., and Sun, X. (2019).
\newblock Adaptive gradient methods with dynamic bound of learning rate.
\newblock {\em 7th International Conference on Learning Representations, ICLR
  2019}.

\bibitem[Mandt et~al., 2017]{mandt2017stochastic}
Mandt, S., Hoffman, M.~D., and Blei, D.~M. (2017).
\newblock Stochastic gradient descent as approximate bayesian inference.
\newblock {\em The Journal of Machine Learning Research}, 18(1):4873--4907.

\bibitem[Massey~Jr, 1951]{massey1951kolmogorov}
Massey~Jr, F.~J. (1951).
\newblock The kolmogorov-smirnov test for goodness of fit.
\newblock {\em Journal of the American statistical Association},
  46(253):68--78.

\bibitem[Meurant and Strako{\v{s}}, 2006]{meurant2006lanczos}
Meurant, G. and Strako{\v{s}}, Z. (2006).
\newblock The lanczos and conjugate gradient algorithms in finite precision
  arithmetic.
\newblock {\em Acta Numerica}, 15:471--542.

\bibitem[Munoz, 2018]{munoz2018colloquium}
Munoz, M.~A. (2018).
\newblock Colloquium: Criticality and dynamical scaling in living systems.
\newblock {\em Reviews of Modern Physics}, 90(3):031001.

\bibitem[Myung, 2003]{myung2003tutorial}
Myung, I.~J. (2003).
\newblock Tutorial on maximum likelihood estimation.
\newblock {\em Journal of mathematical Psychology}, 47(1):90--100.

\bibitem[Neyshabur et~al., 2017]{neyshabur2017exploring}
Neyshabur, B., Bhojanapalli, S., McAllester, D., and Srebro, N. (2017).
\newblock Exploring generalization in deep learning.
\newblock In {\em Advances in Neural Information Processing Systems}, pages
  5947--5956.

\bibitem[Panigrahi et~al., 2019]{panigrahi2019non}
Panigrahi, A., Somani, R., Goyal, N., and Netrapalli, P. (2019).
\newblock Non-gaussianity of stochastic gradient noise.
\newblock {\em arXiv preprint arXiv:1910.09626}.

\bibitem[Papyan, 2019]{papyan2019measurements}
Papyan, V. (2019).
\newblock Measurements of three-level hierarchical structure in the outliers in
  the spectrum of deepnet hessians.
\newblock In {\em International Conference on Machine Learning}, pages
  5012--5021. PMLR.

\bibitem[Pennington and Bahri, 2017]{pennington2017geometry}
Pennington, J. and Bahri, Y. (2017).
\newblock Geometry of neural network loss surfaces via random matrix theory.
\newblock In {\em International Conference on Machine Learning}, pages
  2798--2806. PMLR.

\bibitem[Pennington and Worah, 2018]{pennington2018spectrum}
Pennington, J. and Worah, P. (2018).
\newblock The spectrum of the fisher information matrix of a
  single-hidden-layer neural network.
\newblock {\em Advances in neural information processing systems}, 31.

\bibitem[Reddi et~al., 2019]{reddi2019convergence}
Reddi, S.~J., Kale, S., and Kumar, S. (2019).
\newblock On the convergence of adam and beyond.
\newblock {\em 6th International Conference on Learning Representations, ICLR
  2018}.

\bibitem[Reuveni et~al., 2008]{reuveni2008proteins}
Reuveni, S., Granek, R., and Klafter, J. (2008).
\newblock Proteins: coexistence of stability and flexibility.
\newblock {\em Physical review letters}, 100(20):208101.

\bibitem[Rissanen, 2007]{rissanen2007information}
Rissanen, J. (2007).
\newblock {\em Information and complexity in statistical modeling}.
\newblock Springer Science \& Business Media.

\bibitem[Sagun et~al., 2016]{sagun2016eigenvalues}
Sagun, L., Bottou, L., and LeCun, Y. (2016).
\newblock Eigenvalues of the hessian in deep learning: Singularity and beyond.
\newblock {\em arXiv preprint arXiv:1611.07476}.

\bibitem[Sagun et~al., 2017]{sagun2017empirical}
Sagun, L., Evci, U., Guney, V.~U., Dauphin, Y., and Bottou, L. (2017).
\newblock Empirical analysis of the hessian of over-parametrized neural
  networks.
\newblock {\em arXiv preprint arXiv:1706.04454}.

\bibitem[Sakata and Kaneko, 2020]{sakata2020dimensional}
Sakata, A. and Kaneko, K. (2020).
\newblock Dimensional reduction in evolving spin-glass model: correlation of
  phenotypic responses to environmental and mutational changes.
\newblock {\em Physical Review Letters}, 124(21):218101.

\bibitem[Sankar et~al., 2021]{sankar2021deeper}
Sankar, A.~R., Khasbage, Y., Vigneswaran, R., and Balasubramanian, V.~N.
  (2021).
\newblock A deeper look at the hessian eigenspectrum of deep neural networks
  and its applications to regularization.
\newblock In {\em Proceedings of the AAAI Conference on Artificial
  Intelligence}, volume~35, pages 9481--9488.

\bibitem[Sato and Kaneko, 2020]{sato2020evolutionary}
Sato, T.~U. and Kaneko, K. (2020).
\newblock Evolutionary dimension reduction in phenotypic space.
\newblock {\em Physical Review Research}, 2(1):013197.

\bibitem[Simsekli et~al., 2019]{simsekli2019tail}
Simsekli, U., Sagun, L., and Gurbuzbalaban, M. (2019).
\newblock A tail-index analysis of stochastic gradient noise in deep neural
  networks.
\newblock In {\em International Conference on Machine Learning}, pages
  5827--5837.

\bibitem[Stringer et~al., 2019]{stringer2019high}
Stringer, C., Pachitariu, M., Steinmetz, N., Carandini, M., and Harris, K.~D.
  (2019).
\newblock High-dimensional geometry of population responses in visual cortex.
\newblock {\em Nature}, 571(7765):361--365.

\bibitem[Tang et~al., 2020]{tang2020functional}
Tang, Q.-Y., Hatakeyama, T.~S., and Kaneko, K. (2020).
\newblock Functional sensitivity and mutational robustness of proteins.
\newblock {\em Physical Review Research}, 2(3):033452.

\bibitem[Tang and Kaneko, 2020]{tang2020long}
Tang, Q.-Y. and Kaneko, K. (2020).
\newblock Long-range correlation in protein dynamics: Confirmation by
  structural data and normal mode analysis.
\newblock {\em PLoS computational biology}, 16(2):e1007670.

\bibitem[Tang and Kaneko, 2021]{tang2021dynamics}
Tang, Q.-Y. and Kaneko, K. (2021).
\newblock Dynamics-evolution correspondence in protein structures.
\newblock {\em Physical Review Letters}, 127(9):098103.

\bibitem[Tang et~al., 2017]{tang2017critical}
Tang, Q.-Y., Zhang, Y.-Y., Wang, J., Wang, W., and Chialvo, D.~R. (2017).
\newblock Critical fluctuations in the native state of proteins.
\newblock {\em Physical review letters}, 118(8):088102.

\bibitem[Teh et~al., 2003]{teh2003energy}
Teh, Y.~W., Welling, M., Osindero, S., and Hinton, G.~E. (2003).
\newblock Energy-based models for sparse overcomplete representations.
\newblock {\em Journal of Machine Learning Research}, 4(Dec):1235--1260.

\bibitem[Thomas et~al., 2020]{thomas2020interplay}
Thomas, V., Pedregosa, F., Merri{\"e}nboer, B., Manzagol, P.-A., Bengio, Y.,
  and Le~Roux, N. (2020).
\newblock On the interplay between noise and curvature and its effect on
  optimization and generalization.
\newblock In {\em International Conference on Artificial Intelligence and
  Statistics}, pages 3503--3513. PMLR.

\bibitem[Torlai and Melko, 2016]{torlai2016learning}
Torlai, G. and Melko, R.~G. (2016).
\newblock Learning thermodynamics with boltzmann machines.
\newblock {\em Physical Review B}, 94(16):165134.

\bibitem[Tsuzuku et~al., 2020]{tsuzuku2020normalized}
Tsuzuku, Y., Sato, I., and Sugiyama, M. (2020).
\newblock Normalized flat minima: Exploring scale invariant definition of flat
  minima for neural networks using pac-bayesian analysis.
\newblock In {\em International Conference on Machine Learning}, pages
  9636--9647. PMLR.

\bibitem[Visser, 2013]{visser2013zipf}
Visser, M. (2013).
\newblock Zipf's law, power laws and maximum entropy.
\newblock {\em New Journal of Physics}, 15(4):043021.

\bibitem[Wenzel et~al., 2020]{wenzel2020good}
Wenzel, F., Roth, K., Veeling, B., Swiatkowski, J., Tran, L., Mandt, S., Snoek,
  J., Salimans, T., Jenatton, R., and Nowozin, S. (2020).
\newblock How good is the bayes posterior in deep neural networks really?
\newblock In {\em International Conference on Machine Learning}, pages
  10248--10259. PMLR.

\bibitem[Wu et~al., 2017]{wu2017towards}
Wu, L., Zhu, Z., et~al. (2017).
\newblock Towards understanding generalization of deep learning: Perspective of
  loss landscapes.
\newblock {\em arXiv preprint arXiv:1706.10239}.

\bibitem[Xiao et~al., 2017]{xiao2017fashion}
Xiao, H., Rasul, K., and Vollgraf, R. (2017).
\newblock Fashion-mnist: a novel image dataset for benchmarking machine
  learning algorithms.
\newblock {\em arXiv preprint arXiv:1708.07747}.

\bibitem[Xie et~al., 2020]{xie2020stable}
Xie, Z., Sato, I., and Sugiyama, M. (2020).
\newblock Stable weight decay regularization.
\newblock {\em arXiv preprint arXiv:2011.11152}.

\bibitem[Xie et~al., 2021a]{xie2021diffusion}
Xie, Z., Sato, I., and Sugiyama, M. (2021a).
\newblock A diffusion theory for deep learning dynamics: Stochastic gradient
  descent exponentially favors flat minima.
\newblock In {\em International Conference on Learning Representations}.

\bibitem[Xie et~al., 2022]{xie2022adaptive}
Xie, Z., Wang, X., Zhang, H., Sato, I., and Sugiyama, M. (2022).
\newblock Adaptive inertia: Disentangling the effects of adaptive learning rate
  and momentum.
\newblock In {\em Proceedings of the 39th International Conference on Machine
  Learning}, pages 24430--24459.

\bibitem[Xie et~al., 2021b]{xie2021positive}
Xie, Z., Yuan, L., Zhu, Z., and Sugiyama, M. (2021b).
\newblock Positive-negative momentum: Manipulating stochastic gradient noise to
  improve generalization.
\newblock In {\em International Conference on Machine Learning}, volume 139 of
  {\em Proceedings of Machine Learning Research}, pages 11448--11458. PMLR.

\bibitem[Yao et~al., 2020]{yao2020pyhessian}
Yao, Z., Gholami, A., Keutzer, K., and Mahoney, M.~W. (2020).
\newblock Pyhessian: Neural networks through the lens of the hessian.
\newblock In {\em 2020 IEEE International Conference on Big Data (Big Data)},
  pages 581--590. IEEE.

\bibitem[Yao et~al., 2018]{yao2018hessian}
Yao, Z., Gholami, A., Lei, Q., Keutzer, K., and Mahoney, M.~W. (2018).
\newblock Hessian-based analysis of large batch training and robustness to
  adversaries.
\newblock In {\em Advances in Neural Information Processing Systems}, pages
  4949--4959.

\bibitem[Yu et~al., 2015]{yu2015useful}
Yu, Y., Wang, T., and Samworth, R.~J. (2015).
\newblock A useful variant of the davis--kahan theorem for statisticians.
\newblock {\em Biometrika}, 102(2):315--323.

\bibitem[Zaheer et~al., 2018]{zaheer2018adaptive}
Zaheer, M., Reddi, S., Sachan, D., Kale, S., and Kumar, S. (2018).
\newblock Adaptive methods for nonconvex optimization.
\newblock In {\em Advances in neural information processing systems}, pages
  9793--9803.

\bibitem[Zhang et~al., 2019a]{zhang2019algorithmic}
Zhang, G., Li, L., Nado, Z., Martens, J., Sachdeva, S., Dahl, G., Shallue, C.,
  and Grosse, R.~B. (2019a).
\newblock Which algorithmic choices matter at which batch sizes? insights from
  a noisy quadratic model.
\newblock In {\em Advances in Neural Information Processing Systems}, pages
  8196--8207.

\bibitem[Zhang et~al., 2019b]{zhang2019lookahead}
Zhang, M., Lucas, J., Ba, J., and Hinton, G.~E. (2019b).
\newblock Lookahead optimizer: k steps forward, 1 step back.
\newblock {\em Advances in Neural Information Processing Systems},
  32:9597--9608.

\bibitem[Zhao et~al., 2019]{zhao2019bridging}
Zhao, P., Chen, P.-Y., Das, P., Ramamurthy, K.~N., and Lin, X. (2019).
\newblock Bridging mode connectivity in loss landscapes and adversarial
  robustness.
\newblock In {\em International Conference on Learning Representations}.

\bibitem[Zhu et~al., 2019]{zhu2019anisotropic}
Zhu, Z., Wu, J., Yu, B., Wu, L., and Ma, J. (2019).
\newblock The anisotropic noise in stochastic gradient descent: Its behavior of
  escaping from sharp minima and regularization effects.
\newblock In {\em ICML}, pages 7654--7663.

\end{thebibliography}

\appendix

\section{Kolmogorov-Smirnov Goodness-of-Fit Test}
\label{sec:kstest}

In this section, we introduce how to conduct the Kolmogorov-Smirnov Goodness-of-Fit Test for the self-containedness purpose. 

As we mentioned above, our work used Maximum Likelihood Estimation (MLE) \citep{myung2003tutorial,clauset2009power} for estimating the parameter $\beta$ of the fitted power-law distributions and the Kolmogorov-Smirnov Test (KS Test) \citep{massey1951kolmogorov,goldstein2004problems} for statistically testing the goodness of the fit. The KS test statistic is the KS distance $d_{\rm{ks}}$ between the hypothesized (fitted) distribution and the empirical data, which measures the goodness of fit. Mathematically, the KS distance is defined as 
\begin{align}
\label{eq:ksdistance}
d_{\rm{ks}} = \sup_{\lambda} | F^{\star}(\lambda) - \hat{F}(\lambda) |,
\end{align}
where $F^{\star}(\lambda)$ is the hypothesized cumulative distribution function and $\hat{F}(\lambda)$ is the empirical cumulative distribution function based on the sampled data \citep{goldstein2004problems}. The estimated power exponent via MLE \citep{clauset2009power} can be written as
\begin{align}
\label{eq:mle}
\hat{\beta} = 1 + K \left[\sum_{i=1}^{K} \ln\left(\frac{\lambda_{i}}{\lambda_{\rm{cutoff}}}\right)\right]^{-1},
\end{align}
where $K$ is the number of tested samples and we set $\lambda_{\rm{cutoff}} = \lambda_{k}$.
We note that the Powerlaw library \citep{alstott2014powerlaw} provides a convenient tool to compute the KS distance, $d_{\rm{ks}}$, and estimate the power exponent.

According to the practice of KS Test \citep{massey1951kolmogorov}, we first state {\it \textbf{the power-law hypothesis}} that the tested spectrum is power-law. If $d_{\rm{ks}}$ is higher than the critical value $d_{\rm{c}}$ at the $\alpha=0.05$ significance level, the KS test statistically will support the power-law hypothesis (we cannot reject the power-law hypothesis). We display the critical values in Table \ref{table:kstest}. 

We conducted the KS tests for all of our studied spectrums. We display the KS test statistics and the estimated power exponents $\hat{\beta}$ with standard errors $\sigma$ as well as the corresponding $\hat{s}$ in Tables \ref{table:kslenet-mnist}, \ref{table:kslenet-cifar}, \ref{table:ksfcn}, \ref{table:ksresnet18}, \ref{table:kslnn},and \ref{table:ksprotein}. In the tables, we take the base hyperparameter setting in Appendix \ref{sec:experimentsetting} as the default setting. For better visualization, we color accepting the power-law hypothesis in blue and color rejecting the power-law hypothesis (and the cause) in red. 


 \begin{table*}[t]
\caption{The Table of Kolmogorov-Smirnov Test Critical Values (Significance Level), which was first reported in \citet{massey1951kolmogorov}. If the KS distance $d_{\rm{ks}}$ is lower than a critical value, such as $\frac{1.36}{\sqrt{K}}$ , we would reject the null hypothesis and accept the power-law hypothesis at the $\alpha=0.05$ significance level. Note that $K$ is the number of tested eigenvalues.}
\label{table:kstest}
\begin{center}
\begin{small}
\begin{tabular}{l | lllll}
\toprule
Sample size & $\alpha=0.2$  & $\alpha=0.15$ & $\alpha=0.1$ & $\alpha=0.05$ & $\alpha=0.01$  \\
\midrule
K > 35   &  $\frac{1.07}{\sqrt{k}}$    & $\frac{1.14}{\sqrt{k}}$  & $\frac{1.22}{\sqrt{k}}$  & $\frac{1.36}{\sqrt{k}}$  & $\frac{1.63}{\sqrt{k}}$  \\  
\midrule
K = 50   &  $0.151$    & $0.161$  & $0.173$  & $0.192$  & $0.231$  \\        
K = 1000   &  $0.0338$    & $0.0360$  & $0.0386$  & $0.0430$  & $0.0515$  \\                   
\bottomrule
\end{tabular}
\end{small}
\end{center}
\end{table*}

 \begin{table*}[t]
\caption{The Kolmogorov-Smirnov statistics of LeNet on MNIST and Fashion MNIST. The estimated power exponent $\hat{\beta}$ and slope magnitude $\hat{s}$ are also displayed.}
\label{table:kslenet-mnist}
\begin{center}
\begin{small}
\resizebox{\textwidth}{!}{%
\begin{tabular}{lllll | lllll}
\toprule
Dataset & Model & Training & Sample size & Setting  & $d_{\rm{ks}}$ & $d_{\rm{c}}$ & Power-Law & $\hat{\beta} \pm \sigma$ & $\hat{s}$ \\
\midrule
MNIST & LeNet & \textcolor{red}{Random} & 1000 & -  & 0.0796 & 0.0430 &  \textcolor{red}{No}  \\ 
MNIST & LeNet & SGD & 1000 & -   & 0.00900 & 0.0430 &  \textcolor{blue}{Yes}& $1.991 \pm 0.031$ & 1.009 \\  
MNIST & LeNet & Vanilla SGD & 1000 & -   & 0.0103 & 0.0430 &  \textcolor{blue}{Yes} & $1.914 \pm 0.029$ & 1.094 \\ 
MNIST & LeNet & Adam & 1000 & -  & 0.00962 & 0.0430 &  \textcolor{blue}{Yes} & $1.873 \pm 0.028$ & 1.145  \\ 
MNIST & LeNet & AMSGrad & 1000 & -  & 0.00987 & 0.0430 &  \textcolor{blue}{Yes} & $1.845 \pm 0.027$ & 1.184 \\ 
MNIST & LeNet & AdaBound & 1000 & -  & 0.00889 & 0.0430 &  \textcolor{blue}{Yes} & $1.904 \pm 0.029$ & 1.106 \\ 
MNIST & LeNet & Yogi & 1000 & -  & 0.00966 & 0.0430 &  \textcolor{blue}{Yes} & $1.834 \pm 0.026$ & 1.198 \\ 
MNIST & LeNet & RAdam & 1000 & -  & 0.0164 & 0.0430 &  \textcolor{blue}{Yes} & $1.889 \pm 0.028$ & 1.125 \\ 
MNIST & LeNet & Adai & 1000 & -   & 0.0101 & 0.0430 &  \textcolor{blue}{Yes} & $1.892 \pm 0.028$ & 1.122 \\ 
MNIST & LeNet & PNM & 1000 & -  & 0.0127 & 0.0430 &  \textcolor{blue}{Yes} & $1.846 \pm 0.027$ & 1.181 \\ 
MNIST & LeNet & Lookahead & 1000 & -  & 0.0101 & 0.0430 &  \textcolor{blue}{Yes} & $1.982 \pm 0.031$ & 1.018 \\ 
MNIST & LeNet & DiffGrad & 1000 & -  & 0.0105 & 0.0430 &  \textcolor{blue}{Yes} & $1.834 \pm 0.026$ & 1.198 \\ 
\midrule
MNIST & LeNet & SGD & 1000 & $B=128$  & 0.00900 & 0.0430 &  \textcolor{blue}{Yes}& $1.991 \pm 0.031$ & 1.009 \\  
MNIST & LeNet & SGD & 1000 & $B=512$  & 0.00787 & 0.0430 &  \textcolor{blue}{Yes}& $1.894 \pm 0.028$ & 1.119 \\  
MNIST & LeNet & SGD & 1000 & $B=640$  & 0.0125 & 0.0430 &  \textcolor{blue}{Yes}& $1.838 \pm 0.027$ & 1.194 \\ 
MNIST & LeNet & SGD & 1000 & $\textcolor{red}{B=768}$  & 0.278 & 0.0430 &  \textcolor{red}{No}  \\
MNIST & LeNet & SGD & 1000 & $\textcolor{red}{B=1024}$  & 0.129 & 0.0430 &  \textcolor{red}{No}  \\  
MNIST & LeNet & SGD & 1000 & $\textcolor{red}{B=8192}$  & 0.240 & 0.0430 &  \textcolor{red}{No}  \\ 
MNIST & LeNet & SGD & 1000 & $\textcolor{red}{B=16384}$  & 0.249 & 0.0430 &  \textcolor{red}{No}  \\ 
MNIST & LeNet & SGD & 1000 & $\textcolor{red}{B=32768}$  & 0.201 & 0.0430 &  \textcolor{red}{No}  \\ 
MNIST & LeNet & SGD & 1000 & $\textcolor{red}{B=50000}$  & 0.139 & 0.0430 &  \textcolor{red}{No}  \\ 
MNIST & LeNet & SGD & 1000 & $\textcolor{red}{B=60000}$  & 0.0936 & 0.0430 &  \textcolor{red}{No}  \\ 
\midrule
MNIST & LeNet & SGD & 1000 & $\textcolor{red}{N=600}$  & 0.205 & 0.0430 &  \textcolor{red}{No}  \\ 	
MNIST & LeNet & SGD & 1000 & $N=800$  & 0.0399 & 0.0430 &  \textcolor{blue}{Yes}& $1.995 \pm 0.031$ & 1.004 \\ 	
MNIST & LeNet & SGD & 1000 & $N=1000$  & 0.0198 & 0.0430 &  \textcolor{blue}{Yes}& $2.128 \pm 0.036$ & 0.886 \\ 	
MNIST & LeNet & SGD & 1000 & $N=3000$  & 0.0159 & 0.0430 &  \textcolor{blue}{Yes}& $2.091 \pm 0.034$ & 0.917 \\ 	
MNIST & LeNet & SGD & 1000 & $N=6000$  & 0.0151 & 0.0430 &  \textcolor{blue}{Yes}& $2.001 \pm 0.032$ & 0.999 \\ 	
\midrule
MNIST & LeNet &SGD  & 1000 &  \textcolor{red}{$40\%$ Label Noise}   & 0.180 & 0.0430 &  \textcolor{red}{No}& \\  
MNIST & LeNet & SGD & 1000  & \textcolor{red}{$80\%$ Label Noise}   & 0.157 & 0.0430 &  \textcolor{red}{No}& \\  
MNIST & LeNet & SGD & 1000 &  \textcolor{red}{Random Labels}   & 0.0482 & 0.0430 &  \textcolor{red}{No}& \\  
\midrule
Fashion-MNIST & LeNet & \textcolor{red}{Random}  & 1000 & -   & 0.0971 & 0.0430 &  \textcolor{red}{No}& \\  
Fashion-MNIST & LeNet & SGD & 1000 & -   & 0.0132 & 0.0430 &  \textcolor{blue}{Yes}& $1.939 \pm 0.030 $ & 1.065\\
\midrule
MNIST & LeNet & SGD & 1000 & Eigengap   & 0.0153 & 0.0430 &  \textcolor{blue}{Yes}& $1.550 \pm 0.017$ & 1.817 \\  
Fashion-MNIST & LeNet & SGD & 1000 & Eigengap   & 0.0240 & 0.0430 &  \textcolor{blue}{Yes}& $1.520 \pm 0.017$ & 1.922 \\  
\bottomrule
\end{tabular}
}
\end{small}
\end{center}
\end{table*}

 \begin{table*}[t]
\caption{The Kolmogorov-Smirnov statistics of LeNet on CIFAR-10 and CIFAR-100. The estimated power exponent $\hat{\beta}$ and slope magnitude $\hat{s}$ are also displayed.}
\label{table:kslenet-cifar}
\begin{center}
\begin{small}
\resizebox{\textwidth}{!}{%
\begin{tabular}{lllll | lllll}
\toprule
Dataset & Model & Training & Sample size & Setting  & $d_{\rm{ks}}$ & $d_{\rm{c}}$ & Power-Law & $\hat{\beta} \pm \sigma$ & $\hat{s}$ \\
\midrule
CIFAR-10 & LeNet & \textcolor{red}{Random}  & 1000 & -   & 0.0663 & 0.0430 &  \textcolor{red}{No}& \\  
CIFAR-10 & LeNet & SGD & 1000 & -   & 0.0279 & 0.0430 &  \textcolor{blue}{Yes}& $1.968 \pm 0.031 $ & 1.033 \\ 
CIFAR-10 & LeNet & Vanilla SGD & 1000 & -   & 0.0276 & 0.0430 &  \textcolor{blue}{Yes}& $1.935 \pm 0.030 $ & 1.069 \\  
CIFAR-10 & LeNet & Adam & 1000 & -   & 0.0269 & 0.0430 &  \textcolor{blue}{Yes}& $1.806 \pm 0.025 $ & 1.241 \\  
CIFAR-10 & LeNet & AMSGrad & 1000 & -   & 0.0232 & 0.0430 &  \textcolor{blue}{Yes}& $1.786 \pm 0.025 $ & 1.271 \\  
CIFAR-10 & LeNet & AdaBound & 1000 & -   & 0.0297 & 0.0430 &  \textcolor{blue}{Yes}& $1.901 \pm 0.028$ & 1.110 \\
CIFAR-10 & LeNet & Yogi & 1000 & -   & 0.0184 & 0.0430 &  \textcolor{blue}{Yes}& $1.806 \pm 0.025$ & 1.241 \\
CIFAR-10 & LeNet & RAdam & 1000 & -   & 0.0163 & 0.0430 &  \textcolor{blue}{Yes}& $1.733 \pm 0.023$ & 1.363 \\
CIFAR-10 & LeNet & Adai & 1000 & -   & 0.0310 & 0.0430 &  \textcolor{blue}{Yes}& $1.918 \pm 0.029$ & 1.090 \\
CIFAR-10 & LeNet & PNM & 1000 & -   & 0.0347 & 0.0430 &  \textcolor{blue}{Yes}& $1.911 \pm 0.029$ & 1.098 \\
CIFAR-10 & LeNet & Lookahead & 1000 & -   & 0.0358 & 0.0430 &  \textcolor{blue}{Yes}& $1.964 \pm 0.030$ & 1.037 \\
CIFAR-10 & LeNet & DiffGrad & 1000 & -   & 0.0303 & 0.0430 &  \textcolor{blue}{Yes}& $1.803 \pm 0.024$ & 1.236 \\
\midrule
CIFAR-100 & LeNet & \textcolor{red}{Random}  & 1000 & -   & 0.0944 & 0.0430 &  \textcolor{red}{No}& \\  
CIFAR-100 & LeNet & SGD & 1000 & -   & 0.0315 & 0.0430 &  \textcolor{blue}{Yes}& $1.908 \pm 0.029 $ & 1.101 \\
CIFAR-100 & LeNet & Vanilla SGD & 1000 & -   & 0.0379 & 0.0430 &  \textcolor{blue}{Yes}& $1.903 \pm 0.029 $ & 1.108 \\
\midrule	
CIFAR-100 & LeNet & SGD & 1000 & Evaluated on CIFAR-10  & 0.0306 & 0.0430 &  \textcolor{blue}{Yes}& $1.913 \pm 0.029 $ & 1.095 \\
\bottomrule
\end{tabular}
}
\end{small}
\end{center}
\end{table*}

 \begin{table*}[t]
\caption{The Kolmogorov-Smirnov statistics of FCN. The estimated power exponent $\hat{\beta}$ and slope magnitude $\hat{s}$ are also displayed.}
\label{table:ksfcn}
\begin{center}
\begin{small}
\resizebox{\textwidth}{!}{%
\begin{tabular}{lllll | lllll}
\toprule
Dataset & Model & Training & Sample size & Setting  & $d_{\rm{ks}}$ & $d_{\rm{c}}$ & Power-Law & $\hat{\beta} \pm \sigma$ & $\hat{s}$ \\
\midrule
Avila & 2Layer-FCN & SGD & 50 & -   & 0.0683 & 0.176 &  \textcolor{blue}{Yes}& $1.604 \pm 0.085 $ & 1.656 \\
\midrule
MNIST & \textcolor{red}{1Layer-FCN} & \textcolor{red}{Random} & 1000 & -  & 0.185 & 0.0430 &  \textcolor{red}{No}  \\ 
MNIST & \textcolor{red}{1Layer-FCN} & SGD & 1000 & -   & 0.241 & 0.0430 &  \textcolor{red}{No}  \\  
MNIST & 2Layer-FCN & \textcolor{red}{Random} & 1000 & -  & 0.129 & 0.0430 &  \textcolor{red}{No}  \\ 
MNIST & 2Layer-FCN & SGD & 1000 & -   & 0.0112 & 0.0430 &  \textcolor{blue}{Yes}& $2.209 \pm 0.038$ & 0.827 \\  
MNIST & 4Layer-FCN & \textcolor{red}{Random} & 1000 & -  & 0.0628 & 0.0430 &  \textcolor{red}{No}  \\ 
MNIST & 4Layer-FCN & SGD & 1000 & -   & 0.0141& 0.0430 &  \textcolor{blue}{Yes}& $2.201 \pm 0.038$ & 0.833 \\  
\midrule
MNIST & 2Layer-FCN & SGD & 1000 & \textcolor{red}{Width=10}   & 0.149 & 0.0430 &  \textcolor{red}{No}  \\  
MNIST & 2Layer-FCN & SGD & 1000 & \textcolor{red}{Width=20}   & 0.185 & 0.0430 &  \textcolor{red}{No}  \\  
MNIST & 2Layer-FCN & SGD & 1000 & \textcolor{red}{Width=30}   & 0.0656 & 0.0430 &  \textcolor{red}{No}  \\  
MNIST & 2Layer-FCN & SGD & 1000 &  Width=50   & 0.0187 & 0.0430 &  \textcolor{blue}{Yes}& $2.138 \pm 0.028$ & 0.879 \\  
MNIST & 2Layer-FCN & SGD & 1000 &  Width=70   & 0.0376 & 0.0430 &  \textcolor{blue}{Yes}& $2.271 \pm 0.030$ & 0.787 \\  
MNIST & 2Layer-FCN & SGD & 1000 &  Width=100   & 0.0112 & 0.0430 &  \textcolor{blue}{Yes}& $2.209 \pm 0.038$ & 0.827 \\  
\bottomrule
\end{tabular}
}
\end{small}
\end{center}
\end{table*}

 \begin{table*}[t]
\caption{The Kolmogorov-Smirnov statistics of ResNet18. The estimated power exponent $\hat{\beta}$ and slope magnitude $\hat{s}$ are also displayed.}
\label{table:ksresnet18}
\begin{center}
\begin{small}
\resizebox{\textwidth}{!}{%
\begin{tabular}{lllll | lllll}
\toprule
Dataset & Model & Training & Sample size & Setting  & $d_{\rm{ks}}$ & $d_{\rm{c}}$ & Power-Law & $\hat{\beta} \pm \sigma$ & $\hat{s}$ \\
\midrule
CIFAR-10 & ResNet18 & \textcolor{red}{Random}  & 50 & -   & 0.334 & 0.176 &  \textcolor{red}{No}  \\  
CIFAR-10 & ResNet18 & SGD & 50 & -   & 0.0803 & 0.176 &  \textcolor{blue}{Yes}& $2.146 \pm 0.162 $ & 0.873 \\  
CIFAR-10 & ResNet18 & Vanilla SGD & 50 & -   & 0.0891 & 0.176 &  \textcolor{blue}{Yes}& $2.193 \pm 0.169 $ & 0.838 \\  
CIFAR-10 & ResNet18 & Adam & 50 & -   & 0.0478 & 0.176 &  \textcolor{blue}{Yes}& $2.062 \pm 0.149 $ & 0.950 \\   
CIFAR-10 & ResNet18 & AMSGrad & 50 & -   & 0.0542 & 0.176 &  \textcolor{blue}{Yes}& $2.041 \pm 0.147 $ & 0.961 \\  
CIFAR-10 & ResNet18 & AdaBound & 50 & -   & 0.0588 & 0.176 &  \textcolor{blue}{Yes}& $2.029 \pm 0.146 $ & 0.971 \\ 
CIFAR-10 & ResNet18 & Yogi & 50 & -   & 0.116 & 0.176 &  \textcolor{blue}{Yes}& $1.915 \pm 0.129 $ & 1.092 \\  
CIFAR-10 & ResNet18 & RAdam & 50 & -   & 0.168 & 0.176 &  \textcolor{blue}{Yes}& $1.794 \pm 0.1112 $ & 1.259 \\  
CIFAR-10 & ResNet18 & Adai & 50 & -   & 0.103 & 0.176 &  \textcolor{blue}{Yes}& $2.183 \pm 0.167 $ & 0.845 \\  
CIFAR-10 & ResNet18 & PNM & 50 & -   & 0.138 & 0.176 &  \textcolor{blue}{Yes}& $2.132 \pm 0.160 $ & 0.884 \\
CIFAR-10 & ResNet18 & Lookahead & 50 & -   & 0.110 & 0.176 &  \textcolor{blue}{Yes}& $2.098 \pm 0.155 $ & 0.911 \\  
CIFAR-10 & ResNet18 & DiffGrad & 50 & -   & 0.068 & 0.176 &  \textcolor{blue}{Yes}& $2.055 \pm 0.149 $ & 0.948 \\
\midrule
CIFAR-10 & ResNet18 & SGD & 50 & $B=512$   & 0.0561 & 0.176 &  \textcolor{blue}{Yes}& $2.146 \pm 0.151 $ & 0.936 \\ 
CIFAR-10 & ResNet18 & SGD & 50 & $B=1024$   & 0.0647 & 0.176 &  \textcolor{blue}{Yes}& $2.076 \pm 0.152 $ & 0.929 \\ 
CIFAR-10 & ResNet18 & SGD & 50 & $B=1152$   & 0.0598 & 0.176 &  \textcolor{blue}{Yes}& $2.060 \pm 0.150 $ & 0.944 \\ 
CIFAR-10 & ResNet18 & SGD  & 50 &  \textcolor{red}{$B=1280$}    & 0.331 & 0.176 &  \textcolor{red}{No}  \\  
CIFAR-10 & ResNet18 & SGD  & 50 &  \textcolor{red}{$B=2048$}    & 0.334 & 0.176 &  \textcolor{red}{No}  \\  
CIFAR-10 & ResNet18 & SGD  & 50 &  \textcolor{red}{$B=4096$}    & 0.334 & 0.176 &  \textcolor{red}{No}  \\  
CIFAR-10 & ResNet18 & SGD  & 50 &  \textcolor{red}{$B=16384$}    & 0.343 & 0.176 &  \textcolor{red}{No}  \\  
\midrule
CIFAR-100 & ResNet18 & \textcolor{red}{Random}  & 50 & -   & 0.373 & 0.176 &  \textcolor{red}{No}  \\  
CIFAR-100 & ResNet18 & SGD & 50 & -   & 0.108 & 0.176 &  \textcolor{blue}{Yes}& $2.299 \pm 0.184 $ & 0.770  \\  
\bottomrule
\end{tabular}
}
\end{small}
\end{center}
\end{table*}

\begin{table*}[t]
\caption{The Kolmogorov-Smirnov statistics of LNN. The estimated power exponent $\hat{\beta}$ and slope magnitude $\hat{s}$ are also displayed.}
\label{table:kslnn}
\begin{center}
\begin{small}
\resizebox{\textwidth}{!}{%
\begin{tabular}{lllll | lllll}
\toprule
Dataset & Model & Training & Sample size & Setting  & $d_{\rm{ks}}$ & $d_{\rm{c}}$ & Power-Law & $\hat{\beta} \pm \sigma$ & $\hat{s}$ \\
\midrule
MNIST & \textcolor{red}{4Layer-LNN} & SGD & 1000 & -  & 0.122 & 0.0430 &  \textcolor{red}{No}  \\ 
MNIST & 4Layer-LNN & SGD & 1000 & w/ BatchNorm  & 0.0411& 0.0430 &  \textcolor{blue}{Yes}& $2.190 \pm 0.037$ & 0.840 \\  
MNIST & 4Layer-LNN & SGD & 1000 & w/ ReLu  & 0.0127 & 0.0430 &  \textcolor{blue}{Yes}& $2.054 \pm 0.033$ & 0.949 \\  
\bottomrule
\end{tabular}
}
\end{small}
\end{center}
\end{table*}

 \begin{table*}[t]
\caption{The Kolmogorov-Smirnov Test Statistics of Protein.}
\label{table:ksprotein}
\begin{center}
\begin{small}
\resizebox{\textwidth}{!}{%
\begin{tabular}{ll | lllll}
\toprule
Protein & Sample size  & $d_{\rm{ks}}$ & $d_{\rm{c}}$ & Power-Law & $\hat{\beta} \pm \sigma$ & $\hat{s}$ \\
\midrule
4HHB  & 1000 &  0.0145 & 0.0430 &  \textcolor{blue}{Yes} & $2.008 \pm 0.032$ & 0.992 \\
\bottomrule
\end{tabular}
}
\end{small}
\end{center}
\end{table*}

\section{Experimental Settings}
\label{sec:experimentsetting}

\textbf{Computational environment.} The experiments are conducted on a computing cluster with NVIDIA\textsuperscript{\textregistered} V100 GPUs and Intel\textsuperscript{\textregistered} Xeon\textsuperscript{\textregistered} CPUs.

\subsection{Models, Datasets, and Optimizers}

\textbf{Models:} LeNet \citep{lecun1998gradient}, Fully Connected Networks (FCN), and ResNet18 \citep{he2016deep}. Particularly, we used one-layer FCN, two-layer FCN, four-layer FCN, which have 100 neurons for each hidden layer and use ReLu activations. 

\textbf{Datasets:} MNIST \citep{lecun1998mnist}, Fashion-MNIST \citep{xiao2017fashion}, CIFAR-10/100 \citep{krizhevsky2009learning}, and non-image Avila \citep{de2018reliable}.

\textbf{Optimizers:} SGD, Vanilla SGD, Adam \citep{kingma2014adam}, AMSGrad \citep{reddi2019convergence}, AdaBound \citep{luo2019adaptive}, Yogi \citep{zaheer2018adaptive}, RAdam \citep{liu2019variance}, Adai \citep{xie2022adaptive}, PNM \citep{xie2021positive}, Lookahead \citep{zhang2019lookahead}, and DiffGrad \citep{dubey2019diffgrad}.

\subsection{Image classification on MNIST and Fashion-MNIST}

\textbf{Data Preprocessing For MNIST and Fashion-MNIST:} We perform the common per-pixel zero-mean unit-variance normalization.

\textbf{Hyperparameter Settings:} We select the optimal learning rate for each experiment from $\{0.0001, 0.001, 0.01, 0.1, 1, 10\}$ for SGD and use the default learning rate for adaptive gradient methods. In the experiments on MNIST and Fashion-MNIST: $\eta=0.1$ for SGD, Vanilla SGD, Adai, PNM, and Lookahead; $\eta=0.1$ for Vanilla SGD;$\eta=0.001$ for Adam, AMSGrad, AdaBound, Yogi, RAdam, and DiffGrad. 

We train neural networks for 50 epochs on MNIST and 200 epochs on Fashion-MNIST. For the learning rate schedule, the learning rate is divided by 10 at the epoch of $40\%$ and $80\%$. The batch size is set to $128$ for MNIST and Fashion-MNIST, unless we specify it otherwise. 

The strength of weight decay is default to $\lambda = 0.0005$ as the baseline for all optimizers unless we specify it otherwise.

We set the momentum hyperparameter $\beta_{1} = 0.9$ for SGD and adaptive gradient methods which involve in Momentum. As for other optimizer hyperparameters, we apply the default settings directly.  


\subsection{Image classification on CIFAR-10 and CIFAR-100}

\textbf{Data Preprocessing For CIFAR-10 and CIFAR-100:} We perform the common per-pixel zero-mean unit-variance normalization, horizontal random flip, and $32 \times 32$ random crops after padding with $4$ pixels on each side. 

\textbf{Hyperparameter Settings:} We select the optimal learning rate for each experiment from $\{0.0001, 0.001, 0.01, 0.1, 1, 10\}$ for SGD and use the default learning rate for adaptive gradient methods. In the experiments on CIFAF-10 and CIFAR-100: $\eta=1$ for Vanilla SGD, Adai, and PNM; $\eta=0.1$ for SGD (with Momentum) and Lookahead; $\eta=0.001$ for Adam, AMSGrad, AdaBound, Yogi, RAdam, and DiffGrad. For the learning rate schedule, the learning rate is divided by 10 at the epoch of $\{80,160\}$ for CIFAR-10 and $\{100,150\}$ for CIFAR-100, respectively. The batch size is set to $128$ for both CIFAR-10 and CIFAR-100, unless we specify it otherwise. 

The strength of weight decay is default to $\lambda = 0.0005$ as the baseline for all optimizers unless we specify it otherwise. Recent work \citet{xie2020stable} found that popular optimizers with $\lambda = 0.0005$ often yields test results than $\lambda = 0.0001$ on CIFAR-10 and CIFAR-100. 

We set the momentum hyperparameter $\beta_{1} = 0.9$ for SGD with Momentum. As for other optimizer hyperparameters, we apply the default hyperparameter settings directly.  

\subsection{Learning with noisy labels}

We trained LeNet via SGD (with Momentum) on corrupted MNIST with various (asymmetric) label noise. We followed the setting of \citet{han2018co} for generating noisy labels for MNIST. The symmetric label noise is generated by flipping every label to other labels with uniform flip rates $\{ 40\%, 80\%\}$. In this paper, when we talk about label noise, we mean symmetric label noise.


We also randomly shuffle the labels of MNIST to produce MNIST with random labels, which has little knowledge behind the pairs of instances and labels.

\section{A Bridge to Protein Science} 
\label{sec:protein}

In this section, we discuss recent advancements on protein science and how it inspired our discussions on deep learning. 

As the basic building blocks of biological intelligence, proteins work as the main executors of various vital functions, including catalysis (enzyme), transportation (carrier proteins), defense (antibodies), and so on. The polypeptide chain made up of amino acid residues can fold into its native (energy-minimum) three-dimensional structure from a random coil. For most proteins, their ``correct'' native structures are essential to their functions. These structures (denote as $\vec{r}^0$) are not static. In contrast, they can perform their intrinsic dynamics due to the perturbations from the milieu. When external thermal noise act as force $\vec{\xi}$, the protein's deformation $\Delta \vec{r} = \vec{r} - \vec{r}^0$ can be estimated as: $\Delta \vec{r} = H^{-1} \vec{\xi}$, where $H$ is the elasticity matrix, i.e., the Hessian of the potential energy landscape. Such a framework is known as the elastic network model of the proteins \citep{atilgan2001anisotropy, bahar2010global}. When take the external thermal noises $\vec{\xi}$ as the \textit{input variable}, the deformation $\Delta \vec{r}$ can be recognized as the \textit{output} of a trained network, and the native structure $\vec{r}^0$ encoding the elastic network corresponds to the \textit{parameters} of the network.

\textbf{Proteins as accurate and robust learners.} It is worth noting that, in a long timescale, the native structure (network parameter) $\vec{r}^0$ has been gradually shaped by evolution. Due to random mutations, the native structure (energy-minimum point) and corresponding elastic network have varied. Thus, protein evolution can be recognized as an optimization process of the network parameters \citep{tang2021dynamics}. With the second-order Taylor approximation near a minimum, for a given native structure $\vec{r}^0$, the potential energy can be calculated as:
\begin{equation}
E(\vec{r}^0) = E(\vec{r}^{\star}) + \frac{1}{2} (\vec{r}^0 - \vec{r}^{\star})^{\top} H(\vec{r}^{\star}) (\vec{r}^0 - \vec{r}^{\star}),
\label{eqn:protein}
\end{equation}
in which $\vec{r}^{\star}$ denote the reference structure, a selected ancestral structure in the evolution, and $H(\vec{r}^{\star})$ is its corresponding Hessian. In this way, the evolution of a protein become comparable to the training of artificial neural networks. The potential energy $E$ corresponds to the loss function $L$, while native structure $\vec{r}^0$ corresponds to the model parameters $\theta$.

\textit{(1) Accuracy.} Proteins can respond with high susceptibility when perturbed by the environment, and some can undergo significant structural changes \citep{tang2017critical}. The noise-induced motions are highly anisotropic, with amino acid residues moving collectively in specific directions. These movements are usually related to the protein functions \citep{bahar2010global}. It is analogous to the model's generalization ability which describes how a model (protein) can deal with new data (different external noises) and make accurate predictions (specific functional dynamics). 

\textit{(2) Robustness.} In protein evolution, it is the functions of proteins that act as constraints, so essential functions must withstand most mutations \citep{guo2004protein, tang2021dynamics}. It is observed that, when the environment is stable, organisms tend to evolve in a convergent direction \citep{sato2020evolutionary, sakata2020dimensional}. Upon recognizing the protein as a learner, it is remarkable that the gradients (direction of evolution) on the loss landscape remain relatively stable. During evolution, most mutations do not affect the direction of the principal-component vectors (or the low-dimensional subspaces spanned by these vectors) related to the functions of the protein. This idea aligns with the discussions on the eigengaps in previous sections.

\begin{figure}
\center
\includegraphics[width = 0.6\columnwidth]{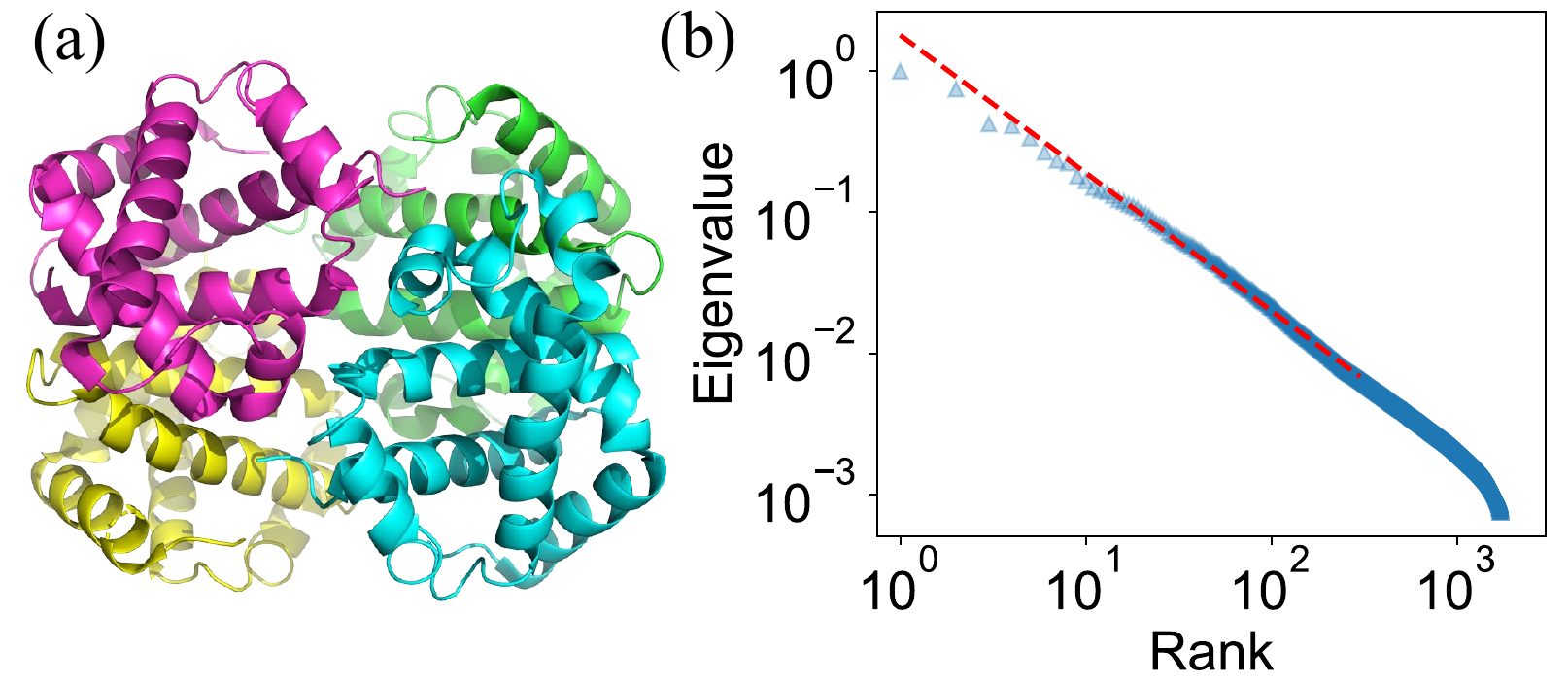}
\caption{(a) The cartoon illustration of human deoxyhemoglobin tetramer (PDB code: 4HHB). (b) The vibrational spectrum of the protein's elastic network. The estimated power exponent (slope magnitude) $\hat{s} = 0.992$ ($\hat{\beta} = 2.008 \pm 0.032$), which is very close to the estimated power exponent in deep learning. }
\label{fig:protein}
\end{figure}

\textbf{Power law and criticality.} We take a protein assembly (shown in Figure \ref{fig:protein}a) as an example to conduct normal mode analysis and obtain the vibration spectrum. This calculation is based on the elastic network model (See Appendix \ref{sec:enmprotein}). The result is shown in Figure \ref{fig:protein}b. We evaluate 9166 kinds of proteins in Section \ref{sec:empirical}. While recent studies \citep{reuveni2008proteins, tang2020long, tang2020functional} implicitly or explicitly suggested that the vibration spectrum of proteins exhibits a power-law distribution, we are the first to conduct large-scale statistical tests on the power-law spectrums for protein. The power-law behaviors suggest the parallels between protein evolution and deep learning. 
Interestingly, similar power laws were observed in various kinds of other natural systems, including brains, bird flocks, insect swarms, and so on \citep{munoz2018colloquium}. In statistical physics, power laws act as the hallmark of ``critical point'' between ordered (robustness) and disordered (plasticity) states. 

\textbf{A Large-Scale Study on 9166 kinds of Proteins.} Surprisingly, we observed $\hat{s} \approx 1$ for the protein in Figure \ref{fig:protein} and in the spectrums of total 9166 kinds of protein molecules from the Protein Data Bank (PDB) \citep{PDB}. Each of the protein molecule has $100\leq N_{\rm{AA}}\leq2000$ amino acid residues. All of the protein spectrums successfully passed the power-law KS tests with $mean(\hat{s})= 1.045$ ($mean(\hat{\beta})= 1.975$). The approximation $\hat{s} \approx 1$ holds even better for large proteins in terms of the mean and the standard error, shown in Table \ref{table:ksproteinsize}. We conjecture there may exist some universal underly mechanism for deep learning and protein. 

\begin{table}[t]
\caption{The estimated $\hat{s}$ for various sized proteins.}
\label{table:ksproteinsize}
\begin{center}
\begin{small}
\begin{sc}
\resizebox{0.6\textwidth}{!}{%
\begin{tabular}{l | lll}
\toprule
  Slope & $300 \leq 3N_{\rm{AA}} \leq 1000$  & $1000 \leq 3N_{\rm{AA}} \leq 3000$ & $3000 \leq 3N_{\rm{AA}} \leq 6000$   \\
  \midrule
$\hat{s}$  & $1.050 \pm 0.175$  &  $1.041 \pm 0.119$  &  $1.002 \pm 0.084$ \\
\bottomrule
\end{tabular}
}
\end{sc}
\end{small}
\end{center}
\end{table}

\section{The Spectral Analysis of Proteins}
\label{sec:enmprotein}

The elastic network models are widely applied to predict and characterize the slow global dynamics of a wide range of proteins and bio-machineries \citep{bahar2010global, tang2020long}. By describing the proteins as mass-and-spring networks, the elastic network models can capture the functional motions of proteins with minimal computational resources by focusing on the movement of proteins nearby the native structure. The movements of the proteins are described as the linear vibrations around the energy minimum of the energy landscape. It is worth noting that the model is not applicable for the dynamics far from the energy minimum, such as protein folding and unfolding problems.

\subsection{Anisotropic Network Model (ANM)}

In this work, we employ a typical form of the ENM, Anisotropic Network Model (ANM), to calculate the vibrational spectrum of proteins \citep{atilgan2001anisotropy, bahar2010global}. Previous research has shown that not only can ANM accurately reproduce the movements of residues as determined by experiments, but the model also fits well with the data on protein structure evolution \citep{tang2021dynamics}.

To introduce the model settings of ANM, let us first consider the sub-system consisting of two nodes (amino acid residues) $i$ and $j$ connected with a harmonic spring. The coordinates of the two nodes are $\vec{r}_i = [x_i, y_i, z_i]$ and $\vec{r}_j = [x_j, y_j, z_j]$, respectively; and their native-state coordinates are $\vec{r}^0_i = [x^0_i, y^0_i, z^0_i]$ and $\vec{r}^0_j = [x^0_j, y^0_j, z^0_j]$.
When the equilibrium distance between them is given by 
$s_{ij}^0 = |\vec{r}^0_i - \vec{r}^0_j| = \sqrt{(x^0_i - x^0_j)^2 + (y^0_i - y^0_j)^2 + (z^0_i - z^0_j)^2}$, 
and the instantaneous distance is given by 
$s_{ij} = |\vec{r}_i - \vec{r}_j| =  \sqrt{(x_i - x_j)^2 + (y_i - y_j)^2 + (z_i - z_j)^2}$, 
then the potential of such a sub-system can be given as:
\begin{equation}
V_{ij} = \frac{\kappa}{2} \left( s_{ij} - s_{ij}^0 \right)^2,
\end{equation}
where $\kappa$ denotes the spring constant.
Then, around the energy minimum, the second-order derivative of the $V_{ij}$ with respect to $\vec{r}_i$ can be given as:
\begin{equation}
    \frac{\partial ^2 V_{ij}}{\partial x_i^2} = \frac{\partial ^2 V_{ij}}{\partial x_j^2} = \frac{\kappa}{s_{ij}^2}(x_j - x_i)^2,
\end{equation}
\begin{equation}
    \frac{\partial ^2 V_{ij}}{\partial x_i \partial y_j} =  \frac{\kappa}{s_{ij}^2}(x_j - x_i)(y_j - y_i).
\end{equation}
Therefore, the corresponding Hessian matrix entries $H_{ij}$ can be given as:
\begin{equation}
    H_{ij} = -\frac{\kappa}{s_{ij}^2} \left[
    \begin{matrix}
    x_j - x_i \\ y_j - y_i \\ z_j - z_i
    \end{matrix} \right] [x_j - x_i, y_j - y_i, z_j - z_i].
\end{equation}

In this way, the $3N_{\rm{AA}} \times 3N_{\rm{AA}}$ Hessian $H$ can be recognized as the direct sum of the $3\times 3$ Hessian matrix $H_{ij}$ and $N_{\rm{AA}} \times N_{\rm{AA}}$ elasticity matrix $\Gamma$. That is, the $3N_{\rm{AA}} \times 3N_{\rm{AA}}$ Hessian matrix $H$ can be recognized as an $N_{\rm{AA}} \times N_{\rm{AA}}$ matrix with entries of $3\times 3$ matrices
\begin{equation}
 H_{ij} = -\frac{\kappa \cdot \Gamma_{ij} } {s^2_{ij}} \left[
 \begin{matrix}
 x_j - x_i \\ y_j - y_i \\ z_j - z_i
 \end{matrix} \right] [x_j - x_i, y_j - y_i, z_j - z_i].
\end{equation}

Here, matrix $\Gamma$ is defined according to the residue-residue contact topology of the native structure. In ANM, only spatial-neighboring residues are considered to be connected. For a pair of amino acid residues ($i$ and $j$), if their mutual distance $r_{ij} \leq r_{\rm{C}}$, then $\Gamma_{ij} = -1$; if $r_{ij} > r_{\rm{C}}$, then $\Gamma_{ij} = 0$; and for the diagonal elements, $\Gamma_{ii} = -\sum_{j\neq i} \Gamma_{ij} = -k_i,$ where $k_i$ denote the degree of node $i$. Note that in a graph theory perspective, matrix $\Gamma$ is also known as the graph Laplacian (or the Kirchhoff matrix) of the residue-residue contact network. In ANM with homogenous contact strength ($\kappa=1$), the only control parameter is the cutoff distance $r_{\rm{C}}$. In the calculation, we take $r_{\rm{C}}= 9.0 $ \r{A}. 

For a protein consisting of $N_{\rm{AA}}$ amino acid residues, the total degrees of freedom is $3 N_{\rm{AA}}$. Among them, there are 6 degrees of freedom related to the rigid body motion, say, three-dimensional translational motions and three-dimensional rotation. Therefore, to fully describe the structural fluctuations of a protein, one need in total $n = 3  N_{\rm{AA}} -6$ parameters.

\subsection{From ANM to SGD: Thermal noise vs. structured noise}
Although the energy landscape around a protein's native state is similar to the loss landscape of a deep neural network, the corresponding noises are entirely different. Protein dynamics are driven by thermal noises, and deep learning is driven by structured noises. Due to these two types of noise, the Hessian matrix and the covariance matrix have different relationships.

For a protein molecule driven by the thermal noises, according to the Boltzmann distribution, the probability of a structure $\Delta \vec{r}$ is given as:
\begin{equation}
p(\Delta \vec{r}) \sim e^ { - \frac{1}{2} \sum_{i,j=1}^{N_{\rm{AA}}} \Delta \vec{r}_i \cdot H_{ij} \cdot \Delta \vec{r}_j },
\end{equation}
in which $\Delta \vec{r}_i = \vec{r}_i - \vec{r}^0_i$ and $\Delta \vec{r}_j = \vec{r}_j - \vec{r}^0_j$ are three-dimensional vectors describing the displacements of residues $i$ and $j$. It is worth noting that $p(\Delta \vec{r})$ is a multivariate Gaussian distribution with covariance matrix $C  \sim H^{-1} $, where $C_{ij} = \langle \Delta \vec{r}_i \cdot \Delta \vec{r}_j \rangle$. 

However, as discussed in the main text, for SGD, the covariance matrix is proportional to the Hessian matrix: $C_{\rm{sgd}}(\theta) \sim H(\theta)$. Due to this difference, it is the spectrum of the inverse (or pseudoinverse) of Hessian $H^{-1}$ that should be applied to compare with the Hessian matrices in deep learning. 

\subsection{Similarities in Hessian spectrums}

By diagonalizing the Hessian matrix $H$, we can obtain all the nonzero eigenvalues and the corresponding eigenvectors describing the motions of the protein, i. e., $H = V \tilde{\Lambda} V^{\top}$, in which the eigenvalues $\tilde{\Lambda} = \text{diag}[\tilde{\lambda}_1, \tilde{\lambda}_2, \cdots \tilde{\lambda}_{n}]$ ($\tilde{\lambda}_1 \leq \tilde{\lambda}_2, \leq \cdots \leq \tilde{\lambda}_{n}$) and eigenvectors $V = [v_1, v_2, \cdots v_{n}]^{\top}$. 

Then, for the inverse Hessian $H^{-1}$, its nonzero eigenvalues $\sigma_i = {1}/{\tilde{\lambda}_i}$, and $\sigma_1  \geq \sigma_2  \geq \cdots \geq \sigma_{n} $. In Fig. \ref{fig:protein}b, as an example, the rank-size distribution of $\sigma_i$ vs $i$ is plotted. Here, we normalize all the eigenvalues as $\hat{\sigma}_i = \sigma_i/\sigma_1$.  The estimated power exponent (slope magnitude) $\hat{s} = 0.992$ (corresponding to $\hat{\beta} = 2.008 \pm 0.032$). This result clearly demonstrates the similarity between the vibrational spectrum of proteins and the Hessian spectrum in deep learning.

\subsection{Protein Dataset}

\begin{figure}
\center
\subfigure[The Slope Magnitude]{\includegraphics[width =0.33\columnwidth ]{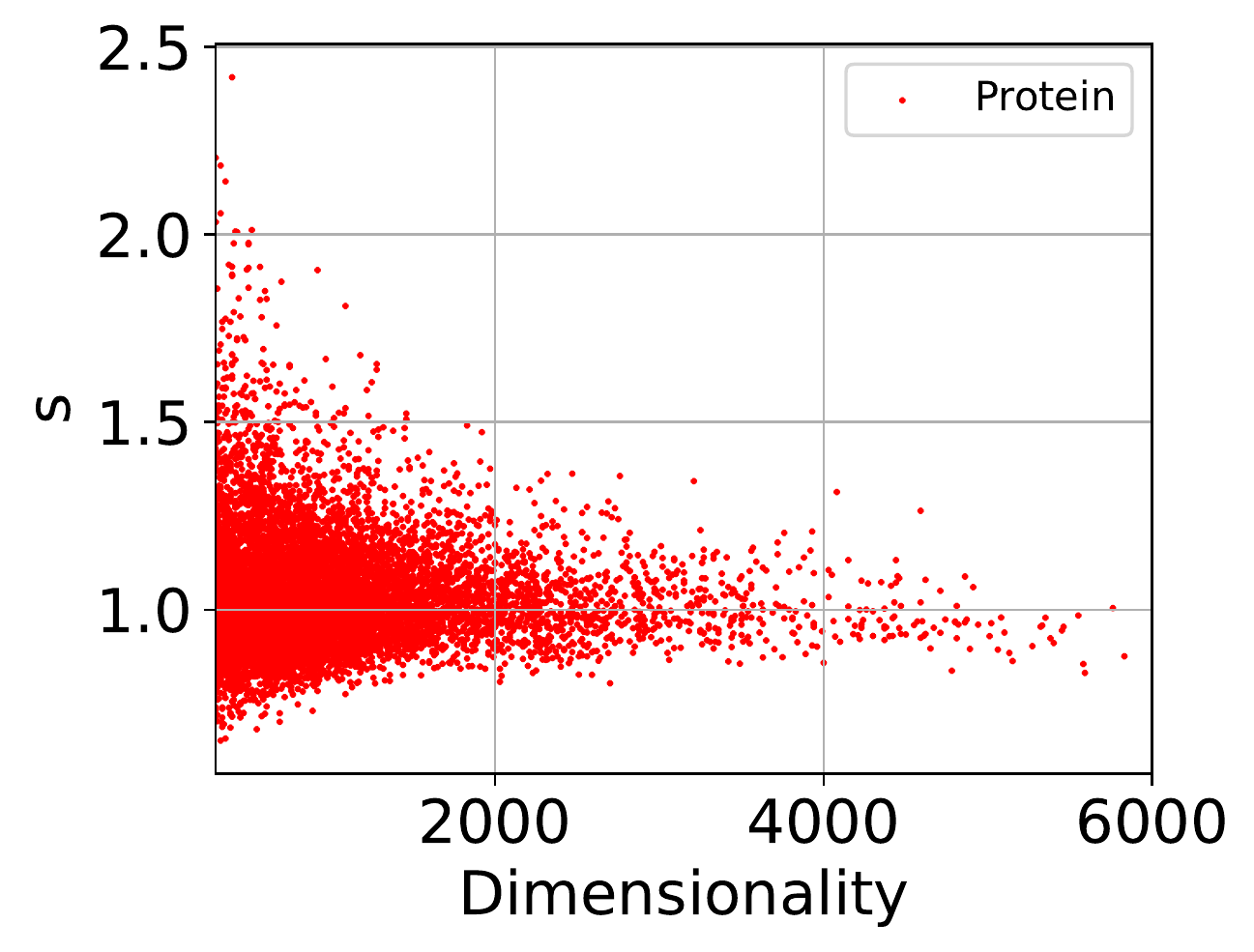}} 
\subfigure[Goodness-of-fit]{\includegraphics[width =0.33\columnwidth ]{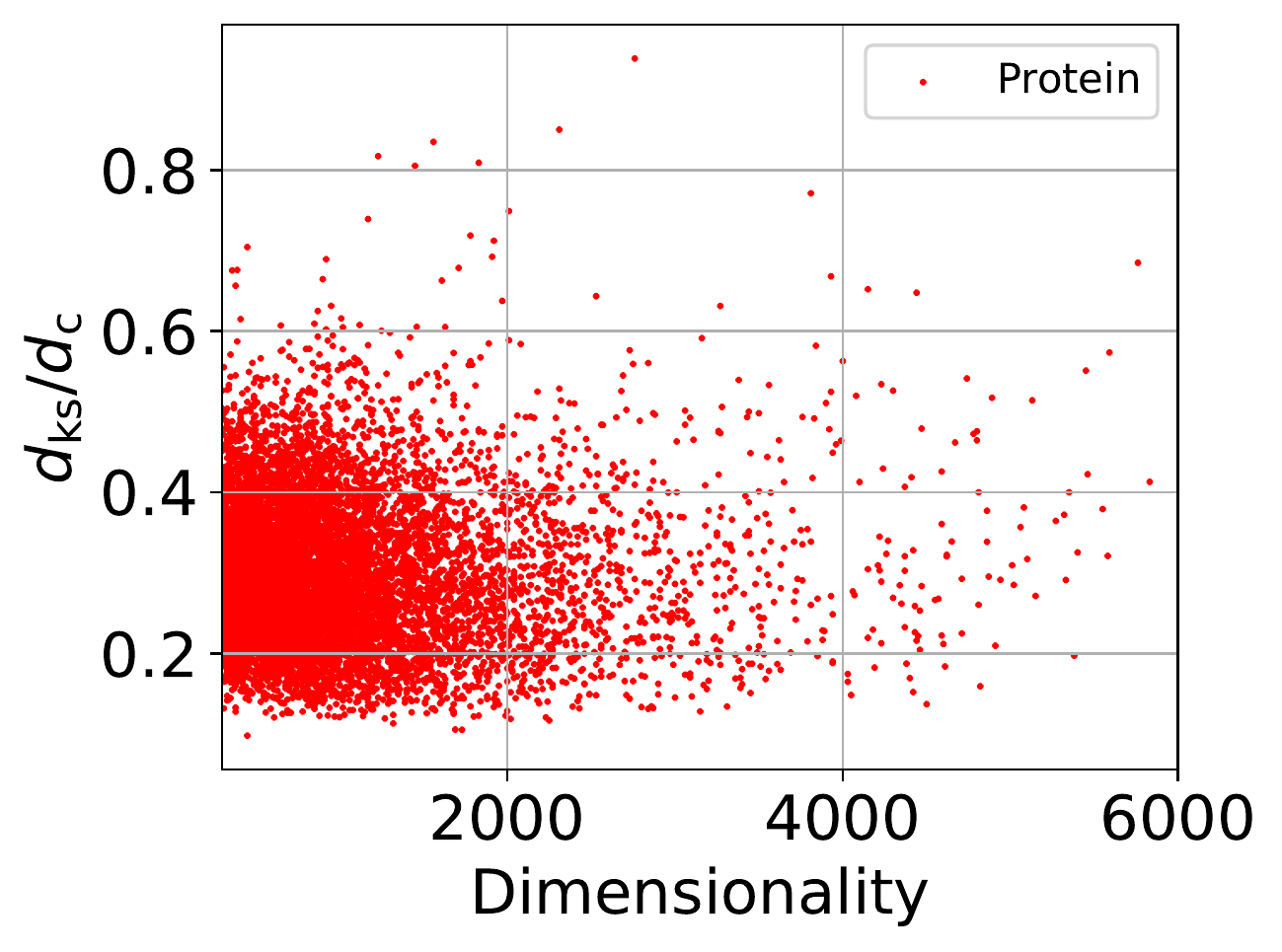}} 
\caption{The KS statistics of the spectrums of 9166 kinds of protein molecule.}
 \label{fig:proteinks}
\end{figure}

\begin{table*}[t]
\caption{The approximation $s \approx 1$ positively correlates with the size of protein.}
\label{table:ksallprotein}
\begin{center}
\begin{small}
\resizebox{\textwidth}{!}{%
\begin{tabular}{l | lll | l}
\toprule
Estimated Parameter & $300 \leq 3N_{\rm{AA}} \leq 1000$  & $1000 \leq 3N_{\rm{AA}} \leq 3000$ & $3000 \leq 3N_{\rm{AA}} \leq 6000$ & $300 \leq 3N_{\rm{AA}} \leq 6000$   \\
\midrule
$\hat{\beta}$  & $1.976 \pm 0.143$ &  $1.973 \pm 0.103$ & $2.004 \pm 0.080$  & $1.975 \pm 0.128$  \\
\midrule
$\hat{s}$  & $1.050 \pm 0.175$  &  $1.041 \pm 0.119$  &  $1.002 \pm 0.084$  & $1.045 \pm 0.155$ \\
\midrule
$d_{\rm{ks}}/d_{\rm{c}}$  & $0.304$ &  $0.292$ & $0.317$ & $0.300$\\
\bottomrule
\end{tabular}
}
\end{small}
\end{center}
\end{table*}

In this work, we evaluated the vibrational spectrums of 9166 kinds of protein molecules from the Protein Data Bank (PDB) \citep{PDB}. The studied proteins have high-resolution and clean structures, are large enough, and have enough diversity. More precisely, the structures of these proteins were all determined via high-resolution X-ray diffraction ($\leq 2.0 \text{\AA}$) without DNA, RNA, or hybrid structures involved. Their chain lengths (number of amino acid residues) are $100 \leq N_{\rm{AA}} \leq 2000$. Their spectrums have no zero eigenvalues except for the six modes correspond to the translational and rotational motions. Every two proteins share less than 30\% sequence similarity. We choose the $\lambda_{\rm{cutoff}}$ as the top $\frac{1}{10}$ largest eigenvalue in each vibrational spectrum for corresponding proteins.

We present the KS statistics of the protein spectrums in Figure \ref{fig:proteinks}. Note that, if $\frac{d_{\rm{ks}}}{d_{\rm{c}}} < 1$, we would reject the null hypothesis and accept the power-law hypothesis. All of the protein spectrums successfully passed the power-law KS tests with $mean(\hat{s})= 1.045$ ($mean(\hat{\beta})= 1.975 \pm 0.128$). We also notice that $\hat{s} \approx 1$ holds even better for larger proteins, which is supported by Table \ref{table:ksallprotein}.


\section{Supplementary Empirical Results}
\label{sec:supresults}

\begin{figure}
\center
\includegraphics[width =0.4\columnwidth ]{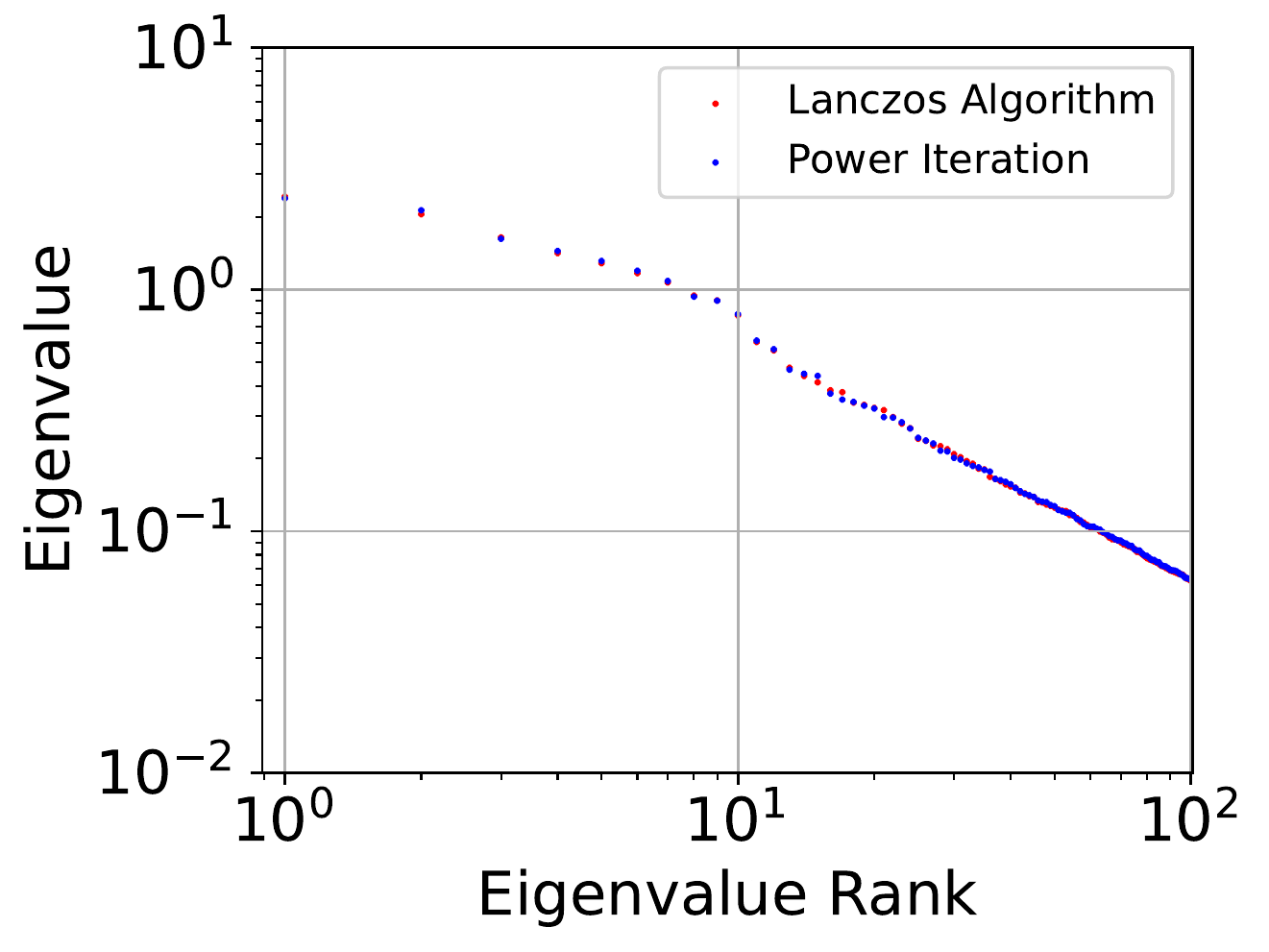}
\caption{The spectrum via Power Iteration Algorithm is highly consistent with the spectrum via Lanczos Algorithm. It also shows that the power-law spectrum is caused by the properties of deep learning rather than the stochasticity of Lanczos Algorithm.}
 \label{fig:poweriteration}
\end{figure}

\begin{figure}
\center
\subfigure[Eigenvalue Rank]{\includegraphics[width =0.33\columnwidth ]{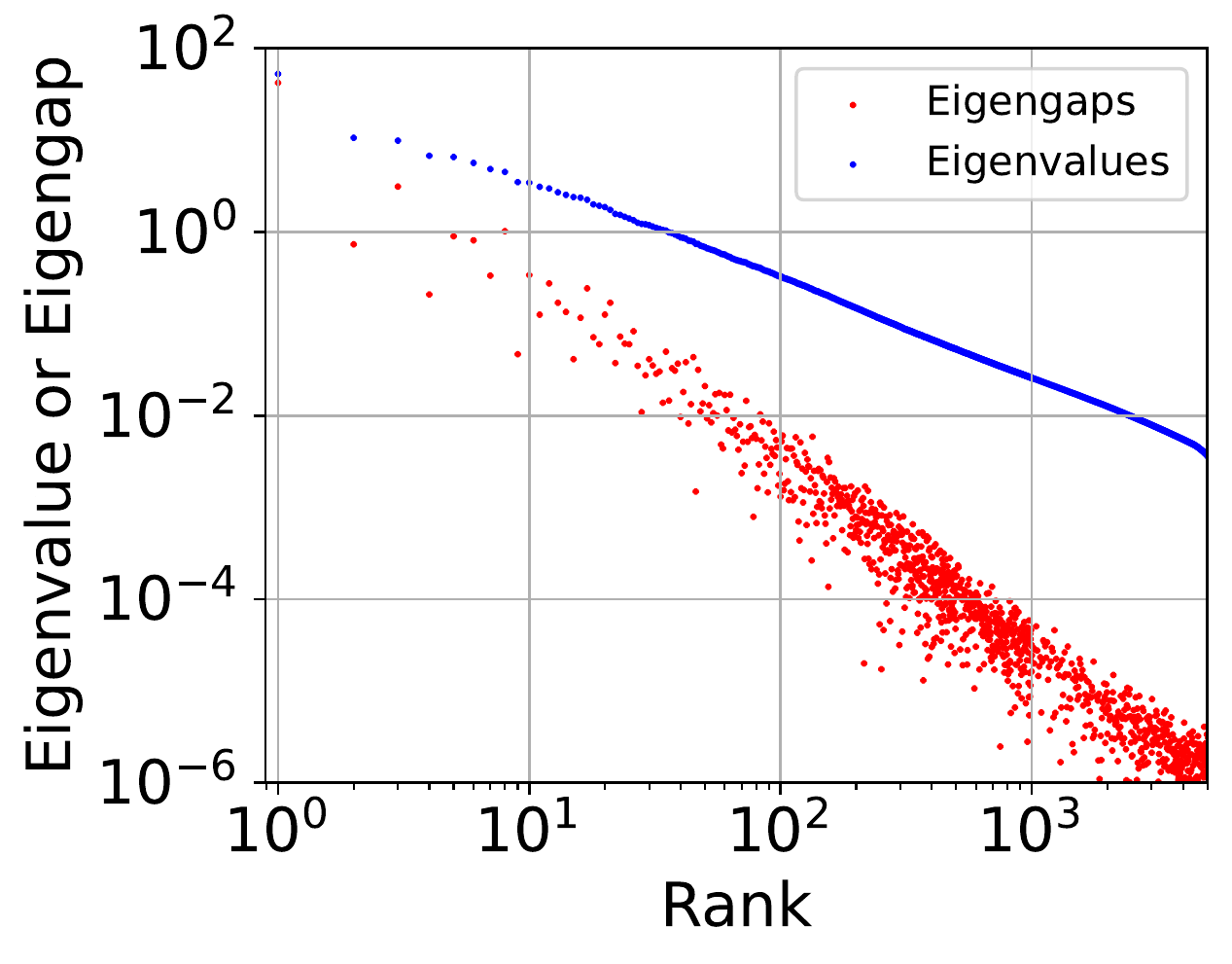}} 
\subfigure[Eigengap Rank]{\includegraphics[width =0.33\columnwidth ]{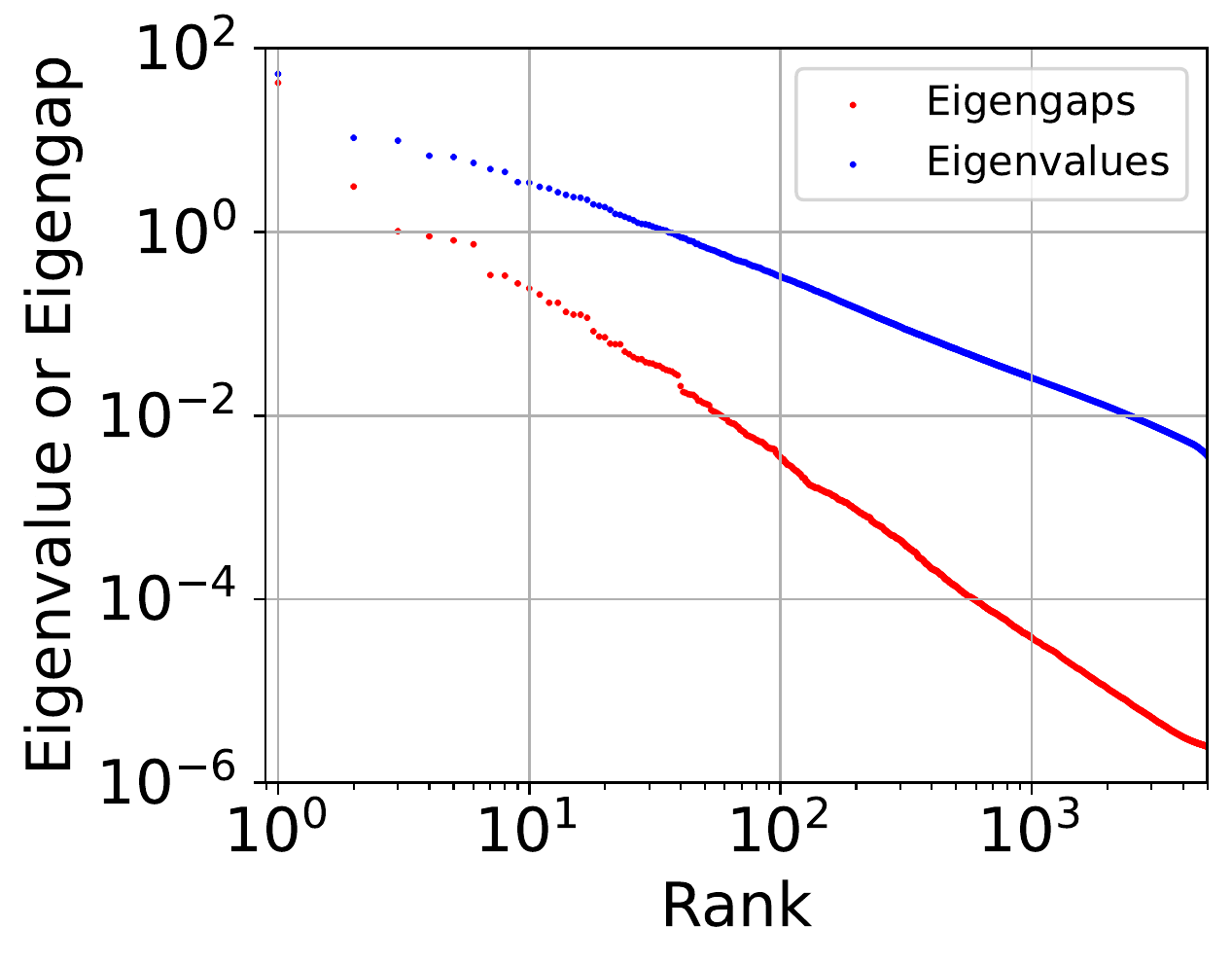}}
\caption{ The power-law Hessian eigengaps in deep learning. Model: LeNet. Datset: Fashion-MNIST. Subfigure (a) displayed the eigengaps by original rank indices sorted by eigenvalues. Subfigure (b) displayed the eigengaps by rank indices re-sorted by eigengaps. }
 \label{fig:fmnisteigengap}
\end{figure}

\begin{figure}
\center
\includegraphics[width =0.4\columnwidth ]{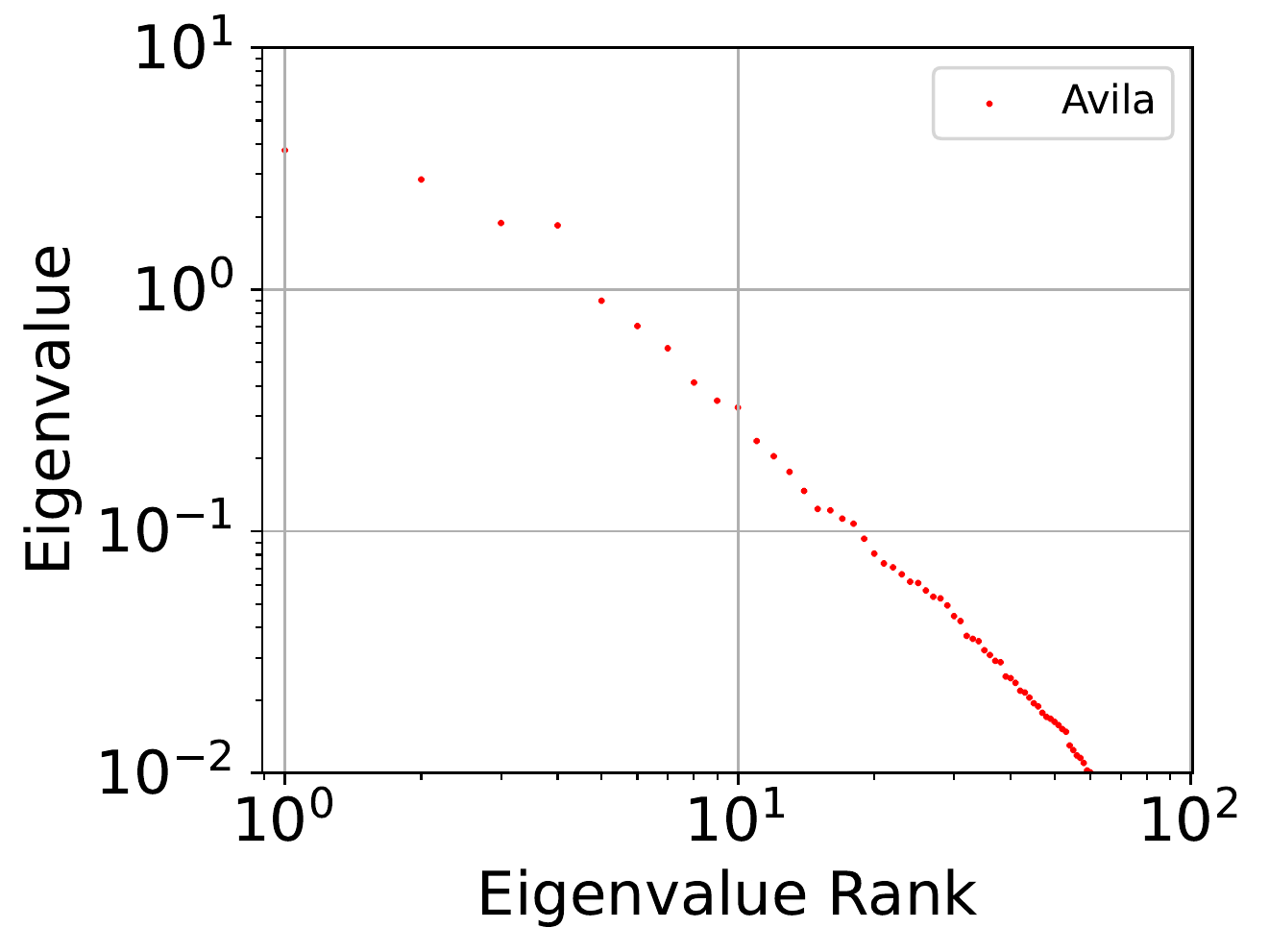}
\caption{The spectrum of FCN on Avila Dataset. It shows that the power-law spectrum of neural networks may also exist in non-image datasets.}
 \label{fig:avilaspectrum}
\end{figure}

We compared the spectrum computed via Power Iteration Algorithm and Lanczos Algorithm in Figure \ref{fig:poweriteration}. It shows the spectrum via Power Iteration Algorithm is highly consistent with the spectrum via Lanczos Algorithm. It also demonstrates that the power-law spectrum is caused by the properties of deep learning rather than the stochasticity of Lanczos Algorithm.

We presented the power-law eigengaps on Fashion-MNIST in Figure \ref{fig:fmnisteigengap}. It shows that the power-law eigengaps on Fashion-MNIST are highly consistent with the power-law eigengaps on MNIST.

\begin{figure}
\center
\subfigure{\includegraphics[width =0.4\columnwidth ]{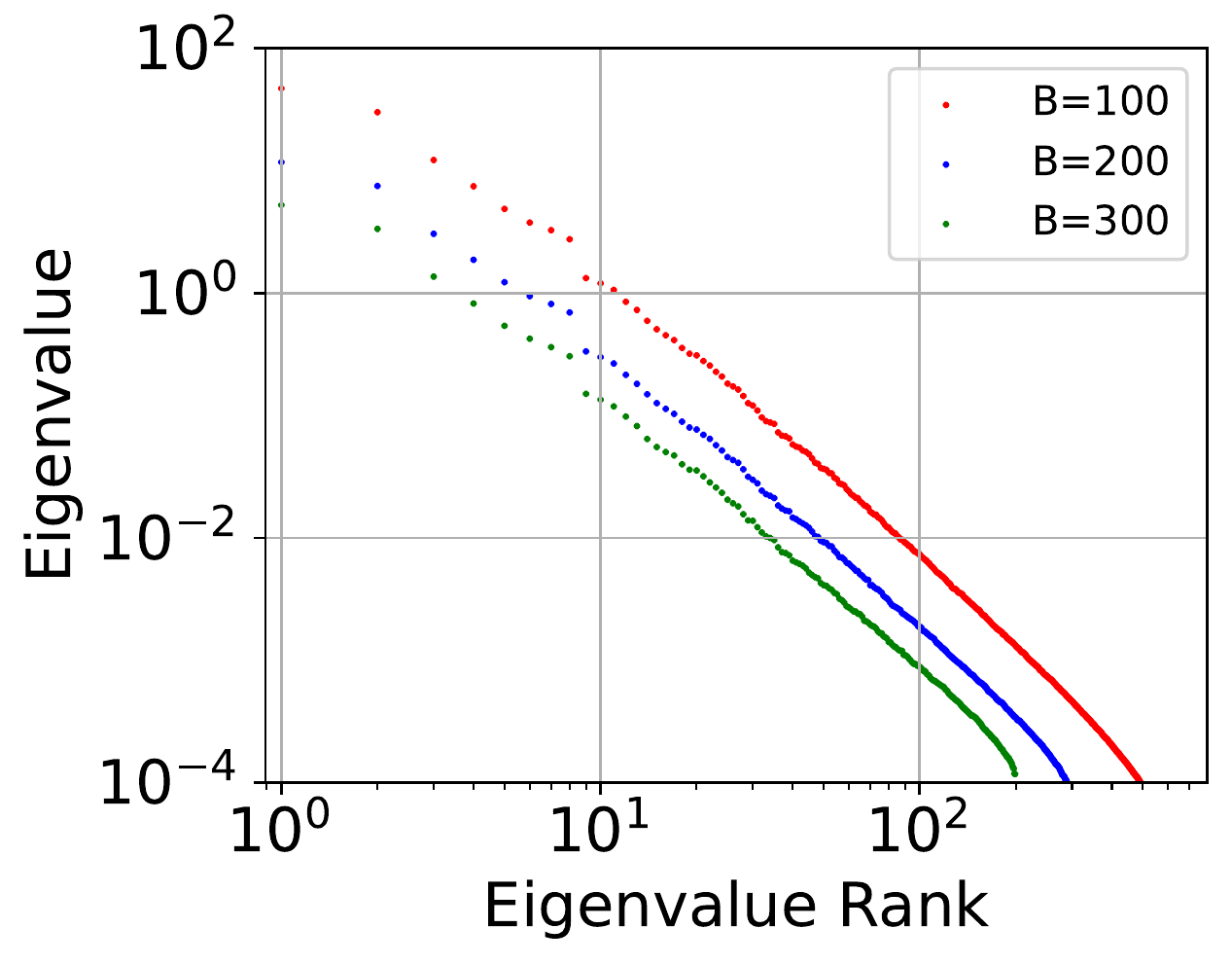}} 
\caption{ The power-law spectrum of gradient noise covariance exist in deep learning for various batch sizes. Model: Fully Connected Network(FCN). Datset: MNIST.}
 \label{fig:noisespectrum}
\end{figure}

We presented the power-law spectrum of the covariance matrix of stochastic gradient noise of FCN on MNIST in Figure \ref{fig:noisespectrum}. As the inverses of the power-law variables are power-law, the covariance spectrum shows heavy-tail properties. It demonstrates that the heavy-tail property belongs to deep neural networks rather than SGD itself.

We presented the power-law spectrum of two-layer FCN on Avila Dataset in Figure \ref{fig:avilaspectrum}. It shows that the power-law spectrum of neural networks may also generally exist in non-convolution neural networks trained on a non-image dataset. Particularly, we note that the Avila Dataset has only ten attributes, including intercolumnar distance, upper margin, lower margin, exploitation, row number, modular ratio, interlinear spacing, weight, peak number, and modular ratio/ interlinear spacing. These attributes are essentially different from the pixels in image datasets. 

\begin{figure}
\center
\includegraphics[width =0.4\columnwidth ]{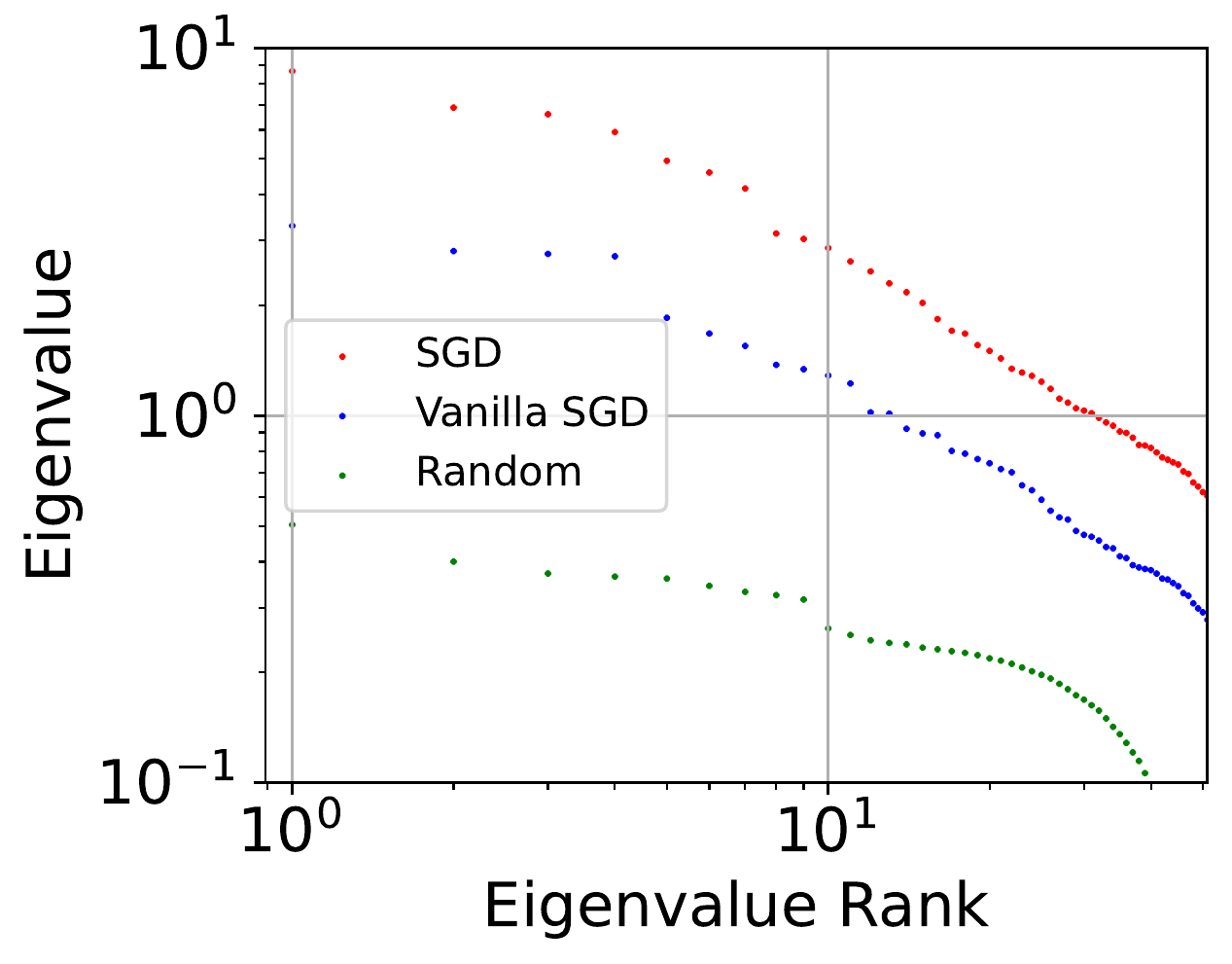}
\caption{The power-law spectrums of ResNet18 on CIFAR-10. It shows that the power-law spectrum of neural networks may also exist in modern neural network architectures (ResNet) as well as simple CNNs/FCNs. }
 \label{fig:resnetspectrum}
\end{figure}

We presented the power-law spectrums of ResNet18 on CIFAR-10 in Figure \ref{fig:resnetspectrum}. It shows that the power-law spectrums hold for ResNet, which is a representative of the modern neural network architectures, as well as simple CNNs/FCNs. Due to the GPU memory limit, we may only display the top 50 eigenvalues for ResNet18. However, the KS test still supports to accept the power-law hypothesis.

\begin{figure}
\center
\subfigure{\includegraphics[width =0.4\columnwidth ]{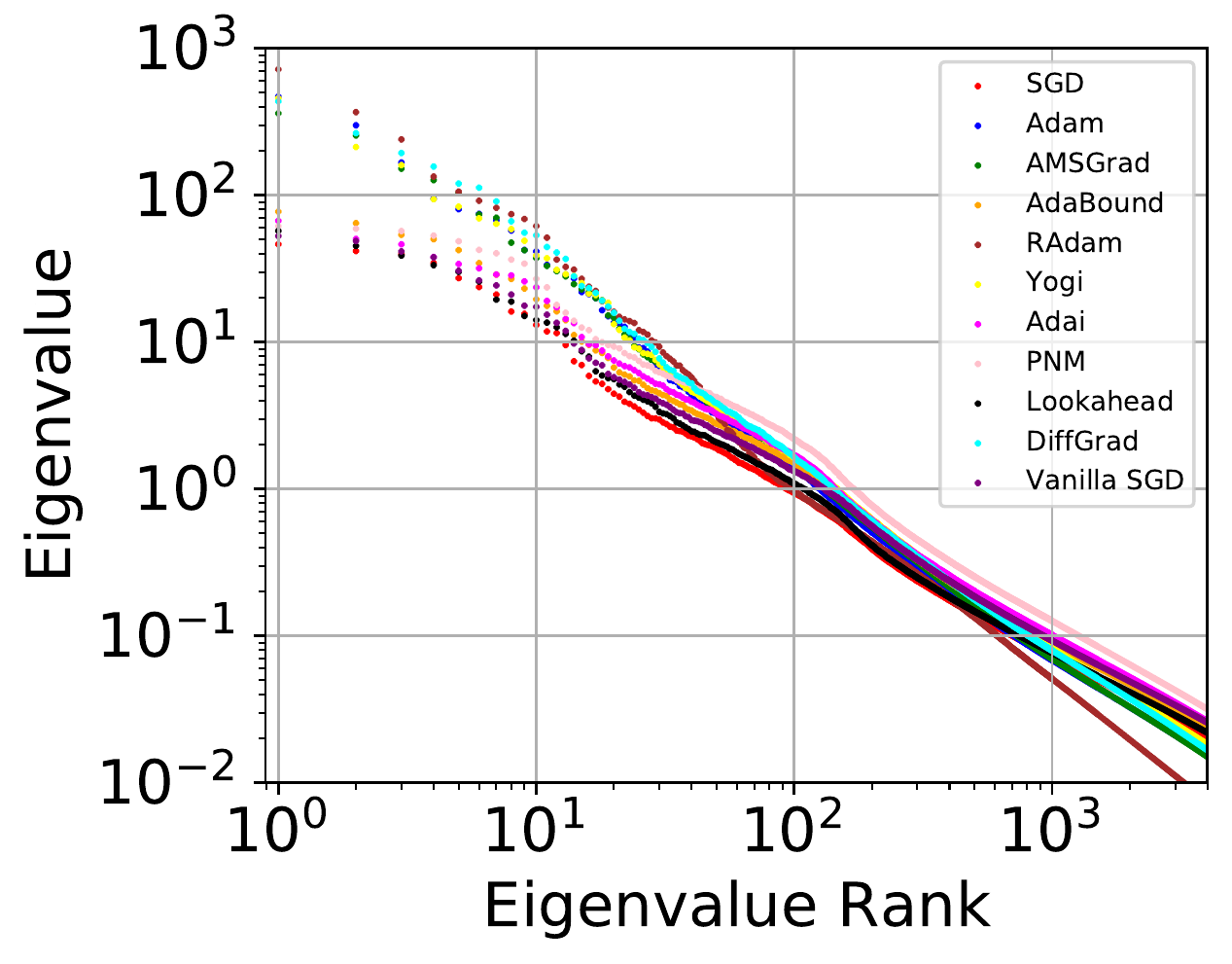}} 
\subfigure{\includegraphics[width =0.4\columnwidth ]{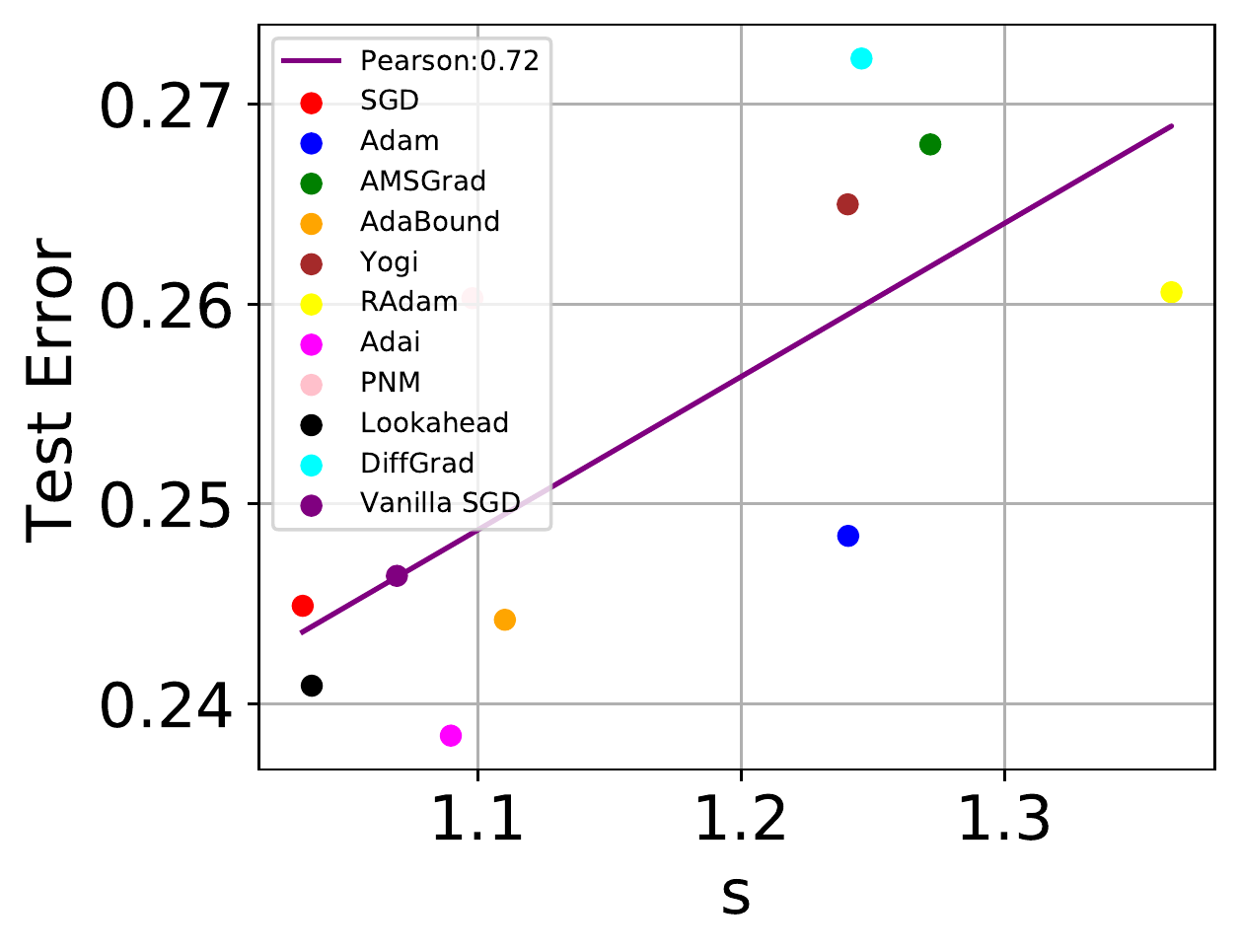}} 
\caption{The power-law spectrums hold across optimizers. Moreover, the slope magnitude $\hat{s}$ is an indicator of minima sharpness and a predictor of test performance. Model:LeNet. Dataset: CIFAR-10.}
 \label{fig:optimizerspectrum-cifar10}
\end{figure}

\begin{figure}
\center
\subfigure{\includegraphics[width =0.4\columnwidth ]{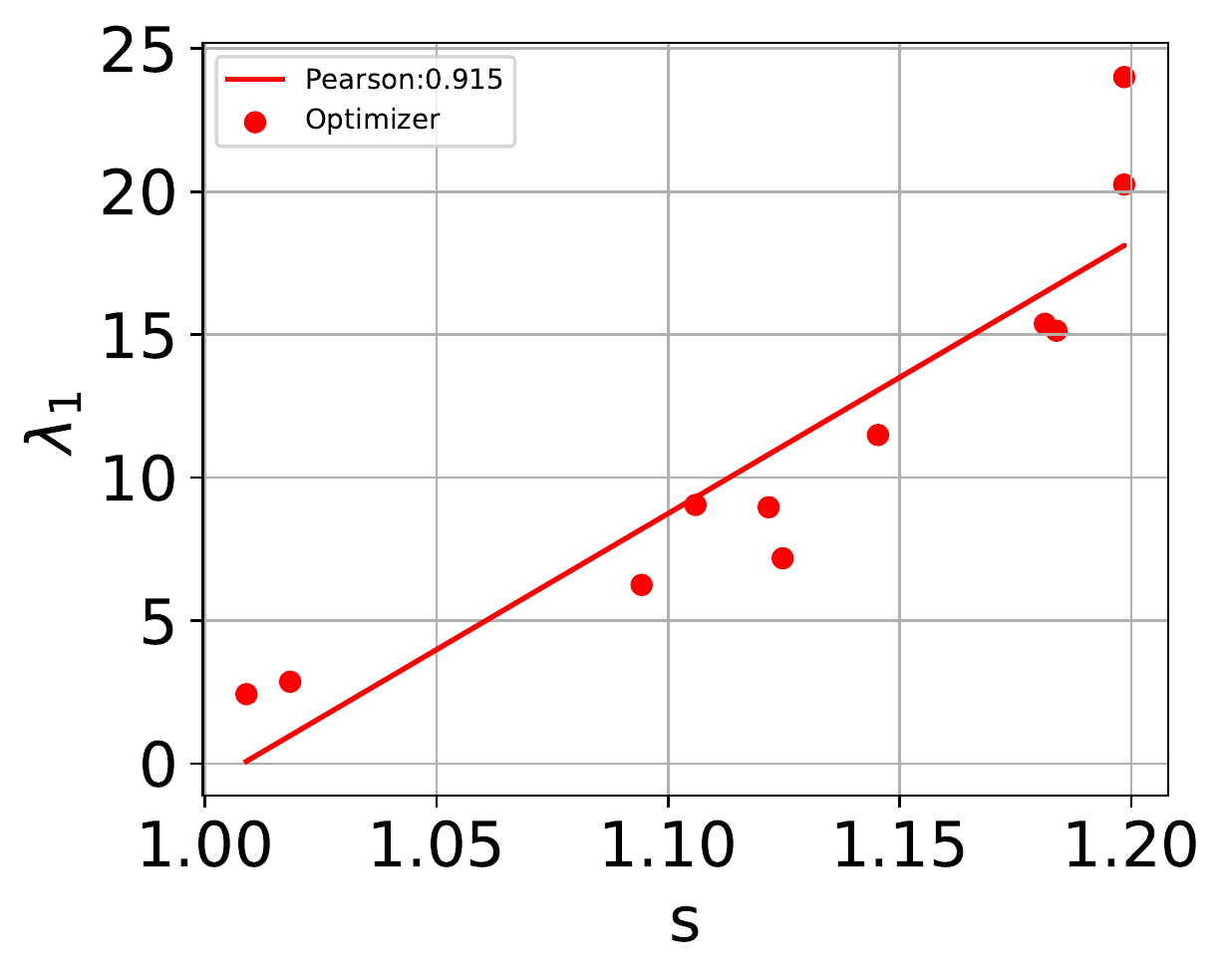}} 
\subfigure{\includegraphics[width =0.4\columnwidth ]{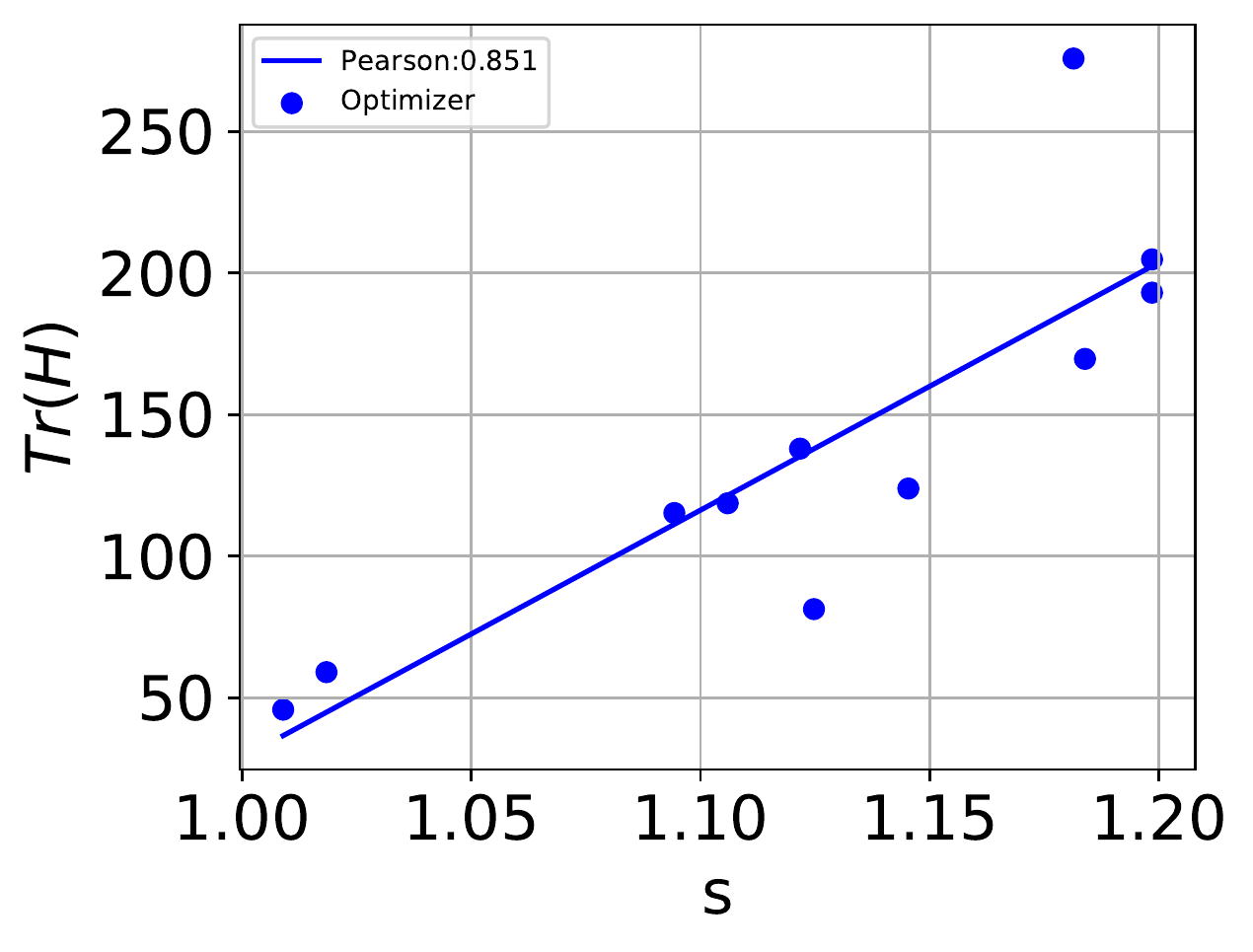}} 
\caption{The slope magnitude $\hat{s}$ closely correlates to the largest Hessian eigenvalue and the Hessian trace. Model:LeNet. Dataset: MNIST.}
 \label{fig:slopesharpness-mnist}
\end{figure}

\begin{figure}
\center
\subfigure{\includegraphics[width =0.4\columnwidth ]{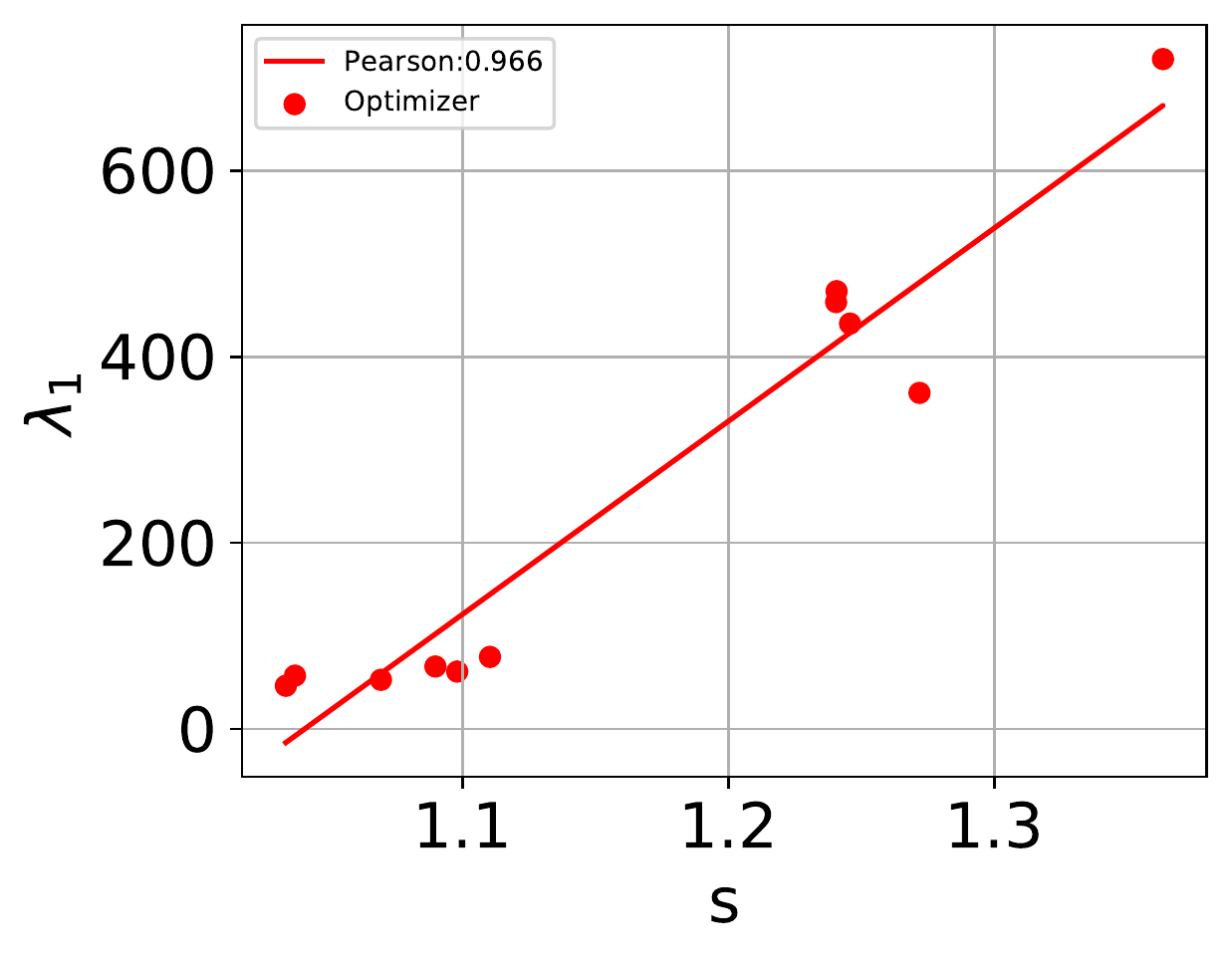}} 
\subfigure{\includegraphics[width =0.4\columnwidth ]{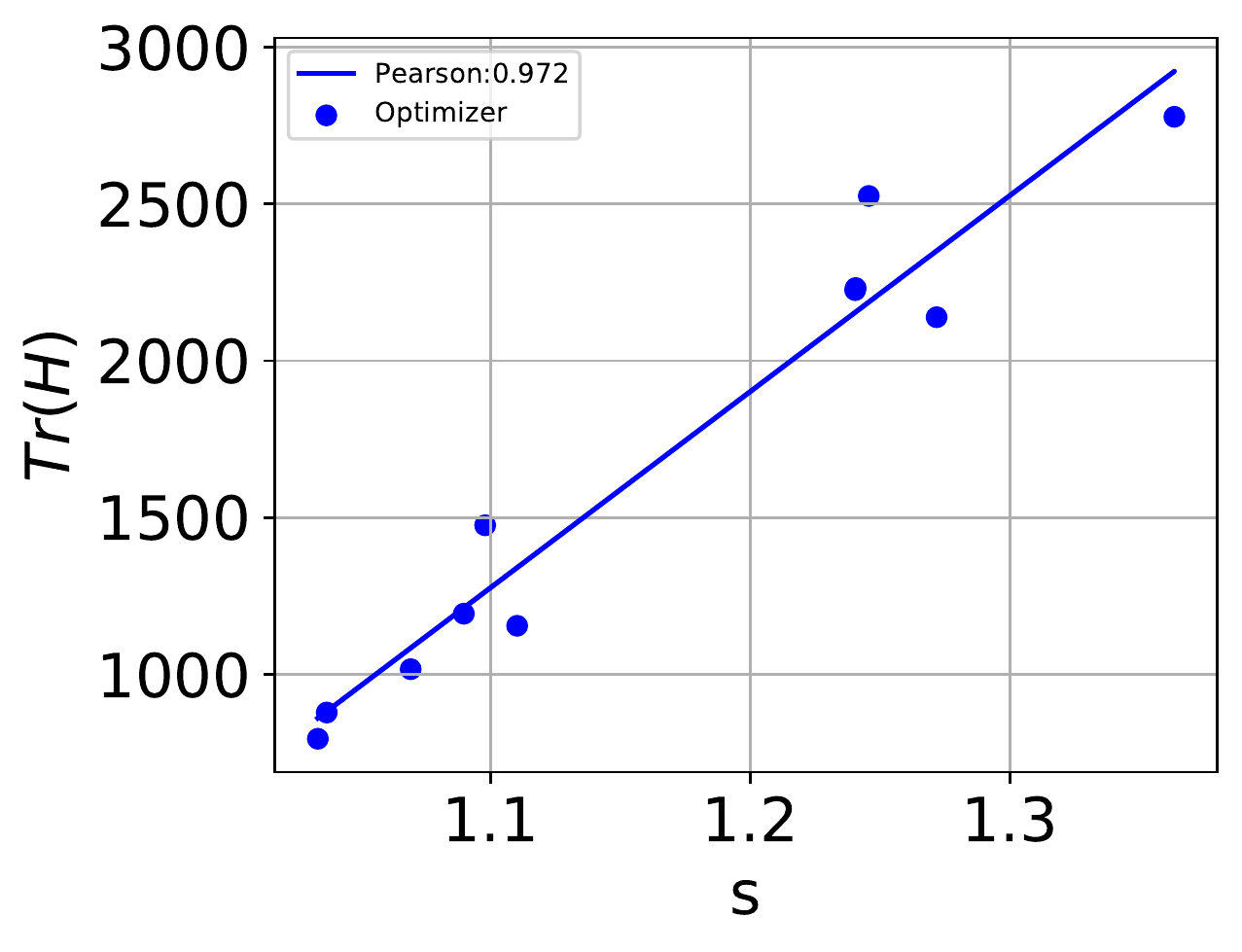}} 
\caption{The slope magnitude $\hat{s}$ closely correlates with the largest Hessian eigenvalue and the Hessian trace. Model:LeNet. Dataset: CIFAR-10.}
 \label{fig:slopesharpness-cifar10}
\end{figure}

The spectrums of LeNet on CIFAR-10 trained via various optimizers are showed in Figure \ref{fig:optimizerspectrum-cifar10}. Figures \ref{fig:slopesharpness-mnist} and \ref{fig:slopesharpness-cifar10} shows that the slope magnitude $\hat{s}$ closely correlates with the largest Hessian eigenvalue and the Hessian trace.

\begin{figure}
\center
\includegraphics[width =0.4\columnwidth ]{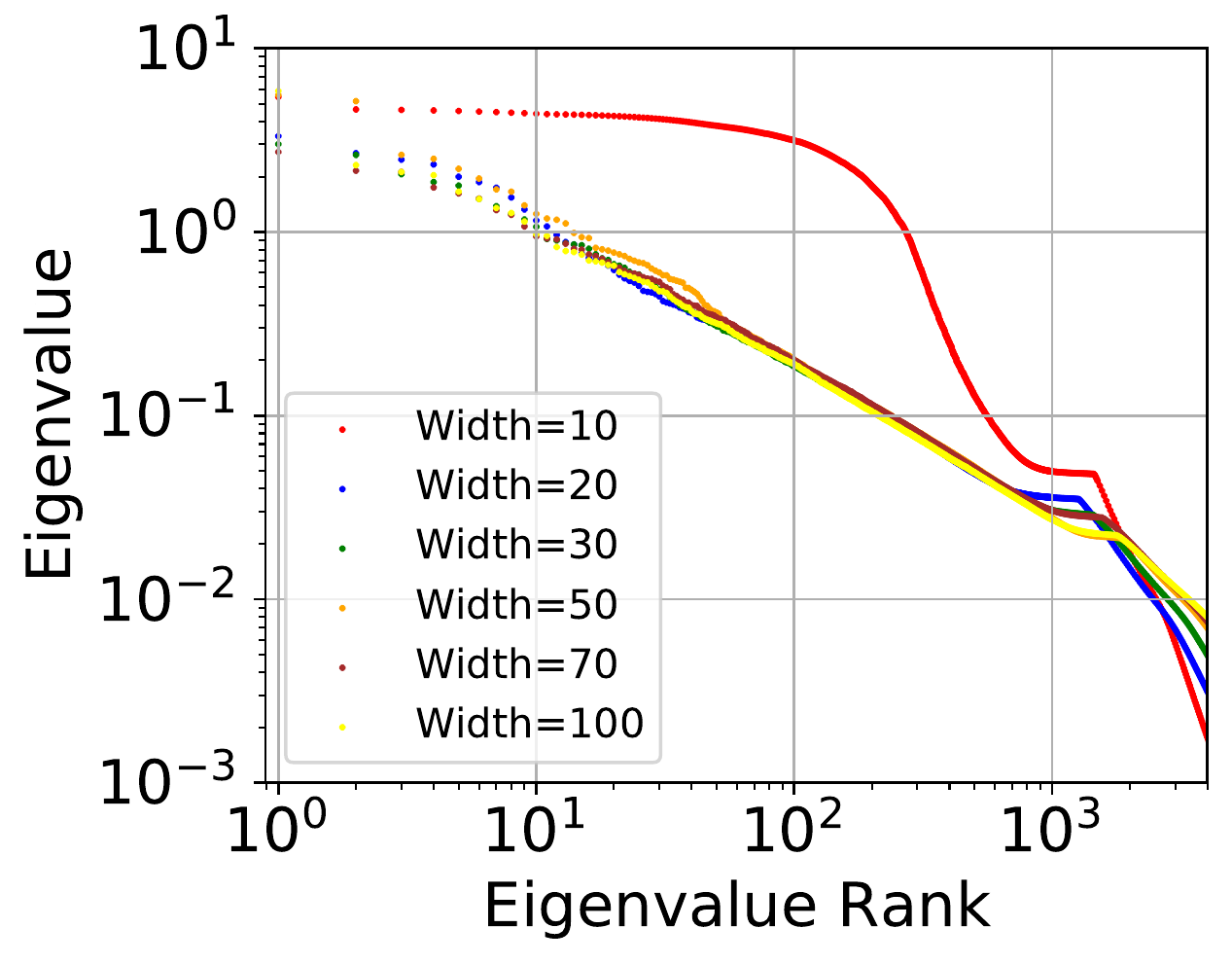}
\caption{The spectrums are not power-law for neural networks with small width($\sim 10$), but gradually become more power-law (more straight in the log-log plot) as the width increases. This may also suggest that the power-law spectrum depends on model capacity. Model: Two-layer FCN. Dataset: MNIST.}
 \label{fig:fcnwidth}
\end{figure}

Figure \ref{fig:fcnwidth} shows that the small width of neural networks may also break the power-law spectrum like small depth. This also supports that overparameterization or large model capacity is necessary for the power-law spectrum. 

\begin{figure}
\center
\subfigure{\includegraphics[width =0.4\columnwidth ]{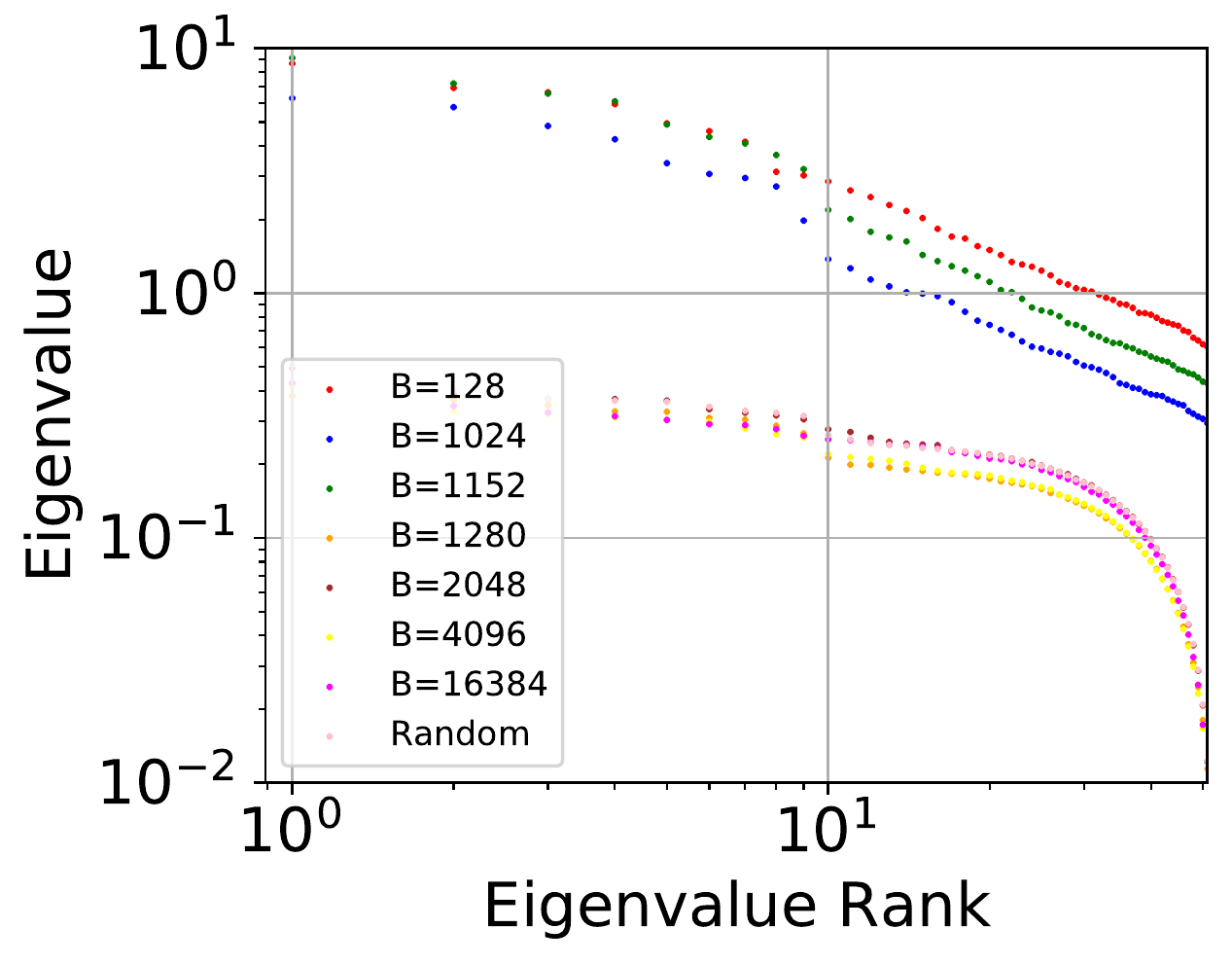}} 
\caption{Batch size matters to the spectrum. Model:ResNet-18. Dataset: CIFAR-10. The sharp phase transition occurs in $1152<B<1280$.}
 \label{fig:batchspectrum-cifar}
\end{figure}

We report the spectrums of large-batch trained ResNet18 on CIFAR-10 in Figure \ref{fig:batchspectrum-cifar}. It indicates that the phase transition behaviors of the spectrums with respect to batch size generally exist. However, it seems that Phase II and Phase III merge into one phase for ResNet18 on CIFAR-10.

 The heavy-tail property of SGD has been a hot and arguable topic recently \citep{simsekli2019tail,panigrahi2019non,gurbuzbalaban2021heavy,hodgkinson2021multiplicative,xie2021diffusion,li2021on}. Note that the power-law distribution is one of the most common heavy-tail distributions in the real world. We argue that the arguable heavy-tail property of SGD may depend on the power-law Hessian spectrum rather than SGD itself, as gradient noise covariance critically depends on the Hessian. We present the power-law spectrums of gradient noise covariance in Figure \ref{fig:noisespectrum}.

\begin{figure}
\center
\includegraphics[width =0.4\columnwidth ]{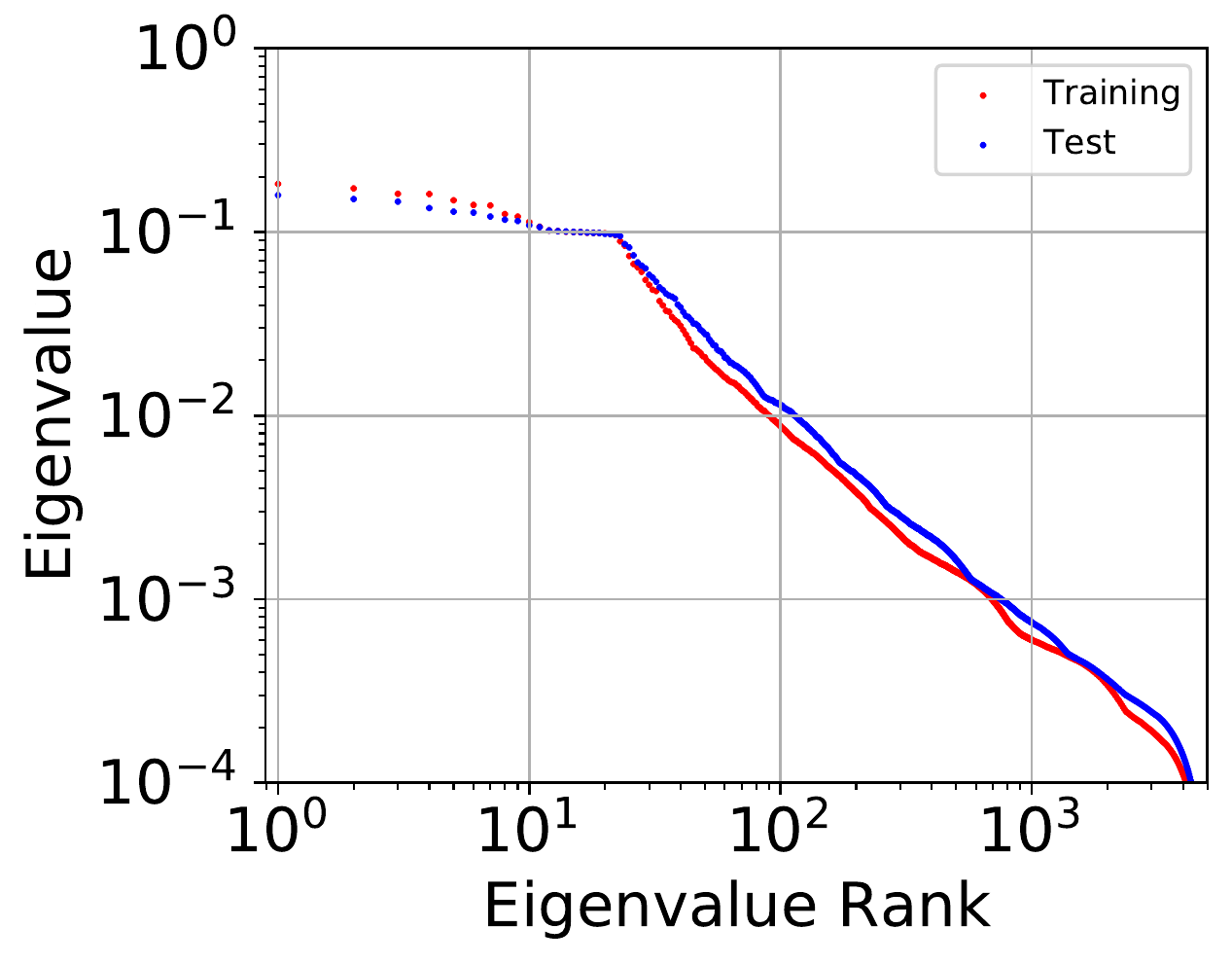}
\includegraphics[width =0.4\columnwidth ]{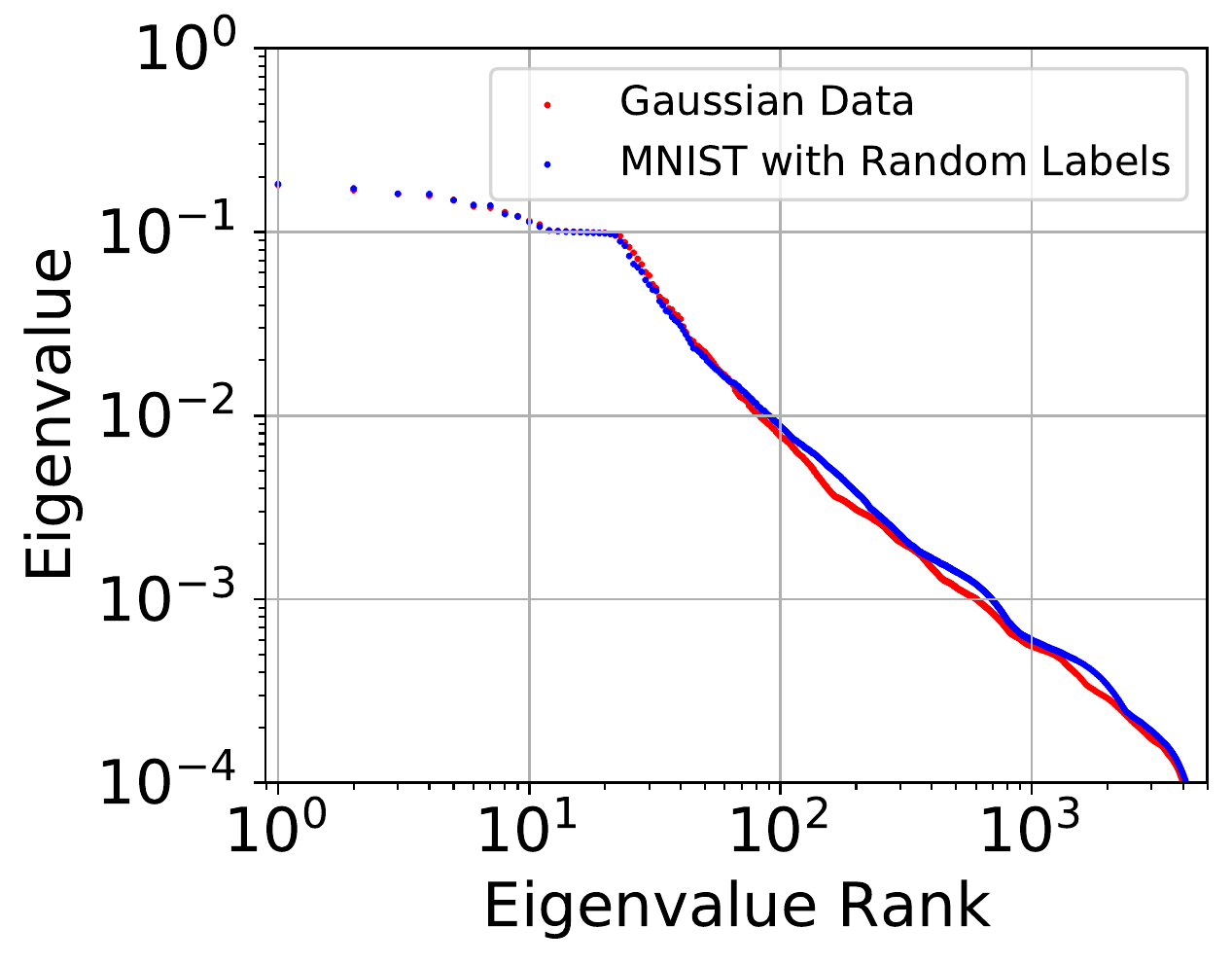}
\caption{ (a) The spectrum of LeNet on (training and test) MNIST with random shuffled labels. (b) The spectrum of LeNet on Gaussian data with random shuffled labels is highly similar to that MNIST with random shuffled labels.}
 \label{fig:randomspectrum}
\end{figure}

We presented the spectrum of learning with clean labels and random labels in Figure \ref{fig:randomspectrum}. The number of top outliers obviously increases, because random labels make the dataset more complex. However, even if the pairs of instances and labels have little knowledge, we still observe the power-law spectrum after the dozens of top outliers. This may suggest that, even if the labels are random, neural networks can still learn useful knowledge from the instances only. 

\end{document}